\documentclass{article}
\usepackage{graphicx} 
\usepackage{amsmath}
\usepackage{amsfonts}
\usepackage{amssymb}
\usepackage{float}
\usepackage{url}
\usepackage[toc,page]{appendix}
\usepackage{tabularx}
\usepackage[margin=1in]{geometry}
\usepackage{color}
\usepackage{longtable}
\usepackage{lineno}

\usepackage{booktabs}
\usepackage{xcolor}
\usepackage{colortbl}

\definecolor{lightgreen}{RGB}{225, 250, 225}
\definecolor{lightblue}{RGB}{220, 235, 255}

\usepackage[normalem]{ulem}

\newcommand{\mstrike}[1]{}

\title{Mantis: A Foundation Model for Mechanistic Disease Forecasting}
\author{Carson Dudley\footnote{Corresponding author, cdud@umich.edu}, Reiden Magdaleno, Christopher Harding, \\ Ananya Sharma, Emily Martin, Marisa Eisenberg}
\date{\today}

\begin{document}

\maketitle


\section*{Abstract}
Infectious disease forecasting in novel outbreaks or low-resource settings is hampered by the need for large disease and covariate data sets, bespoke training, and expert tuning, all of which can hinder rapid generation of forecasts for new settings. To help address these challenges, we developed Mantis, a foundation model trained entirely on mechanistic simulations, which enables out-of-the-box forecasting across diseases, regions, and outcomes, even in settings with limited historical data. We evaluated Mantis against 78 forecasting models across sixteen diseases with diverse modes of transmission, assessing both point forecast accuracy (mean absolute error) and probabilistic performance (weighted interval score and coverage). Despite using no real-world data during training, Mantis achieved lower mean absolute error than all models in the CDC's COVID-19 Forecast Hub when backtested on early pandemic forecasts which it had not previously seen. Across all other diseases tested, Mantis consistently ranked in the top two models across evaluation metrics. Mantis further generalized to diseases with transmission mechanisms not represented in its training data, demonstrating that it can capture fundamental contagion dynamics rather than memorizing disease-specific patterns. These capabilities illustrate that purely simulation-based foundation models such as Mantis can provide a practical foundation for disease forecasting: general-purpose, accurate, and deployable where traditional models struggle.

\section*{Introduction}

Accurate forecasting of infectious disease dynamics is fundamental to effective public health response. During the COVID-19 pandemic, forecasts shaped many aspects of daily life, guiding decisions about how to allocate scarce hospital beds, whether restaurants could reopen, and if children could safely attend school \cite{mcbryde2020role}. Despite their importance, however, disease forecasts often struggle when they are most needed. In the first two years of the COVID-19 pandemic, the performance of many models in the CDC Forecast Hub was worse than a naïve baseline that simply projected the most recent value forward \cite{evaluationpnas}. This highlights a key limitation of many operational systems: accurate, actionable forecasts are most difficult when uncertainty is highest, but this is often also when decisions are most urgent.

Several structural limitations contribute to this challenge. First, many models require large volumes of disease-specific historical data, which are rarely available early in outbreaks.  Second, they rely on strong computational resources and extensive manual tuning by expert teams, limiting scalability and access in under-resourced health departments. Third, they generalize poorly across diseases, locations, or targets, necessitating new models for every context. 

In addition, forecasting horizons are typically relatively short: most models predict 1–4 weeks ahead \cite{evaluationpnas, deepcovid, deepgleam}, in some cases too short-term for public health decisions. Key decisions, such as ramping up hospital capacity, acquiring ventilators, or coordinating vaccination campaigns, often require 6–8 weeks of lead time \cite{einav_surge_2014, dichter_mass_2022}. Without reliable long-range forecasts, these decisions are either delayed or made reactively, often at considerable cost \cite{funk2019ebola}. 
Typically, projections beyond 4 weeks are restricted to scenario analysis, which explores hypothetical outcomes conditional on specific policy interventions or behavioral changes \cite{reich2022, runge2024}. However, unconditional forecasts at longer horizons remain valuable for public health planning. We define forecasts as unconditional projections of expected outcomes, versus scenario analyses that specify hypothetical interventions (i.e. conditional projections). By this definition, temporal horizon does not determine whether a projection is a forecast or a scenario---an 8-week unconditional projection is a forecast, while a 2-week conditional projection is still scenario analysis (given that policy changes, new variants, and other scenarios can, in principle, happen on any timeframe).

Addressing the challenges of limited historical data, poor cross-disease generalization, restricted forecast horizons, and the need for extensive expert tuning requires a different approach to model training. Foundation models offer a promising paradigm: by training large neural networks on massive, diverse datasets, these models learn generalizable representations that enable strong out-of-the-box performance across a wide range of tasks without task-specific retraining \cite{gpt3, alphafold2}. This approach has been highly successful across multiple domains, such as natural language processing and computer vision, where a single pretrained model can be deployed immediately for diverse applications \cite{gpt3, clip}. A foundation model for disease forecasting could provide similar benefits: accurate predictions across pathogens, geographies, and targets without the expert tuning and historical data typically required for each new setting. However, the data requirements for foundation models present a fundamental obstacle in to applications in infectious disease epidemiology. Although disease outbreaks occur near-continuously worldwide, the volume of well-observed outbreak data available for model training is nonetheless limited: many pathogens are not reportable, many outbreaks go undetected, and those captured by surveillance systems vary widely in quality, resolution, and completeness across regions and time periods. Moreover, no historical dataset can span the full diversity of potential pathogens. Real-world disease data alone cannot provide the scale and coverage needed to train a truly general-purpose model.

Mechanistic simulation offers an alternate approach. Epidemiological models encoding disease transmission, progression, and intervention dynamics can generate unlimited synthetic outbreak data spanning scenarios never observed in practice. By training neural networks on these simulations rather than historical observations, we can achieve the data scale required for foundation models while embedding mechanistic scientific knowledge about contagion dynamics directly into the learning process. We previously developed this approach, called Simulation-Grounded Neural Networks (SGNNs), in a separate study, where we demonstrated its effectiveness across multiple scientific domains, including initial tests in epidemiological forecasting \cite{sgnns2025}. However, while that initial exploration demonstrated preliminary results that showed promise for infectious disease forecasting, the model used in that setting lacked the wide range of mechanisms that would be critical for training a general purpose infectious disease foundation model.

\begin{figure}[ht]
    \centering
    \includegraphics[width=0.9\textwidth]{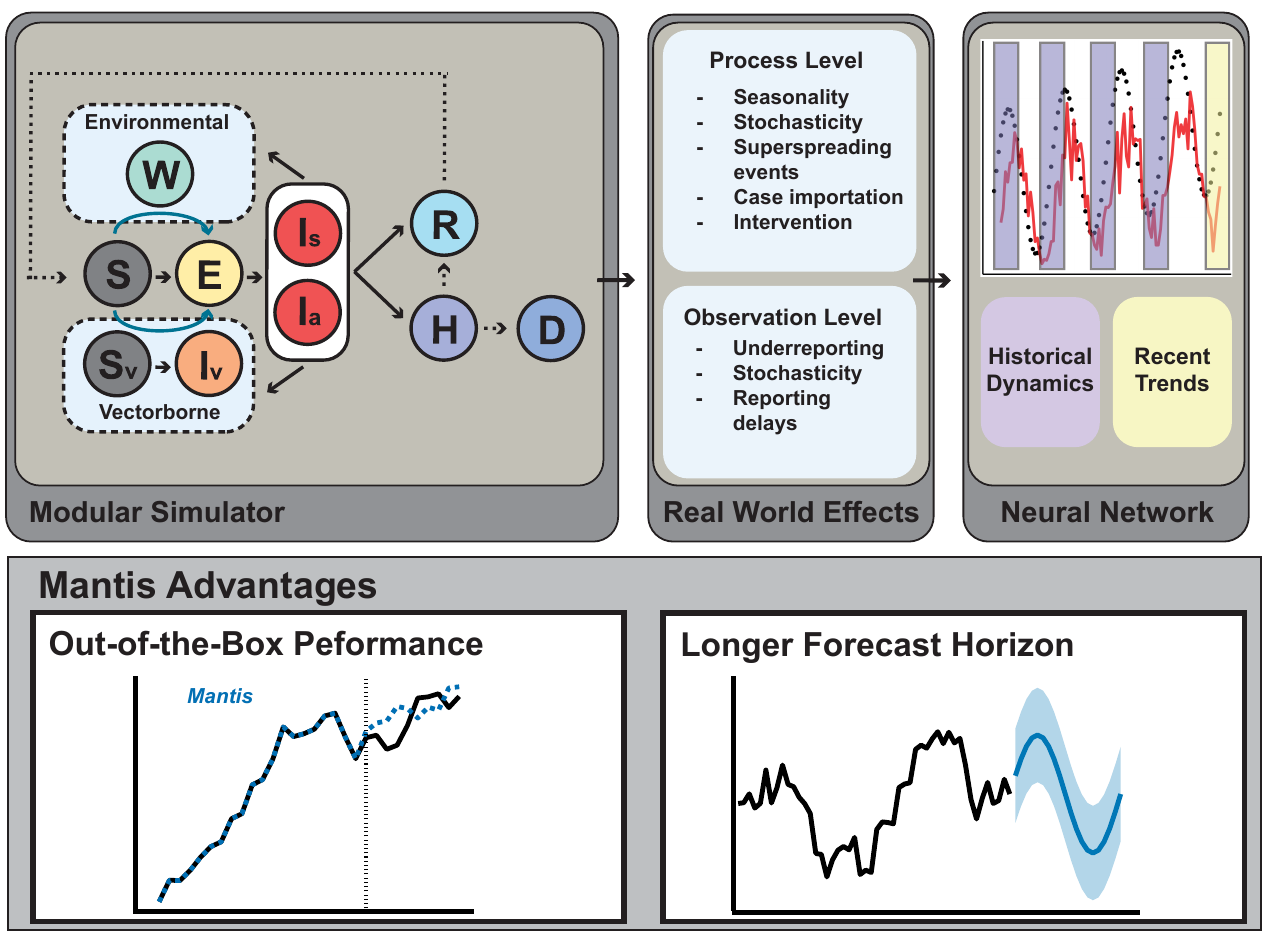}
    \caption{
        \textbf{Conceptual overview of Mantis.}
        Mantis is a simulation-grounded foundation model trained entirely on synthetic outbreaks generated by mechanistic epidemiological models. The training pipeline begins with a modular simulator that encodes diverse outbreak mechanisms, including multiple transmission modes (human-to-human, vectorborne, environmental), progression dynamics, intervention strategies, and population structures. These simulations are then passed through an observation model that incorporates real-world surveillance effects such as underreporting, stochasticity, and reporting delays. Trained on over 400 million simulated days, Mantis forecasts directly from real-world time series at inference time. Mantis enables accurate out-of-the-box performance and longer-range forecasting capability.
    }
    \label{fig:conceptual}
\end{figure}

In this study, we present Mantis, a foundation model for disease forecasting  using the SGNN framework. We expand our training simulations to synthetic outbreaks from mechanistic models that capture transmission, progression, immunity, seasonality, and intervention effects. We simulate human-to-human disease spread, as well as environmental and vector-borne diseases. These simulations encode core epidemiological mechanisms, allowing Mantis to internalize general contagion dynamics rather than purely extrapolating from historical trends. Unlike traditional models, Mantis requires no disease-specific tuning or retraining, but still produces accurate probabilistic forecasts out-of-the-box across diseases, geographies, and targets (Figure \ref{fig:conceptual}).

We evaluated Mantis against 78 models across forecasting tasks spanning 16 real-world diseases, including seasonal influenza, early COVID-19, dengue, hepatitis B, measles, malaria, cholera, and ebola. These evaluations cover multiple transmission modes, outbreak phases, surveillance systems, and historical eras. We assessed both point forecast accuracy and uncertainty quantification. When benchmarked against CDC Forecast Hub models on early pandemic COVID-19 forecasts, Mantis achieved the lowest mean absolute error despite using no COVID-specific data or covariates in training and was tied for the second lowest weighted interval score. Across all other diseases---including those with transmission mechanisms absent from its training data---Mantis consistently ranked among the top two models on all evaluation metrics, demonstrating robust generalization to diverse epidemiological settings.

To support deployment, we release Mantis as an open-source repository. Users can generate forecasts from a single historical time series without any machine learning expertise.

\section*{Results}
We assessed Mantis's forecasting accuracy across a set of real-world forecasting tasks spanning six diseases: seasonal influenza \cite{mistartmap, delphi_epidata}, early COVID-19 \cite{cdccovid}, dengue \cite{tychodengue}, hepatitis B \cite{tychohepatitisb}, smallpox \cite{tychosmallpox}, and scarlet fever \cite{tychoscarletfever}. For each task, we benchmarked Mantis against expert-tuned statistical and machine learning models, including published baselines and submissions to the CDC COVID-19 Forecast Hub \cite{evaluationpnas} (see Methods). We evaluated both point forecast accuracy using mean absolute error (MAE) and probabilistic forecasting performance using weighted interval score (WIS) and coverage metrics (e.g., how often did the 95\% CI actually contain the true value), as well as performing statistical testing using the Diebold-Mariano test (Appendix A). In Appendix \ref{appendix:additionalresults}, we evaluated Mantis's performance on an additional 10 diseases spanning parasitic, zoonotic, and enteric pathogens across multiple other countries, along with formal statistical tests of forecasting improvements. Pretrained models, example data, and an interactive forecasting demo are available at {\texttt{github.com/carsondudley1/Mantis}}.

\subsection*{Mantis Achieves Competitive Performance Across Diseases and Tasks}

For each disease, we evaluated performance across multiple spatial and temporal resolutions, including U.S. states and smaller regions within states. We evaluated both point forecast performance (mean absolute error, MAE) and probabilistic metrics (weighted interval score, WIS; 50\% and 95\% coverage). WIS is the standard probabilistic evaluation metric for disease forecasting \cite{wis}, providing a comprehensive assessment that accounts for both forecast accuracy and uncertainty calibration across predicted quantiles. Following the evaluation framework established in \cite{evaluationpnas}, we report relative MAE and relative WIS, where metrics are normalized by the performance of a naïve baseline that forecasts the current value forward with uncertainty intervals derived from historical variance (see Methods). Relative metrics offer two key advantages over absolute error: they provide an interpretable measure of the information contributed by a model beyond simple persistence, and they enable meaningful comparisons across different forecast targets and geographic scales, where absolute case or death counts may vary by orders of magnitude.

We compared Mantis against a range of published and re-implemented models, including top submissions to the CDC COVID-19 Forecast Hub, statistical baselines (exponential smoothing and SARIMA), LSTM neural networks retrained at each forecast point, and Chronos, a state-of-the-art time-series foundation model \cite{chronos2024}.

Mantis achieved competitive performance across diseases and metrics (Table \ref{tab:results}). For early pandemic COVID-19 forecasts at the state level, Mantis achieved the lowest MAE among evaluated models, including the CDC Forecast Hub ensemble that combines dozens of expert-tuned submissions. The CDC ensemble achieved slightly better WIS (0.61 vs 0.63), while UMass-MechBayes achieved marginally better 95\% coverage (0.93 vs 0.91), demonstrating that multiple approaches can excel on different aspects of probabilistic forecasting. Although results have shown that some general time series forecasting foundation models can confer reasonable results for infectious disease forecasting \cite{time_series_fms}, we did not find that any we tested (Chronos \cite{chronos2024} and TimePFN \cite{timepfn}) gave competitive performance with Mantis or our other baseline models, so we did not include them in our analysis. 

In Appendix \ref{appendix:additionalresults}, we evaluate Mantis' performance specifically in the early weeks of the pandemic, showing that Mantis outperforms both a standard LSTM model and the naive baseline in the early weeks. These results illustrate how the use of mechanistic models in training Mantis can enable it to perform well even in extremely limited data scenarios. Across the 15 additional diseases tested, Mantis consistently ranked first or second on both point forecast accuracy (MAE) and probabilistic performance (WIS).

Example forecast trajectories illustrate Mantis's ability to match the timing and magnitude of real-world outbreaks across epidemiological settings (Figure \ref{fig:mantis_forecasts}, with over 100 more in the supplement).  Forecasts include uncertainty intervals derived from the model's predicted quantiles, enabling direct use in probabilistic planning. Because Mantis is trained entirely on broad mechanistic simulations rather than historical observations, it requires no model fitting or fine-tuning for new targets.

\begin{table}[ht!]
\centering
\caption{Forecasting results across diseases and models. For each metric, the best model is highlighted in \colorbox{lightgreen}{\textbf{green and bold}}, and the second-best is \colorbox{lightblue}{\underline{blue and underlined}}. }
\label{tab:results}
\begin{tabular}{l|l|cccc}
\toprule
\textbf{Disease} & \textbf{Model} & \rotatebox{60}{\textbf{Relative MAE}} & \rotatebox{60}{\textbf{Relative WIS}} & \rotatebox{60}{\textbf{50\% coverage}} & \rotatebox{60}{\textbf{95\% coverage}} \\
\midrule
State-level COVID-19 \cite{cdccovid} & Mantis & \colorbox{lightgreen}{\textbf{0.65}} & \colorbox{lightblue}{\underline{0.63}} & \colorbox{lightgreen}{\textbf{0.51}} & \colorbox{lightblue}{\underline{0.91}} \\
 & CDC ensemble & \colorbox{lightblue}{\underline{0.66}} & \colorbox{lightgreen}{\textbf{0.61}} & \colorbox{lightblue}{\underline{0.53}} & 0.90 \\
 & UMass-MechBayes & 0.67 & \colorbox{lightblue}{\underline{0.63}} & 0.56 & \colorbox{lightgreen}{\textbf{0.93}} \\
 & GT-DeepCOVID & 0.82 & 0.75 & 0.41 & 0.84 \\
 & UCSD\_NEU-DeepGLEAM & 0.78 & 0.83 & 0.70 & \colorbox{lightblue}{\underline{0.91}} \\
 & SARIMA & 1.08 & 1.10 & 0.44 & 0.81 \\
 & LSTM & 1.37 & 1.54 & 0.33 & 0.70 \\
\midrule
Local-scale flu hosp. \cite{mistartmap} & Mantis & \colorbox{lightgreen}{\textbf{0.85}} & \colorbox{lightgreen}{\textbf{0.86}} & \colorbox{lightgreen}{\textbf{0.51}} & \colorbox{lightgreen}{\textbf{0.92}} \\
 & ETS & 1.15 & 1.32 & 0.39 & 0.78 \\
 & LSTM & 1.21 & 1.33 & 0.40 & \colorbox{lightblue}{\underline{0.79}} \\
 & SARIMA & \colorbox{lightblue}{\underline{1.10}} & \colorbox{lightblue}{\underline{1.00}} & \colorbox{lightblue}{\underline{0.47}} & 0.78 \\
\midrule
Dengue cases \cite{tychodengue} & Mantis &\colorbox{lightgreen}{\textbf{0.75}} & \colorbox{lightgreen}{\textbf{0.80}} & \colorbox{lightgreen}{\textbf{0.47}} & \colorbox{lightblue}{\underline{0.90}} \\
 & ETS & 0.99 & \colorbox{lightblue}{\underline{1.04}} & \colorbox{lightblue}{\underline{0.45}} & \colorbox{lightblue}{\underline{0.90}} \\
 & LSTM & 1.34 & 1.62 & 0.37 & 0.73 \\
 & SARIMA & \colorbox{lightblue}{\underline{0.94}} & 1.58 & 0.73 & \colorbox{lightgreen}{\textbf{0.92}} \\
\midrule
Smallpox cases \cite{tychosmallpox} & Mantis & \colorbox{lightgreen}{\textbf{0.83}} & \colorbox{lightgreen}{\textbf{0.84}} & \colorbox{lightblue}{\underline{0.45}} & \colorbox{lightblue}{\underline{0.87}} \\
 & ETS & 1.00 & 1.06 & 0.44 & 0.86 \\
 & LSTM & 1.31 & 1.52 & 0.33 & 0.71 \\
 & SARIMA & \colorbox{lightblue}{\underline{0.95}} & \colorbox{lightblue}{\underline{1.04}} & \colorbox{lightgreen}{\textbf{0.53}} & \colorbox{lightgreen}{\textbf{0.88}} \\
\midrule
ILI syndromic \cite{delphi_epidata} & Mantis & \colorbox{lightgreen}{\textbf{0.83}} & \colorbox{lightgreen}{\textbf{0.88}} & \colorbox{lightgreen}{\textbf{0.46}} & \colorbox{lightgreen}{\textbf{0.89}} \\
& ETS & \colorbox{lightblue}{\underline{1.00}} & 1.05 & 0.43 & 0.83 \\
 & LSTM & 1.19 & 1.30 & 0.41 & 0.80 \\
 & SARIMA & 1.03 & \colorbox{lightblue}{\underline{0.97}} & \colorbox{lightblue}{\underline{0.45}} & \colorbox{lightblue}{\underline{0.88}} \\
\midrule
Scarlet Fever cases \cite{tychoscarletfever} & Mantis & \colorbox{lightgreen}{\textbf{0.80}} & \colorbox{lightgreen}{\textbf{0.81}} & \colorbox{lightgreen}{\textbf{0.47}} & \colorbox{lightgreen}{\textbf{0.91}} \\
 & ETS & 0.89 & \colorbox{lightblue}{\underline{0.95}} & \colorbox{lightblue}{\underline{0.46}} & 0.89 \\
 & LSTM & \colorbox{lightblue}{\underline{0.87}} & 1.04 & 0.36 & 0.80 \\
 & SARIMA & 1.05 & 1.08 & 0.56 & \colorbox{lightblue}{\underline{0.90}} \\
\midrule
Hepatitis B cases \cite{tychohepatitisb} & Mantis & \colorbox{lightblue}{\underline{0.85}} & \colorbox{lightgreen}{\textbf{0.84}} & \colorbox{lightgreen}{\textbf{0.47}} &\colorbox{lightblue}{\underline{0.90}} \\
 & ETS & 0.86 & \colorbox{lightblue}{\underline{0.89}} & \colorbox{lightblue}{\underline{0.46}} & 0.88 \\
 & LSTM & \colorbox{lightgreen}{\textbf{0.82}} & 1.07 & 0.37 & 0.78 \\
 & SARIMA & 0.86 & \colorbox{lightgreen}{\textbf{0.84}} & 0.58 & \colorbox{lightgreen}{\textbf{0.91}} \\
\bottomrule
\end{tabular}
\end{table}

\subsection*{Covariates Improve Forecast Accuracy Across Horizons}

We evaluated two variants of Mantis to assess the impact of additional information on forecasting performance: a covariate-free version, which receives only the time series to be forecast (e.g., deaths), and a covariate-aware version, which also receives related signals (e.g., cases or hospitalizations). (See Methods for training details for each model.)

For example, when forecasting COVID-19 deaths, the covariate-aware Mantis version (used in evaluations) receives both historical deaths and upstream hospitalization data, while the covariate-free version sees only the death time series. Both models use identical architecture and training data; the only difference is the presence or absence of additional input channels.

\begin{figure}[ht!]
    \centering
    \includegraphics[width=0.95\textwidth]{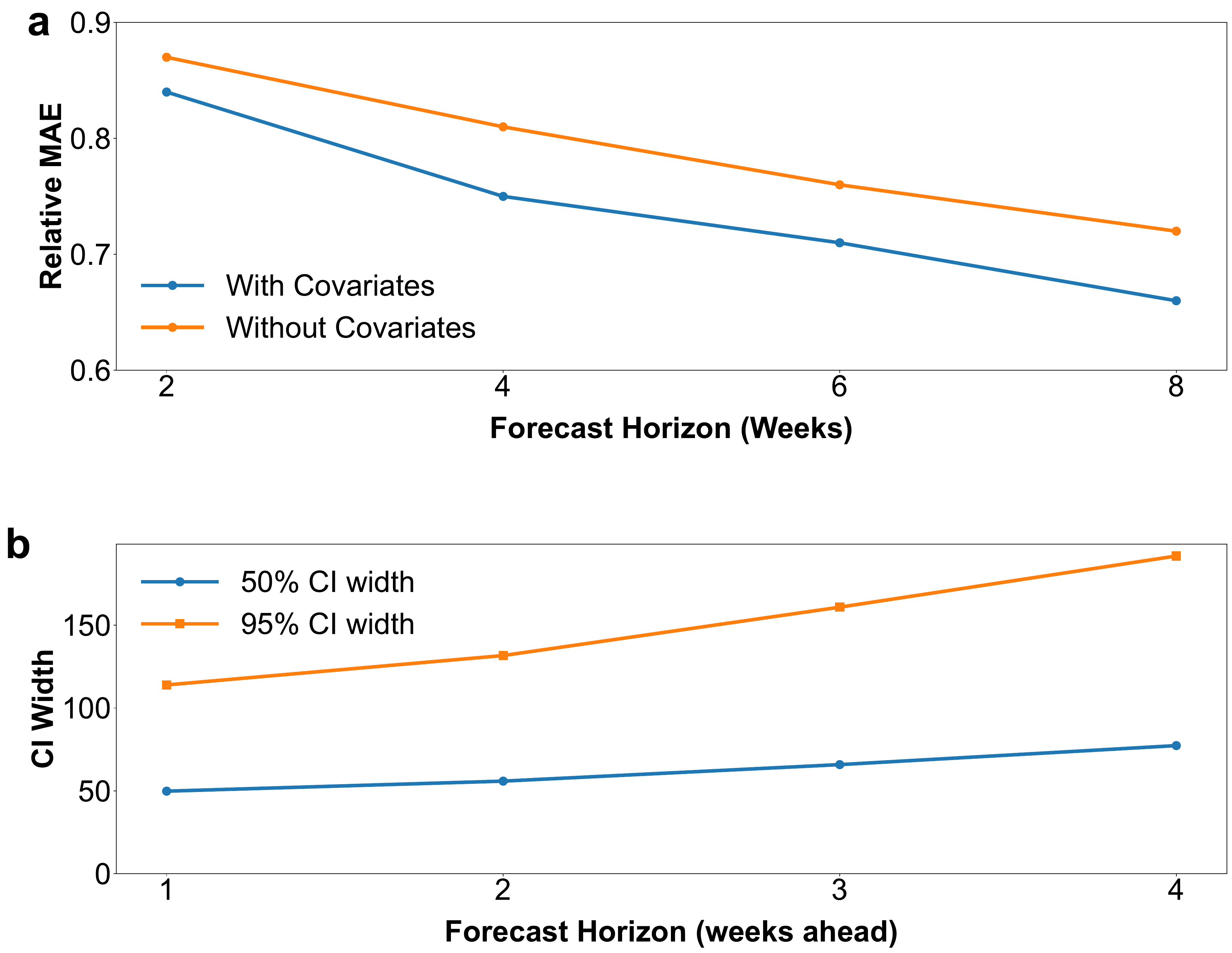}
    \caption{
        \textbf{Mantis performance and uncertainty across forecast horizons with and without covariates.}
        \textbf{(a) Covariate integration improves accuracy. } Including covariates (e.g., using cases to predict hospitalizations) consistently improves Mantis's accuracy across all forecast horizons. Relative MAE vs. naive baseline shown for COVID-19 mortality forecasts with (blue) and without (orange) hospitalization covariates across 2, 4, 6, and 8-week horizons.
        \textbf{(b) Mantis maintains calibrated uncertainty across forecast horizons.} Mantis's prediction interval widths increase appropriately with forecast horizon, reflecting growing uncertainty in longer-range predictions. The 50\% confidence interval width (blue) grows from approximately 50 to 80 deaths, while the 95\% interval width (orange) expands from approximately 110 to 180 deaths over 1 to 4 weeks ahead when forecasting COVID-19 mortality. This systematic widening demonstrates the model becoming appropriately less confident as prediction horizon increases.
    }
    \label{fig:covariate_uncertainty}
\end{figure}

Across forecast horizons (2, 4, 6, and 8 weeks), the covariate-aware version consistently outperformed its covariate-free counterpart (Figure \ref{fig:covariate_uncertainty}a), achieving lower mean absolute error across diseases and targets. These improvements were particularly pronounced at longer horizons, where upstream indicators (e.g., cases leading to hospitalizations, then deaths) provide early signals that improve downstream predictions.

These results highlight the benefit of incorporating complementary epidemiological signals when available and demonstrate that Mantis can flexibly leverage additional information without retraining. In data-rich settings, covariate-aware forecasts yield improved performance; in low-data contexts, the covariate-free version retains strong out-of-the-box accuracy, making both variants useful in practice depending on context and data availability.

\subsection*{Mantis Enables Accurate Forecasts at Longer Horizons}

We evaluated Mantis’s long-range forecasting performance at an 8-week horizon across the same benchmark tasks and baselines described above. Mantis consistently achieved the lowest relative MAE across all diseases and targets tested, demonstrating its ability to extrapolate epidemiological dynamics further into the future than prior models. At the 8-week horizon, Mantis achieved lower relative error than all models in the CDC Forecast Hub at 4 weeks, and lower raw error than the median 4-week error of Forecast Hub models, demonstrating long-range accuracy that exceeds the short-term limits of some existing systems.

\begin{figure}[ht!]
    \centering
    \includegraphics[width=0.9\textwidth]{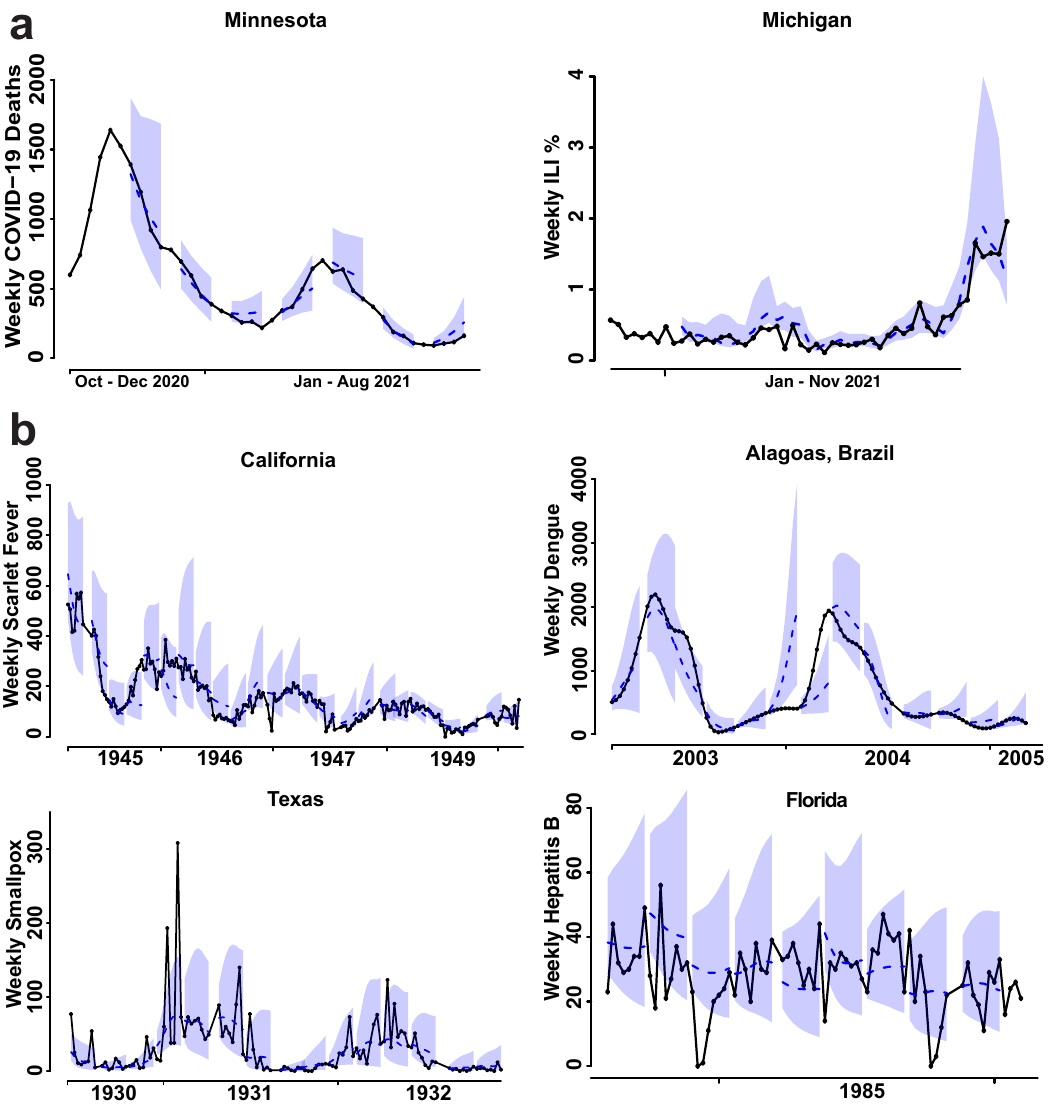}
    \caption{
        \textbf{Mantis Produces Accurate and Generalizable Forecasts Across Diseases and Geographies.}
        \textbf{(a)} Four-week-ahead forecasts (blue dashed line and shaded 90\% CI) compared to observed outcomes (black) for COVID-19 mortality in Minnesota and influenza-like illness (ILI) in Michigan. In the latter, Mantis demonstrates its foundation model capacity by accurately forecasting syndromic inputs despite never being trained with syndromic data.
        \textbf{(b)} Eight-week-ahead forecasts for four historical outbreaks, highlighting Mantis’s ability to generalize across time, space, and transmission mode. In Alagoas, Brazil, Mantis forecasts the 2004 dengue surge prematurely, highlighting a failure in temporal calibration due to the absence of covariates, but still estimates the peak height with high accuracy despite being 18 weeks early. Forecasts for hepatitis B in Florida---a chronic, bloodborne disease type absent from Mantis’s training set---also demonstrate its capacity to generalize to out-of-distribution disease profiles while maintaining forecast stability and accuracy.
    }
    \label{fig:mantis_forecasts}
\end{figure}

Visual examples of 8-week forecasts are shown in Figure \ref{fig:mantis_forecasts}b. Prediction intervals widen with time (figure \ref{fig:covariate_uncertainty}b), maintaining calibrated uncertainty that reflects forecast difficulty (calibration results in Table \ref{tab:results}).

This capability is not enabled by manual tuning or disease-specific details, but by scale: Mantis learns from millions of simulated outbreak trajectories, encompassing diverse transmission patterns and intervention regimes. Unlike historically trained models, which rely on comparatively narrow slices of observed data, simulation-trained models like Mantis can internalize general patterns that support long-range prediction.

\subsection*{Testing Mantis as a Foundation Model: Generalizing to New Mechanisms and Inputs}

We next evaluated Mantis's out-of-the-box forecasting capacity across both held-out transmission mechanisms and new input data modalities.

First, we benchmarked Mantis on diseases governed by transmission and progression mechanisms never represented in training simulations. For example, hepatitis B exhibits bloodborne transmission and chronic infection dynamics that are markedly different from the acute, self-limiting outbreaks used in pretraining. Although the evaluation data consist of reported acute cases from U.S. notifiable disease surveillance, the underlying biology of hepatitis B differs fundamentally from Mantis's training simulations: the Mantis training set included no models without recovery (all simulated diseases eventually transition out of the infectious state, with the longest average infectious period of just 10 days), whereas hepatitis B infection can persist for years with chronic carriage. Despite this fundamental mechanistic mismatch, Mantis achieved competitive performance on hepatitis B forecasts, ranking second in MAE (0.85 vs 0.82) and achieving the best WIS (0.84) among models we tested (Table \ref{tab:results}). These results demonstrate that Mantis can generalize across mechanistic boundaries---not just to new pathogens, but to qualitatively different epidemic regimes, such as chronic persistence versus outbreak spread. However, it is important to note that this forecasting task is limited: hepatitis B data exhibit relatively stable but noisy dynamics without substantial outbreak structure. While Mantis performs well compared to baselines, there is relatively little signal to exploit beyond tracking a noisy equilibrium, and 8-week forecasts of endemic incidence are rarely a high-priority public health need. The point of this benchmark is not to demonstrate operational utility for hepatitis B, but to show that Mantis generalizes to qualitatively different epidemic regimes rather than memorizing a narrow set of outbreak patterns from training. However, given the relatively simple dynamics here, further testing of this point would be a useful step for future work.

Next, we tested Mantis's ability to generalize across data modalities. Although trained using cases, hospitalizations, and deaths, Mantis can incorporate new input types at inference time. We evaluated Mantis on syndromic surveillance data (percent of outpatient visits for influenza-like illness), a data type never represented in training simulations. Despite this distribution shift, Mantis achieved the lowest MAE (0.83) and WIS (0.88) on ILI forecasting among models we tested (Table \ref{tab:results}), demonstrating generalization to novel surveillance indicators. This capability is especially valuable in settings where conventional case counts may be missing or delayed, but syndromic indicators are available in near real-time.

Across all seven disease forecasting tasks, spanning respiratory, vector-borne, bloodborne, and syndromic targets, Mantis consistently ranked first or second on both point and probabilistic metrics (Table \ref{tab:results}). This level of generalization, across diseases, mechanisms, and modalities, demonstrates Mantis's capacity as a foundation model and the advantages of scaling afforded by simulation data. Rather than relying on domain-specific tuning, Mantis captures structural features of contagion that transfer across contexts, enabling immediate deployment in settings where other models would require retraining, expert tuning, or disease-specific data collection.

In Appendix \ref{appendix:additionalresults}, we extend these generalization tests. We evaluate Mantis on 10 additional diseases across Asia and Africa, including additional pathogens with transmission mechanisms not represented in training (e.g., Ebola, zoonotic spillover with contact transmission). In addition to new diseases, we also test Mantis's ability to incorporate entirely novel input modalities at inference time. Despite being trained exclusively on cases, hospitalizations, and deaths, Mantis was able to leverage weather covariates (e.g., temperature, rainfall) to improve forecasts for climate-sensitive diseases in Thailand, and could use disease trajectories from adjacent locations as spatial covariates to improve COVID-19 death forecasts in U.S. states.

\subsection*{Ablation Studies: The Role of Scale and Mechanistic Pretraining}

To understand the key factors driving Mantis's performance, we conducted two ablation studies examining the roles of architectural capacity and mechanistic pretraining in enabling accurate out-of-the-box forecasting.

\subsubsection*{Architectural Capacity Requirements}

We first evaluated whether Mantis's forecasting capabilities could be achieved with a simpler, more compact architecture. Using the same 400 million days of synthetic training data, we replaced Mantis's hybrid CNN-transformer architecture with the dual-branch LSTM architecture from DEFSI \cite{defsi}. This LSTM model contains approximately 200 times fewer parameters than Mantis, making it significantly more lightweight and computationally efficient.

Despite identical training data and procedures, the reduced architectural capacity led to substantially degraded performance. When evaluated on COVID-19 mortality forecasting across U.S. states, the LSTM architecture achieved mean absolute error (MAE) that was 89\% higher than the full Mantis model. This demonstrates that while synthetic training is useful, it requires sufficient architectural capacity to effectively learn and generalize from the diverse mechanistic patterns encoded in the simulation data.

\subsubsection*{Mechanistic Pretraining is Essential for Novel Outbreak Performance}

To isolate the contribution of mechanistic pretraining, we trained the full Mantis architecture exclusively on real-world data, eliminating simulations from the training process. We evaluated on COVID-19 death forecasting. We implemented a rolling training strategy that provided the strongest possible baseline: the model was initially trained on the first few months of pandemic data, then progressively retrained with expanding datasets as more real-world observations became available. This approach mimics operational practice and ensures that the real-world-only model has access to all available historical information at each forecast point.

The results demonstrate the importance of mechanistic pretraining. When trained only on real-world COVID-19 data, forecast error increased by 262\% compared to the simulation-pretrained Mantis model. This performance degradation reflects the fundamental challenge of learning outbreak dynamics from limited historical data during a novel pandemic. Real-world training data, even when accumulated over months, cannot capture the full spectrum of epidemic behaviors that might emerge. In contrast, the simulation-pretrained model draws from hundreds of millions of days of diverse outbreak scenarios, enabling it to recognize and extrapolate patterns even in unprecedented contexts.

These ablation results underscore the core insight behind Mantis: architectural capacity provides the necessary capability to represent complex patterns, while large-scale simulation-grounded training paradigm provides abundant data to learn generalizable principles of epidemic dynamics and utilize the enhanced architectural capacity. Neither component is sufficient in isolation. Rather, the combination of high-capacity modern neural architectures and large-scale mechanistic simulations enables out-of-the-box generalization to novel diseases and outbreak contexts.

\section*{Discussion}

Our results using Mantis demonstrate that accurate forecasting across diverse diseases can be achieved through training on simulated outbreaks, without access to real-world data. By learning from mechanistic simulations, Mantis delivers competitive performance across real-world forecasting tasks spanning multiple transmission modes, outbreak phases, and surveillance systems. These capabilities position Mantis and other simulation-based foundation models as a practical tool for public health use, especially in low-resource or emerging outbreak settings where historical data, computational resources, and specialized modeling expertise are often limited. Because Mantis can generate forecasts with only a short initial observation window, it enables early situational awareness when conventional models require longer histories to achieve comparable accuracy. Mantis's advantage in data-scarce settings stems from its ability to avoid overfitting to sparse initial data. Rather than extrapolating from a handful of observations, Mantis produces appropriately hedged forecasts that reflect the true uncertainty across potential outbreak trajectories in our evaluations, enabling more reliable probabilistic planning even when historical context is minimal. Mantis avoids overfitting because using synthetic data allows us to generate as many simulations as necessary, avoiding the overfitting or memorization risks inherent in epoch-based training on fixed datasets. 

This approach offers a potential solution to a fundamental challenge in epidemiological modeling: real-world disease data are limited by data availability, reporting inconsistencies, and lack of surveillance of most outbreaks. Simulations, by contrast, can generate unlimited training data spanning diverse scenarios. Whether simulation-based training ultimately proves superior to training on massive real-world datasets---if such data could be assembled---remains an open question. However, given current data availability constraints, the simulation-based approach offers a practical path for building general-purpose forecasting systems. 

It is important to note, though, that this approach depends critically on developing simulations that are sufficiently realistic and diverse to capture the dynamics a model will encounter in practice. Our results suggest that a broad distribution of acute outbreak simulations can generalize even to chronic disease dynamics not explicitly modeled, but the boundaries of this generalization are unclear. Future work should investigate when simulation-trained models fail and whether certain epidemic regimes require explicit representation in training, or whether sufficiently diverse simulations can approximate arbitrary contagion dynamics.

Beyond its practical utility, Mantis's ability to generalize across transmission mechanisms raises a more fundamental question: why does a model trained on acute respiratory, vector-borne, and environmental outbreak simulations transfer to diseases with substantially different mechanisms and natural histories, including chronic bloodborne infections it has never encountered? One interpretation is that epidemic time series, regardless of their generating mechanism, share common dynamical signatures (e.g., acceleration, deceleration, peak structure, seasonal modulation, noise characteristics) that are largely determined by the interaction between transmission, immunity, and observation processes rather than by the specific biological route of transmission. Indeed, many common patterns are well-known in infectious disease epidemiology (e.g.  stuttering chains, exponential growth, and epidemic, endemic, and seasonal illnesses, among many others \cite{brouwer2026infectious, cobey2020modeling, keeling2008modeling, blumberg2013inference}).  Mantis appears to learn these shared structural features, enabling it to recognize and extrapolate epidemic patterns even when the underlying mechanism differs from anything in its training data.

This interpretation connects to recent work identifying universal statistical properties of epidemics and universality in disease dynamics \cite{marani2021, paul2021emergence,das2021scaling}. For example, \cite{marani2021} found that epidemic intensity follows a common power-law distribution across centuries of outbreaks from diverse pathogens, suggesting that fundamental regularities govern epidemic dynamics independent of specific transmission biology. Our results provide complementary evidence from a modeling perspective: if epidemic dynamics were not governed by such shared structure, a model trained on one class of simulations should not generalize to mechanistically distinct diseases. Together, these findings point toward potentially universal patterns in contagion dynamics that may inform both theoretical understanding and practical forecasting.

Mantis represents an initial attempt at a foundation model for disease forecasting, and several important questions remain unresolved. First, the optimal neural architecture for this task is unclear. While our hybrid CNN-transformer design performs well, alternative architectures may offer better performance or efficiency trade-offs. Second, the ideal balance between simulation diversity and computational cost requires further investigation. Our training set spans 400 million simulated days, generated continuously until model convergence without repeating data. Whether this scale is necessary or whether comparable performance could be achieved with fewer but more strategically designed simulations remains to be seen. More real-time benchmarking of Mantis will also more clearly demonstrate the settings where it succeeds and where it struggles, which will become more clear in coming seasons.

While Mantis shows broad generalization, several limitations merit consideration. Though we demonstrate generalization beyond the specific regimes directly simulated during training, the model's performance depends on the diversity and realism of training simulations. Additionally, Mantis inherits any systematic biases present in the underlying mechanistic models used for training. If the simulation framework systematically misrepresents certain disease dynamics---for example, categorically overestimating intervention effects or inadequately capturing stochasticity in small populations, which are typically underrepresented in forecasting evaluation, Mantis included---these biases will propagate to the trained model. Future work should explore approaches to identify and fill critical simulation gaps, and develop methods to validate that training simulations adequately span the space of real-world disease dynamics. 

An important limitation of the current framework is the absence of explicit spatial structure. Each forecast is generated from a single location's time series, without incorporating mobility networks or spatial coupling between regions. However, numerous studies have demonstrated the importance of spatial dynamics in disease spread (e.g. \cite{riley2007large, sattenspiel2009geographic, chowell2018spatial, bertuzzo2010spatially}, among others). For example, work on Italy's COVID-19 epidemic demonstrated that transport mechanisms and inter-provincial mobility were essential for modeling the spatial propagation of outbreaks across regions and quantifying the impact of geographically targeted containment measures \cite{gattospatial}. While Mantis addresses a different task (forecasting the temporal trajectory of an outbreak within a given location rather than its spatial spread), spatial information could nonetheless improve single-location forecasts. In a preliminary evaluation (Appendix~\ref{appendix:additionalresults}), we found that providing Mantis with case data from an adjacent state as a covariate modestly improved COVID-19 death forecasts ($\sim$5\% reduction in MAE across $\sim$4,000 forecasts), despite the model never having been trained on spatially coupled data. Mantis's multi-wave training simulations capture some spatial effects implicitly, as abrupt shifts in transmission rates can mimic the arrival of epidemic waves from connected regions, but cannot represent the mechanistic process of spatial spread. A model explicitly trained on spatially coupled simulations would likely yield larger gains. Incorporating metapopulation dynamics or mobility-informed covariates into the simulation framework represents a natural extension that could improve forecasting performance, particularly for anticipating incoming waves before they manifest in local surveillance data.

Mantis illustrates a potential path forward for general-purpose disease forecasting systems that can be quickly deployed without requiring disease-specific tuning. Its out-of-the-box generalization due to having been fully trained prior to use makes it especially well-suited for early outbreak response and low-resource environments---settings where traditional models often fail due to sparse or delayed data, or due to the expertise and computational needs involved in training modern machine learning models. Because Mantis has been pre-trained prior to deployment on simulation data, it can be run rapidly and with comparatively little technical knowledge relative to most AI/ML models. It is also important to note that in many settings, surveillance reporting lags of 6--8 weeks mean that even an 8-week forecast is partly a nowcast of current conditions. However, Mantis's extended forecast horizon remains valuable in these contexts: a model limited to 4-week-ahead predictions, under such reporting delays, provides no forward-looking information at all. Moreover, Mantis can be integrated into modular pipelines alongside existing nowcasting methods: a nowcasting system could first estimate current disease levels from incomplete recent data, and Mantis could then generate forward-looking forecasts from that estimate as input. By pairing simulation-based priors with flexible neural architectures, this approach offers a scalable alternative to expert-tuned, disease-specific models. In future work, we plan to make this approach even more accessible by developing a web-based interface that will allow forecast generation from any device with a browser. Nonetheless, as a first demonstration of simulation-grounded foundation models in infectious disease forecasting, Mantis establishes both the promise and the open challenges of this paradigm for future decision support systems in global public health.

\section*{Methods}

A complete description of the methods is given in the Appendix.

\subsection*{Training Mantis on Simulated Outbreaks}

Mantis was trained entirely on synthetic outbreaks generated using mechanistic epidemiological models, following the framework developed in \cite{sgnns2025}. Simulations were based on compartmental disease models with stochastic transitions and heterogeneous parameterizations, using three primary transmission routes: human-to-human respiratory illness models, vectorborne disease models (primarily inspired by arbovirus dynamics), and environmental transmission models (inspired by cholera and other environmentally transmitted pathogens). These models varied in transmission dynamics, disease severity, and intervention strategies to create a broad and diverse training distribution (see Appendix~\ref{appendix:simulation} for details). Each simulation generated daily time series of cases, hospitalizations, and deaths under a wide range of epidemic scenarios. In total, the training dataset included over 400 million days of data spanning thousands of parameter combinations.

\subsection*{Neural Architecture and Training Procedure}

Mantis uses a sequence-to-sequence neural network with convolutional and transformer-based components (Appendix~\ref{appendix:architecture}). The model takes an input context window of up to 112 weeks of observed time series and outputs probabilistic forecasts for multiple future targets (e.g., cases, hospitalizations, deaths). Forecasts are trained using a weighted quantile loss. We optimized the neural network using AdamW \cite{adamw}. No real-world data are used during training.

\subsection*{Evaluation on Real-World Forecasting Tasks}

We evaluated Mantis across six diverse real-world infectious disease forecasting scenarios to assess generalization across pathogens, transmission modes, geographic scales, and historical periods (Appendix~\ref{appendix:evaluation}). Our evaluation comprised two primary tasks and four additional challenges representing varying degrees of distribution shift from the model’s training data. Together, these datasets encompass over 39 forecasting tasks and more than 150,000 individual forecast evaluations, enabling rigorous assessment of Mantis’s capabilities across diverse epidemiological conditions, spatial resolutions, and outcome types. All Mantis evaluations were conducted in a strict zero-shot setting, with no fine-tuning or retraining on any evaluation data.

\subsection*{Baseline Models}
We compared Mantis against two classes of baseline models: operational forecasts from the CDC COVID-19 Forecast Hub and models we re-implemented for evaluation across all diseases. For the CDC COVID-19 Forecast Hub comparison, we used forecasts as originally submitted by modeling teams during the pandemic and, to the best of our ability, evaluated both them and Mantis using the data as it was available at the time of each forecast (i.e., without any backfill or cleaning). This retrospective evaluation means our dataset contains data quality issues (e.g., missing/underreported data) that were present in the original surveillance data and likely affected all models' performance, including Mantis. These CDC models were retrained or refitted each week to incorporate newly available data, and modeling teams could manually adjust forecasts if their automated systems produced anomalous predictions. Mantis forecasts were not modified in any way. This represents the strongest possible real-world baseline, reflecting operational forecasting practice with expert oversight under realistic data conditions.

For baselines we implemented---including statistical models (exponential smoothing and SARIMA) and deep learning models (LSTM)---we likewise retrained or refitted each model at every forecast point as new data became available. This rolling retraining procedure provides the strongest possible baseline performance for each method. Full baseline model implementation details can be found in Appendix \ref{appendix:evaluation}.

To compute relative MAE and relative WIS, we normalized all forecast errors by the performance of a naïve baseline model. This baseline generates point forecasts by projecting the most recent observed value forward to all future time points. Uncertainty intervals are constructed using the procedure from the COVIDhub-baseline model \cite{evaluationpnas}: we collect all first differences from the historical time series along with their negatives (i.e., $y_t - y_{t-1}$ and $-(y_t - y_{t-1})$ for all past times $t$). We form a piecewise linear approximation to the empirical cumulative distribution function of these differences, then generate 100,000 Monte Carlo samples by independently drawing sequences of $h$ differences and adding them to the most recent observed value. The resulting distribution is truncated at zero (to prevent negative forecasts) and summarized using quantiles, with the median set equal to the last observed value to eliminate sampling noise. This baseline captures the natural variability in historical changes without making assumptions about future trends, providing a neutral reference point for evaluating forecast skill.

\section*{Author Contributions}
CD conceived of the work, led the study, and developed the code and initial draft of the manuscript. CH and AS did the ablation studies and RM developed figure visualizations. CD, MCE, and RM all contributed to the analysis plan and made substantial edits and revisions for the final manuscript. 

\section*{Acknowledgments}
This project was made possible by the Insight Net cooperative agreement with University of Michigan (5 NU38FT000002-02-00) from the CDC’s Center for Forecasting and Outbreak Analytics (CDC-RFA-FT-23-0069). Its contents are solely the responsibility of the authors and do not necessarily represent the official views of the Centers for Disease Control and Prevention. The authors thank Julie Gilbert for her assistance with gathering data for an earlier version of the manuscript. The authors thank Kevin M. Bakker for his assistance in finding and collecting data from the Thai Ministry of Public Health. The authors also thank the two anonymous reviewers whose comments have greatly strengthened the manuscript.

\bibliographystyle{unsrt}

\begin{appendices}

Below we provide additional results (Section~\ref{appendix:additionalresults}), a more complete description of the methods, including the mechanistic simulators used for training (Section~\ref{appendix:simulation}), the model architecture (Section~\ref{appendix:architecture}), and the evaluation tasks and data (Section~\ref{appendix:evaluation}).

\section{Additional Results}\label{appendix:additionalresults}

\subsection{Mantis Maintains Accuracy Early in Novel Outbreaks Without Overfitting}
\label{appendix:earlypandemic}

A critical challenge for data-driven forecasting models is performance during the early phase of a novel outbreak, when only a few weeks of observations are available. We analyzed Mantis's relative MAE over the first 32 weeks of the COVID-19 pandemic across all U.S. states (representing roughly the first wave, with models beginning generating forecasts after 8 weeks), comparing against an LSTM baseline \cite{LSTM} retrained at each forecast point using all available data. Both models are evaluated against the same naïve persistence baseline described in the Methods, and relative MAE below 1.0 indicates improvement over that baseline.

The results show a clear divergence between the two approaches. From the first forecast week, Mantis outperforms or matches the naïve baseline while the LSTM performs substantially worse, beginning at roughly 1.5 times the naïve baseline error and deteriorating to nearly 3 times by week 31. Mantis, by contrast, drops below the naïve baseline within the first few forecast weeks, with only a brief and modest rise near the peak of the first wave (weeks 22-23) consistent with the well-documented challenge of forecasting epidemic turning points. By the end of the evaluation window, Mantis achieves relative MAE of approximately 0.6 — 40\% better than the naïve baseline — while the LSTM reaches nearly 3.0.

This divergence reflects a structural challenge specific to novel outbreak settings rather than a general limitation of data-driven models. The LSTM's deteriorating performance does not occur because more data is harmful in general — indeed, LSTMs trained on large, diverse datasets can perform well. Rather, the problem is that the data accumulating during this window consists entirely of a single, still-incomplete epidemic wave: the model is repeatedly retrained on an expanding but structurally homogeneous record that provides no information about what happens after the peak. Without having observed a complete outbreak trajectory, the LSTM has no basis for learning when or how the wave will resolve, and instead overfits to the rising phase it has seen. Mantis avoids this failure because its parameters are fixed after simulation-based pretraining on millions of complete outbreak trajectories: it does not retrain on observed COVID-19 data and therefore is not misled by the incomplete signal of a single rising wave.

This overfitting concern has shaped operational practice throughout the pandemic. Because data-driven deep learning models are known to overfit under sparse observations, many forecasting teams either did not deploy such models during the early pandemic or relied primarily on mechanistic models during this period. For example, DeepCOVID \cite{deepcovid}, one of the best-performing deep learning models in the CDC Forecast Hub, was only released toward the latter portion of this evaluation window, after sufficient training data had accumulated to make data-driven approaches more viable. More broadly, we cannot make direct comparisons against most CDC Forecast Hub model submissions during the early weeks of the pandemic, as many hub models were not yet operational or publicly available at that time. The common response to this limitation---defaulting to mechanistic compartmental models in early outbreak settings---underscores a gap that Mantis is specifically designed to fill. By grounding its forecasts in mechanistic simulations rather than observed outbreak histories, Mantis can be deployed immediately at the onset of an outbreak without the overfitting risks that constrain other data-driven methods.

\begin{figure}[H]
    \centering
    \includegraphics[width=0.85\textwidth]{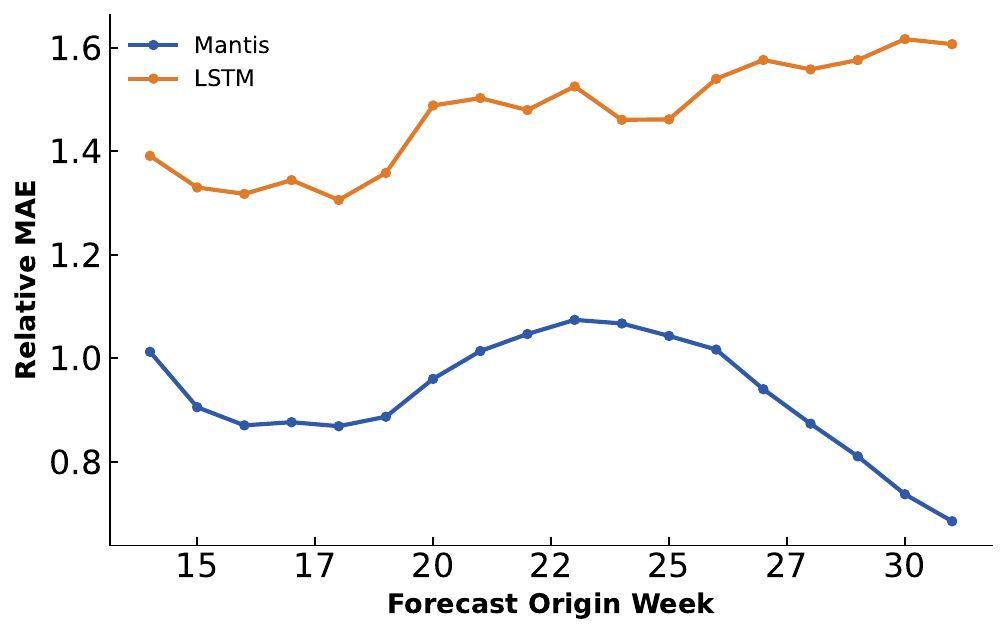}
    \caption{
        \textbf{Mantis avoids overfitting during the early pandemic while the LSTM does not.} Relative MAE (model MAE / naïve baseline MAE) for Mantis (blue) and a retrained LSTM baseline (orange) across the early COVID-19 pandemic, aggregated across U.S. states. Values below 1.0 indicate improvement over the naïve persistence baseline. Mantis's relative MAE declines as the outbreak progresses, beating the baseline on more than two-thirds of forecast weeks. The LSTM, retrained at each forecast point on all available data, overfits to limited early-pandemic observations and fails to beat the baseline throughout the evaluation period.
    }
    \label{fig:early_pandemic_mae}
\end{figure}

\subsection*{Expanded Disease Evaluation and Statistical Significance Testing}
\label{appendix:expanded}

To evaluate Mantis's generalization across a broader range of pathogens, transmission mechanisms, and geographic settings, we conducted additional forecasting evaluations on 10 diseases not included in the main text analysis. Details on each data set are given in Appendix D, but in brief, these diseases span enteric, zoonotic, parasitic, and respiratory transmission modes across multiple continents:

\begin{itemize}
    \item \textbf{Hand, Foot, and Mouth Disease (HFMD)}: enterovirus, 78 provinces in Thailand
    \item \textbf{Leptospirosis}: zoonotic/environmental, 78 provinces in Thailand
    \item \textbf{Malaria}: parasitic/vector-borne, 78 provinces in Thailand
    \item \textbf{Ebola}: zoonotic/close contact, North Kivu province, Democratic Republic of the Congo
    \item \textbf{Mpox}: zoonotic/close contact, Sierra Leone and Uganda
    \item \textbf{Cholera}: enteric/environmental, Ethiopia
    \item \textbf{Acute diarrhea}: primarily enteric, 78 provinces in Thailand
    \item \textbf{Scrub typhus}: zoonotic/vector-borne, 78 provinces in Thailand
    \item \textbf{Enteric fever} (typhoid and paratyphoid): enteric, 78 provinces in Thailand
    \item \textbf{Measles}: respiratory, 78 provinces in Thailand
\end{itemize}

For each disease, we compared Mantis against three baselines: exponential smoothing (ETS), SARIMAX, and LSTM \cite{LSTM}. As in the main text, all non-Mantis baseline models were retrained at each forecast time point to incorporate newly available data. A key feature of this evaluation is that Mantis required no tuning whatsoever across all 10 diseases and locations---the same pretrained model, code, and inference pipeline were used for every setting, from Ebola in the DRC to malaria in Thailand to cholera in Ethiopia. Generating forecasts for a new disease required only providing the input time series. Due to the substantially larger scale of these evaluations (e.g., each Thai disease involves years of weekly data across 78 provinces), we did not perform per-disease hyperparameter tuning for the non-Mantis baselines as was done for the primary evaluations---a process that would need to be repeated weekly, separately for each location and disease. Instead, we used a standard configuration for each baseline method (described in Appendix~\ref{appendix:evaluation}).

For the Thai disease evaluations, we additionally assessed the impact of incorporating weather covariates on forecasting performance. Specifically, we provided provincial-level relative humidity as a covariate input alongside the target case series for all models. For Mantis, the inclusion of relative humidity as a leading covariate improved forecasting MAE by approximately 15--20\% across most Thai diseases, consistent with the known climate sensitivity of vector-borne, enteric, and environmentally mediated pathogens in tropical settings. This result further demonstrates Mantis's ability to flexibly leverage novel covariate types at inference time: relative humidity was never represented in any training simulation, yet the model extracted useful predictive signal from it without any retraining or architectural modification. As with the spatial covariate evaluation described below, the covariate-aware variant used identical model weights and code---the only difference was the addition of the humidity time series as an input channel.

Mantis achieved the lowest MAE in 9 of 10 diseases, with ETS achieving the lowest MAE for scrub typhus (Table~\ref{tab:expanded_results}). Mantis ranked second for scrub typhus with nearly identical MAE (2.03 vs.\ 2.01). Across all 10 diseases, Mantis maintained well-calibrated uncertainty, with 90\% coverage intervals containing the true value between 83\% and 98\% of the time.

\subsubsection*{Statistical Significance Testing via Diebold-Mariano Test}

To provide formal statistical comparisons between Mantis and each baseline, we conducted two-sided Diebold-Mariano (DM) tests \cite{dmtest}. The null hypothesis is that there is no statistically significant difference in predictive accuracy between Mantis and the comparison model (in either direction). We computed the test using raw (unnormalized) mean absolute errors rather than the relative MAE reported elsewhere, because the DM test requires the raw loss differential series to satisfy its distributional assumptions (although we note Mantis and the other top performing models still reliably beat the naive baseline). We note that the DM test assumes covariance stationarity of the loss differentials, an assumption that may not be fully satisfied in epidemiological settings where forecast difficulty varies across outbreak phases. We report p-values for transparency but encourage interpreting them alongside the magnitude of MAE differences rather than relying on significance thresholds alone.

Results are presented in Table~\ref{tab:expanded_results}. Mantis's improvement over the baselines was statistically significant 38 out of 45 cases, and the improvement over ETS was statistically significant ($p < 0.05$) for 10 of 15 diseases. We report only p value vs ETS in the table to save space and because Mantis' improvement vs the other baselines was statistically significant for each disease except Hepatitis B. We excluded COVID-19 from this table because the CDC forecast hubs that we primarily compare Mantis to were not available throughout the entire test period, and many of them were created months into the pandemic, so we could not conduct a standardized test.

\begin{table*}[ht!]
\centering
\caption{Expanded disease evaluation results. For MAE and 90\% coverage, the best model is highlighted in \colorbox{lightgreen}{\textbf{green and bold}} and the second-best in \colorbox{lightblue}{\underline{blue and underlined}}. DM $p$-values report two-sided Diebold-Mariano tests comparing Mantis to each baseline (* denotes $p < 0.05$). Mantis significantly outperformed LSTM and SARIMAX ($p < 0.05$) for all 10 diseases; only ETS $p$-values are shown as they exhibit meaningful variation. Disease locations are listed in the text above.}
\label{tab:expanded_results}
\begin{tabular}{l|cccc|cccc|c}
\toprule
& \multicolumn{4}{c|}{\textbf{MAE}} & \multicolumn{4}{c|}{\textbf{95\% Coverage}} & \\
\textbf{Disease} & \rotatebox{60}{\textbf{Mantis}} & \rotatebox{60}{\textbf{ETS}} & \rotatebox{60}{\textbf{LSTM}} & \rotatebox{60}{\textbf{SARIMAX}} & \rotatebox{60}{\textbf{Mantis}} & \rotatebox{60}{\textbf{ETS}} & \rotatebox{60}{\textbf{LSTM}} & \rotatebox{60}{\textbf{SARIMAX}} & \rotatebox{60}{\textbf{DM $p$ (ETS)}} \\
\midrule
Dengue & \colorbox{lightgreen}{\textbf{249.29}} & 321.16 & 434.70 & \colorbox{lightblue}{\underline{304.94}} & \colorbox{lightblue}{\underline{0.90}} & \colorbox{lightblue}{\underline{0.90}} & 0.73 & \colorbox{lightgreen}{\textbf{0.92}} & $<$0.001* \\
Smallpox & \colorbox{lightgreen}{\textbf{21.40}} & 25.78 & 33.78 & \colorbox{lightblue}{\underline{24.49}} & \colorbox{lightblue}{\underline{0.87}} & 0.86 & 0.71 & \colorbox{lightgreen}{\textbf{0.88}} & $<$0.001* \\
Flu hosp. & \colorbox{lightgreen}{\textbf{12.21}} & 16.52 & 17.38 & \colorbox{lightblue}{\underline{15.80}} & \colorbox{lightgreen}{\textbf{0.92}} & 0.78 & \colorbox{lightblue}{\underline{0.79}} & 0.78 & 0.004* \\
Scarlet fever & \colorbox{lightgreen}{\textbf{72.31}} & 80.43 & \colorbox{lightblue}{\underline{78.63}} & 94.89 & \colorbox{lightgreen}{\textbf{0.91}} & 0.89 & 0.80 & \colorbox{lightblue}{\underline{0.90}} & 0.012* \\
Hepatitis B & \colorbox{lightblue}{\underline{7.71}} & 7.80 & \colorbox{lightgreen}{\textbf{7.44}} & 7.81 & \colorbox{lightblue}{\underline{0.90}} & 0.88 & 0.78 & \colorbox{lightgreen}{\textbf{0.91}} & 0.434 \\
HFMD & \colorbox{lightgreen}{\textbf{9.44}} & \colorbox{lightblue}{\underline{10.51}} & 14.56 & 12.26 & \colorbox{lightgreen}{\textbf{0.93}} & \colorbox{lightblue}{\underline{0.60}} & 0.42 & 0.51 & 0.027* \\
Measles & \colorbox{lightgreen}{\textbf{1.72}} & \colorbox{lightblue}{\underline{1.81}} & 4.03 & 4.12 & \colorbox{lightgreen}{\textbf{0.97}} & \colorbox{lightblue}{\underline{0.77}} & 0.17 & 0.51 & 0.270 \\
Scrub typhus & \colorbox{lightblue}{\underline{2.03}} & \colorbox{lightgreen}{\textbf{2.01}} & 4.46 & 7.31 & \colorbox{lightgreen}{\textbf{0.97}} & \colorbox{lightblue}{\underline{0.72}} & 0.17 & 0.42 & 0.705 \\
Enteric fever & \colorbox{lightgreen}{\textbf{1.60}} & \colorbox{lightblue}{\underline{1.62}} & 4.92 & 4.42 & \colorbox{lightgreen}{\textbf{0.97}} & \colorbox{lightblue}{\underline{0.80}} & 0.11 & 0.54 & 0.684 \\
Acute diarrhea & \colorbox{lightgreen}{\textbf{134.82}} & \colorbox{lightblue}{\underline{149.01}} & 207.86 & 199.37 & \colorbox{lightgreen}{\textbf{0.95}} & \colorbox{lightblue}{\underline{0.57}} & 0.49 & 0.71 & $<$0.001* \\
Ebola & \colorbox{lightgreen}{\textbf{2.74}} & \colorbox{lightblue}{\underline{3.10}} & 3.41 & 4.76 & \colorbox{lightgreen}{\textbf{0.91}} & \colorbox{lightblue}{\underline{0.85}} & 0.19 & 0.65 & $<$0.001* \\
Leptospirosis & \colorbox{lightgreen}{\textbf{1.38}} & \colorbox{lightblue}{\underline{1.51}} & 2.76 & 3.47 & \colorbox{lightgreen}{\textbf{0.98}} & \colorbox{lightblue}{\underline{0.75}} & 0.17 & 0.51 & 0.003* \\
Malaria & \colorbox{lightgreen}{\textbf{4.86}} & \colorbox{lightblue}{\underline{5.20}} & 18.76 & 17.45 & \colorbox{lightgreen}{\textbf{0.97}} & \colorbox{lightblue}{\underline{0.78}} & 0.12 & 0.57 & 0.147 \\
Cholera & \colorbox{lightgreen}{\textbf{3.96}} & \colorbox{lightblue}{\underline{4.43}} & 4.95 & 6.18 & \colorbox{lightblue}{\underline{0.82}} & \colorbox{lightgreen}{\textbf{0.84}} & 0.11 & 0.68 & 0.010* \\
Mpox & \colorbox{lightgreen}{\textbf{0.89}} & \colorbox{lightblue}{\underline{1.21}} & 1.37 & 1.27 & \colorbox{lightgreen}{\textbf{0.97}} & \colorbox{lightblue}{\underline{0.88}} & 0.03 & 0.69 & 0.105 \\
\bottomrule
\end{tabular}
\end{table*}

\subsection*{Preliminary Spatial Evaluation: Adjacent-Region Covariates}

Numerous studies have highlighted the importance of spatial transport mechanisms in shaping epidemic dynamics \cite{riley2007large, sattenspiel2009geographic, chowell2018spatial, bertuzzo2010spatially,gattospatial}. \cite{gattospatial} demonstrated that modeling inter-provincial mobility was critical for understanding COVID-19's geographic spread across Italy and quantifying the impact of spatially targeted interventions. Although Mantis was designed to forecast disease trajectories within a single location and was not trained on any spatially coupled simulation data, we conducted a preliminary evaluation to assess whether the model could extract useful signal from spatial covariates at inference time.

We evaluated Mantis on COVID-19 death forecasting across U.S. states during the early pandemic period (February 2020 through November 2021), comparing two configurations: (1) a baseline in which Mantis received only the target state's death time series with no covariate, and (2) a spatial variant in which Mantis additionally received the death time series from the target state's most populous neighboring state as a covariate input. Neighboring states were defined by shared borders, and the largest neighbor by population was selected under the assumption that it would be the most likely source of epidemic spillover. All other model settings, including architecture and weights, were identical between conditions.

\begin{figure}[ht!]
    \centering
    \includegraphics[width=0.5\textwidth]{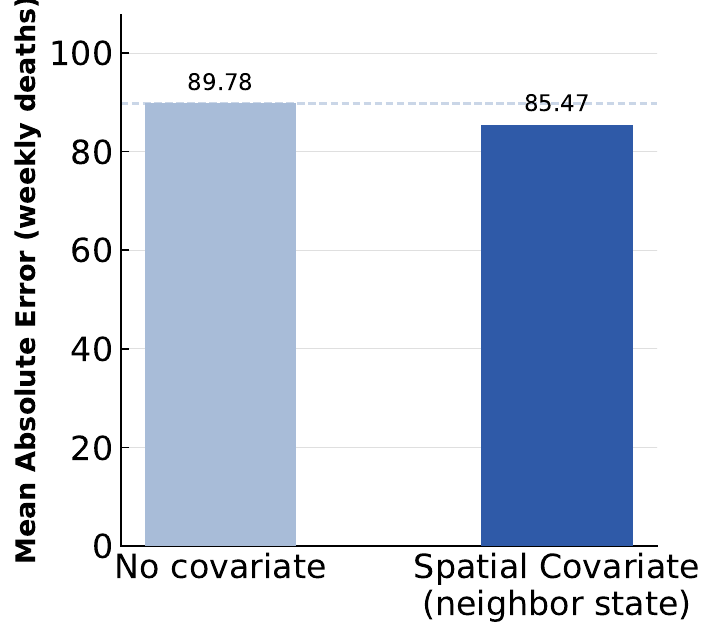}
    \caption{\textbf{Spatial covariates modestly improve Mantis forecasts despite no spatial training.} Mean absolute error for COVID-19 death forecasts across U.S. states, comparing Mantis with no covariate (baseline) versus Mantis provided with deaths from the largest neighboring state as a covariate input. The spatial variant achieved 4.8\% lower MAE across approximately 4,000 forecasts, suggesting that Mantis can extract useful spatial signal at inference time even without any spatially coupled training data.}
    \label{fig:spatial_eval}
\end{figure}

Across approximately 4,000 individual forecasts, the spatial variant achieved 4.8\% lower mean absolute error than the no-covariate baseline (Figure~\ref{fig:spatial_eval}). This improvement, while modest, is notable given that Mantis was never trained on spatially coupled data of any kind. The result suggests that the model can leverage cross-location signals when provided, even without explicit training on spatial dynamics.

We emphasize that this evaluation represents a simple proof-of-concept rather than any sort of comprehensive spatial forecasting framework. The covariate used here (deaths in a single adjacent state) captures only a crude proxy for spatial coupling, omitting mobility networks, commuting patterns, and multi-hop transmission pathways that are known to drive geographic spread \cite{gattospatial}. A model explicitly trained on spatially coupled simulations---for example, incorporating metapopulation dynamics with gravity-model or radiation-model mobility kernels \cite{masucci2013gravity}---would likely leverage spatial information far more effectively. Extending the Mantis simulation framework to include such spatial structure represents a promising direction for future work.

\subsection{Performance Consistency Across Population Scales}
\label{appendix:population}

To assess whether Mantis's performance varies systematically with jurisdiction size, we analyzed the relationship between state population and forecast accuracy for the COVID-19 mortality forecasting task. We computed the mean relative MAE for each U.S. state and territory across all forecast dates in the evaluation period (April 2020 through November 2021) and examined the correlation with 2020 Census population estimates.

\begin{figure}[H]
    \centering
    \includegraphics[width=\textwidth]{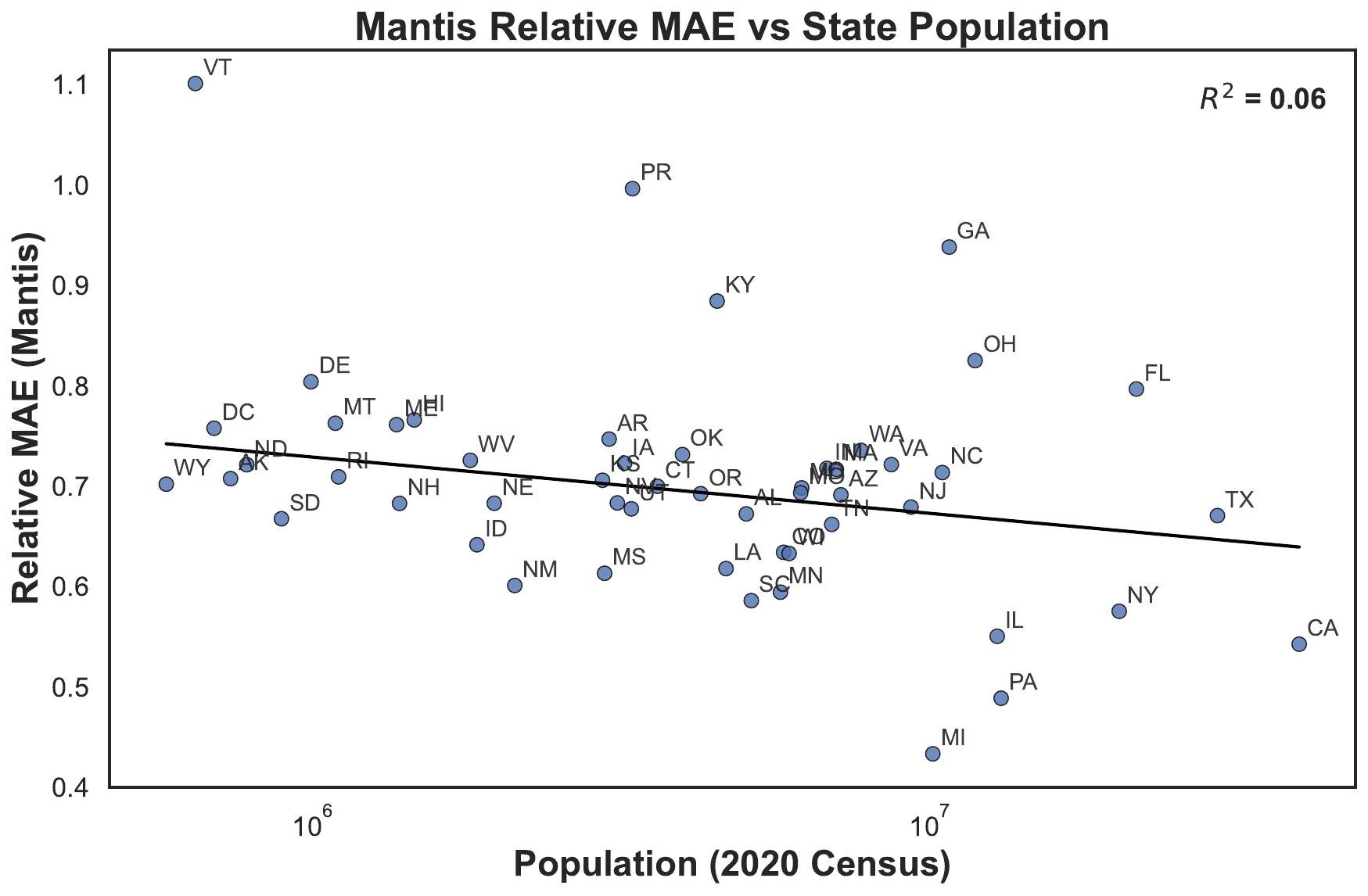}
    \caption{
        \textbf{Mantis delivers consistent performance across population scales.}
        Relative MAE versus state population for COVID-19 mortality forecasts across 51 U.S. states and territories (Vermont excluded as an outlier). Each point represents the mean relative MAE for a jurisdiction across all forecast dates from April 2020 through November 2021. Population is shown on a logarithmic scale (2020 Census). A weak negative correlation ($R^2 = 0.06$) indicates slightly better performance in larger states, but the low coefficient of determination suggests that Mantis achieves relatively uniform accuracy across diverse population sizes.
    }
    \label{fig:population_mae}
\end{figure}

Figure~\ref{fig:population_mae} shows Mantis's relative MAE versus state population on a logarithmic scale. We excluded Vermont from this analysis due to anomalous worse performance, perhaps due to data quality issues: in Vermont's COVID-19 surveillance data, hospitalizations significantly lag deaths (rather than preceding them as expected), suggesting potential reporting artifacts that confound model evaluation. Among the remaining 51 jurisdictions, we observed a weak negative correlation between log population and relative MAE ($R^2 = 0.06$), meaning Mantis performs slightly better in larger states. However, the low $R^2$ value suggests that Mantis performance is relatively consistent across diverse population scales.

This consistency is positive given that smaller jurisdictions present distinct forecasting challenges due to noisier case trajectories and sometimes less available data. The weak relationship between population and performance suggests that Mantis's simulation-based training enables robust forecasting across the spectrum of population sizes from states like Wyoming to California.

\section{Training Data}\label{appendix:simulation}

\subsection*{Human-to-Human Transmission Simulator}

The human-to-human transmission simulator implements a stochastic compartmental model that captures the complex dynamics of respiratory pathogens spreading through direct contact. This simulator generates diverse outbreak scenarios by randomly sampling from wide parameter ranges, producing synthetic epidemics that span the spectrum from influenza-like seasonal patterns to COVID-like pandemic dynamics.

\subsubsection*{Model Structure}

\paragraph{Compartmental Framework}
The core model follows a flexible SEAIR structure (Susceptible, Exposed, Asymptomatic, Infectious, Recovered) with optional latent periods, asymptomatic transmission, and waning immunity. These structural features are included probabilistically during simulation generation, allowing Mantis to learn from a wide range of outbreak regimes, including both acute and chronic infections, and both simple and complex transmission pathways. For example, in some simulations, all exposed individuals transition directly to the infectious state (i.e., no latent period); in others, asymptomatic infections are excluded, reflecting diseases without subclinical spread.

When fully enabled, the population is divided into five compartments:

\begin{align}
\frac{dS}{dt} &= -\lambda S + \omega R + b N - d S \\
\frac{dE}{dt} &= \lambda S - \sigma E - d E \\
\frac{dA}{dt} &= p_a \sigma E - \gamma_a A - d A \\
\frac{dI}{dt} &= (1-p_a) \sigma E - \gamma I - d I \\
\frac{dR}{dt} &= \gamma_a A + \gamma I - \omega R - d R
\end{align}

where:
\begin{itemize}
    \item $S$: Susceptible individuals
    \item $E$: Exposed (latent) individuals  
    \item $A$: Asymptomatic infectious individuals
    \item $I$: Symptomatic infectious individuals
    \item $R$: Recovered/immune individuals
    \item $N = S + E + A + I + R$: Total population
\end{itemize}

\paragraph{Force of Infection}
The force of infection $\lambda$ incorporates contributions from both symptomatic and asymptomatic individuals:

\begin{align}
\lambda(t) = \beta(t) \cdot s(t) \cdot \left(\frac{I + \alpha A}{N}\right) \cdot m(t)
\end{align}

where:
\begin{itemize}
    \item $\beta(t)$: Time-varying transmission rate (allows for multiple waves)
    \item $s(t)$: Seasonal forcing factor
    \item $\alpha$: Relative transmissibility of asymptomatic individuals
    \item $m(t)$: Super-spreading multiplier
\end{itemize}

\paragraph{Model Parameters}
The model includes the following epidemiological parameters:
\begin{itemize}
    \item $\sigma$: Progression rate from exposed to infectious ($1/\sigma$ = latent period)
    \item $\gamma$: Recovery rate for symptomatic individuals ($1/\gamma$ = infectious period)
    \item $\gamma_a$: Recovery rate for asymptomatic individuals
    \item $\omega$: Waning immunity rate ($1/\omega$ = duration of immunity)
    \item $p_a$: Probability of asymptomatic infection
    \item $b$: Birth rate
    \item $d$: Death rate
\end{itemize}

\subsubsection*{Stochastic Implementation}

The continuous-time model is implemented as a discrete-time stochastic process using daily time steps. At each time step $t$, transitions between compartments are drawn from binomial distributions (equivalent to tau-leaping with large tau, but we found that it improved performance):

\begin{align}
\text{New exposures} &\sim \text{Binomial}(S_t, 1-e^{-\lambda_t}) \\
\text{E} \to \text{I/A transitions} &\sim \text{Binomial}(E_t, 1-e^{-\sigma}) \\
\text{Recoveries from I} &\sim \text{Binomial}(I_t, 1-e^{-\gamma}) \\
\text{Recoveries from A} &\sim \text{Binomial}(A_t, 1-e^{-\gamma_a}) \\
\text{Waning immunity} &\sim \text{Binomial}(R_t, 1-e^{-\omega})
\end{align}

Among newly infectious individuals, the split between asymptomatic and symptomatic follows:
\begin{align}
\text{New asymptomatic} &\sim \text{Binomial}(\text{New infectious}, p_a)
\end{align}

\subsubsection*{Multi-Wave Dynamics}

To capture realistic outbreak patterns with multiple waves, the transmission rate $\beta(t)$ is implemented as a piecewise constant function:

\begin{align}
\beta(t) = \beta_i \quad \text{for } t \in [t_{i-1}, t_i)
\end{align}

where $t_0, t_1, \ldots, t_k$ are randomly chosen wave change points and $\beta_1, \beta_2, \ldots, \beta_k$ are independently sampled transmission rates. The number of wave changes is drawn from a discrete uniform distribution, and change points are sampled uniformly within the simulation period.

\subsubsection*{Seasonal Forcing}

Seasonal transmission patterns are modeled using multiple harmonic components:

\begin{align}
s(t) = 1 + \sum_{j=1}^{n_h} A_j \cos\left(\frac{2\pi(t - \phi_j + \delta_y)}{P_j}\right) + \epsilon_t
\end{align}

where:
\begin{itemize}
    \item $n_h$: Number of harmonic components (1-4)
    \item $A_j$: Amplitude of harmonic $j$
    \item $P_j$: Period of harmonic $j$ (365.0, 182.5, or 91.25 days)
    \item $\phi_j$: Phase offset for harmonic $j$
    \item $\delta_y$: Annual peak jitter, sampled independently each year to slightly shift seasonal peaks and avoid perfectly periodic recurrence
    \item $\epsilon_t$: Daily multiplicative noise $\sim \mathcal{N}(1, 0.05^2)$
\end{itemize}

\subsubsection*{Super-Spreading Events}

Super-spreading is modeled through a stochastic multiplier applied to the force of infection. At each time step, a subset of infectious individuals becomes super-spreaders with probability $p_{ss}$. The transmission enhancement follows a Gamma distribution:

\begin{align}
m(t) = \left(1 - \frac{n_{ss}}{n_I}\right) + \frac{n_{ss}}{n_I} \cdot M
\end{align}

where $n_{ss} \sim \text{Binomial}(n_I, p_{ss})$ and $M \sim \text{Gamma}(k_{ss}, \theta_{ss})$.

\subsubsection*{Intervention Policies}

The simulator includes an optional universal intervention policy that modulates transmission based on case thresholds. The policy operates through a contact reduction factor applied to $\beta(t)$:

\begin{align}
\beta_{eff}(t) = \beta(t) \cdot r_{contact}(t)
\end{align}

where $r_{contact}(t) \in [0.2, 1.0]$ represents the intervention strength. The policy activates when daily cases exceed an on-threshold and deactivates after cases remain below an off-threshold for a specified duration.

\subsubsection*{Endemic Dynamics}

For simulations with endemic circulation, demographic processes are included:

\begin{align}
\text{Births} &\sim \text{Poisson}(b \cdot N) \\
\text{Deaths} &\sim \text{Binomial}(X, \min(d, 1)) \quad \forall X \in \{S,E,A,I,R\} \\
\text{Importations} &\sim \text{Poisson}(\iota)
\end{align}

New births enter the susceptible compartment, deaths occur proportionally across all compartments, and imported cases enter the exposed or infectious compartment.

\subsubsection*{Observation Model}

True epidemic dynamics are converted to observed case counts through a comprehensive noise injection pipeline that simulates real-world surveillance artifacts:

\paragraph{Multiplicative Noise}
Base case counts are perturbed with log-normal noise:
\begin{align}
\hat{C}_t = C_t \cdot \epsilon_t, \quad \epsilon_t \sim \text{LogNormal}(0, \sigma_{mult}^2)
\end{align}

\paragraph{Underreporting}
Reporting rates evolve logistically over time:
\begin{align}
r_t = r_0 + (r_\infty - r_0) \cdot \frac{1}{1 + e^{-k(t - t_{mid})}}
\end{align}

Reported cases follow: $\tilde{C}_t \sim \text{Binomial}(\hat{C}_t, r_t)$

\paragraph{Reporting Delays}
Cases are distributed across future days using inverse-weighted delay distributions:
\begin{align}
P(\text{delay} = d) \propto \frac{\alpha}{d + 1}, \quad d \in [0, D_{max}]
\end{align}

\paragraph{Day-of-Week Effects}
Reporting varies by weekday with multiplicative factors:
\begin{align}
\tilde{C}_t' = \tilde{C}_t \cdot w(\text{day-of-week}(t))
\end{align}

where $w(\cdot)$ ranges from 0.4 (weekend) to 1.2 (Tuesday).

\subsubsection*{Hospitalization and Death Generation}

Secondary outcomes (hospitalizations and deaths) are derived from symptomatic cases using time-varying probabilities and delay distributions. For each wave segment $i$, hospitalization and death probabilities ($p_{h,i}$ and $p_{d,i}$) are independently sampled.

\paragraph{Hospitalization Process}
\begin{align}
H_t &= \sum_{\tau=0}^{t} \text{Binomial}(I_\tau, p_{h,wave(\tau)}) \cdot f_h(t-\tau) \\
f_h(d) &\sim \text{Gamma}(k_h, \theta_h)
\end{align}

\paragraph{Death Process}
Deaths occur among hospitalized individuals with additional delays:
\begin{align}
D_t &= \sum_{\tau=0}^{t} \sum_{s=0}^{\tau} \text{Binomial}(H_{\tau,s}, p_{d,wave(s)}) \cdot f_d(t-\tau) \\
f_d(d) &\sim \text{Gamma}(k_d, \theta_d)
\end{align}

where $H_{\tau,s}$ represents hospitalizations on day $\tau$ originating from infections on day $s$.

\subsubsection*{Parameter Ranges}

Table~\ref{tab:h2h_params} summarizes all parameters and their sampling distributions used in the human-to-human transmission simulator. Parameters were chosen to be broadly consistent with respirator pathogens.

\begin{longtable}{|p{3.5cm}|p{2.5cm}|p{6cm}|p{2cm}|}

\caption{Human-to-Human Transmission Simulator Parameters}
\label{tab:h2h_params} \\
\hline
\textbf{Parameter} & \textbf{Symbol} & \textbf{Range/Distribution} & \textbf{Units} \\
\hline
\endfirsthead 

\multicolumn{4}{c}%
{{\bfseries \tablename\ \thetable{} -- continued from previous page}} \\
\hline
\textbf{Parameter} & \textbf{Symbol} & \textbf{Range/Distribution} & \textbf{Units} \\
\hline
\endhead 

\hline \multicolumn{4}{r}{{Continued on next page}} \\ \hline
\endfoot

\hline
\endlastfoot

\multicolumn{4}{|c|}{\textbf{Population and Simulation}} \\
\hline
Population size & $N$ & Log-uniform[50,000, 40,000,000] \cite{zhou_epidemic_2021,yerlanov_effective_2023,mogelmose_incorporating_2022} & individuals \\
Simulation days & $T$ & 2,000 \cite{zhou_epidemic_2021,getz_adequacy_2019} & days \\
\hline

\multicolumn{4}{|c|}{\textbf{Core Transmission Parameters}} \\
\hline
Base transmission rate & $\beta$ & Uniform[0.2, 0.235] \cite{hussein_meta-analysis_2021,guerra_basic_2017} & day$^{-1}$ \\
Recovery rate & $\gamma$ & Uniform[0.1, 0.33] \cite{carrat_time_2008,stone_risk_2025} & day$^{-1}$ \\
Latent progression rate & $\sigma$ & Uniform[0.2, 0.4] \cite{lessler_incubation_2009,lauer_incubation_2020,wu_incubation_2022} & day$^{-1}$ \\
Waning immunity rate & $\omega$ & Uniform[0.001, 0.0075] \cite{berec_protection_2022,goldberg_protection_2022,ferdinands_intraseason_2017} & day$^{-1}$ \\
\hline

\multicolumn{4}{|c|}{\textbf{Asymptomatic Dynamics}} \\
\hline
Asymptomatic probability & $p_a$ & Beta(3, 7) \cite{oran_proportion_2021,alene_magnitude_2021} & proportion \\
Asymp. transmissibility & $\alpha$ & Beta(2, 5) \cite{wu_assessing_2021,buitrago-garcia_occurrence_2022} & proportion \\
Has asymptomatic compartment & - & Bernoulli(0.5) \cite{wu_assessing_2021} & boolean \\
\hline

\multicolumn{4}{|c|}{\textbf{Model Structure Options}} \\
\hline
Has latent period & - & Bernoulli(0.7) \cite{getz_adequacy_2019} & boolean \\
Has waning immunity & - & Bernoulli(0.9) \cite{berec_protection_2022,goldberg_protection_2022} & boolean \\
\hline

\multicolumn{4}{|c|}{\textbf{Multi-Wave Dynamics}} \\
\hline
Multiple waves enabled & - & Configurable (default: True) \cite{an_seasonality_2024} & boolean \\
Number of wave changes & - & Uniform[0, 4] \cite{an_seasonality_2024} & count \\
Wave change days & - & Uniform[50, 1800] \cite{an_seasonality_2024} & days \\
\hline

\multicolumn{4}{|c|}{\textbf{Super-Spreading}} \\
\hline
Super-spreader probability & $p_{ss}$ & Uniform[0.0005, 0.02] \cite{adam_clustering_2020,wang_superspreading_2021} & proportion \\
Multiplier shape & $k_{ss}$ & 4.0 \cite{wang_superspreading_2021} & - \\
Multiplier scale & $\theta_{ss}$ & 1.5 \cite{goyal_viral_2021} & - \\
\hline

\multicolumn{4}{|c|}{\textbf{Seasonality}} \\
\hline
Use seasonality & - & Bernoulli(0.8) \cite{an_seasonality_2024,price_association_2019} & boolean \\
Number of harmonics & $n_h$ & Uniform[1, 4] \cite{an_seasonality_2024} & count \\
Harmonic amplitude & $A_j$ & Base $\times$ Uniform[0.3, 1.0] \cite{an_seasonality_2024} & proportion \\
Base amplitude & - & Uniform[0.1, 0.5] \cite{an_seasonality_2024} & proportion \\
Phase offset & $\phi_j$ & Uniform[0, 365] + Uniform[-60, 60] \cite{an_seasonality_2024} & days \\
Possible periods & $P_j$ & Choice[365.0, 182.5, 91.25] \cite{an_seasonality_2024} & days \\
Daily noise std & $\sigma_{\epsilon}$ & 0.05 \cite{an_seasonality_2024} & proportion \\
Annual peak jitter & $\delta_y$ & Uniform[-30, 30] \cite{an_seasonality_2024} & days \\
\hline

\multicolumn{4}{|c|}{\textbf{Endemic Demographics}} \\
\hline
Enable endemic & - & Bernoulli(0.8) \cite{mogelmose_incorporating_2022} & boolean \\
Birth rate & $b$ & Uniform[0.00002, 0.00012] \cite{mogelmose_incorporating_2022} & day$^{-1}$ \\
Death rate factor & - & Uniform[0.8, 1.3] \cite{mogelmose_incorporating_2022} & proportion \\
Importation rate & $\iota$ & Exp(Uniform[ln(0.01), ln(0.5)]) \cite{kucharski_effectiveness_2020} & day$^{-1}$ \\
\hline

\multicolumn{4}{|c|}{\textbf{Intervention Policy}} \\
\hline
Enable intervention & - & Bernoulli(0.25) \cite{kucharski_effectiveness_2020} & boolean \\
On threshold & - & $N \times$ Uniform[1e-5, 1e-3] & cases/day \\
Off threshold & - & On threshold $\times$ Uniform[0, 1] & cases/day \\
Contact reduction & $r_{contact}$ & Uniform[0.2, 0.6] \cite{kucharski_effectiveness_2020} & proportion \\
Trigger delay & - & Uniform[0, 21] \cite{kucharski_effectiveness_2020} & days \\
Minimum duration & - & Uniform[14, 35] \cite{kucharski_effectiveness_2020} & days \\
Maximum duration & - & Choice[None, Uniform[60, 120]] \cite{kucharski_effectiveness_2020} & days \\
Consecutive off days & - & Uniform[1, 50] \cite{kucharski_effectiveness_2020} & days \\
\hline

\multicolumn{4}{|c|}{\textbf{Observation Noise}} \\
\hline
Multiplicative noise std & $\sigma_{mult}$ & 0.1 \cite{dixon_completeness_2017} & log-scale \\
Overdispersion parameter & $r$ & 1200 \cite{dixon_completeness_2017} & - \\
\hline

\multicolumn{4}{|c|}{\textbf{Reporting Effects}} \\
\hline
Apply reporting rate & - & Bernoulli(0.8) \cite{swaan_timeliness_2018,dixon_completeness_2017} & boolean \\
Initial reporting rate & $r_0$ & Uniform[0.05, 0.4] \cite{dixon_completeness_2017} & proportion \\
Final reporting rate & $r_\infty$ & Uniform[0.25, 0.85] \cite{dixon_completeness_2017} & proportion \\
Days to max testing & - & Uniform[30, 365] \cite{swaan_timeliness_2018} & days \\
Logistic steepness & $k$ & Uniform[4, 8] \cite{swaan_timeliness_2018} & - \\
\hline

Apply weekday effects & - & Bernoulli(0.8) \cite{dixon_completeness_2017} & boolean \\
Monday factor & - & Normal(1.2, 0.15) \cite{dixon_completeness_2017} & proportion \\
Tue-Thu factor & - & Normal(1.0, 0.1) \cite{dixon_completeness_2017} & proportion \\
Friday factor & - & Normal(0.9, 0.12) \cite{dixon_completeness_2017} & proportion \\
Saturday factor & - & Normal(0.6, 0.2) \cite{dixon_completeness_2017} & proportion \\
Sunday factor & - & Normal(0.4, 0.2) \cite{dixon_completeness_2017} & proportion \\
\hline

Apply lab noise & - & Bernoulli(0.8) \cite{effler_statewide_1999} & boolean \\
Lab batch size & - & Poisson(100) \cite{effler_statewide_1999} & samples \\
Bad batch rate & - & 0.005 \cite{effler_statewide_1999} & proportion \\
Bad batch accuracy & - & Uniform[0.7, 0.85] \cite{effler_statewide_1999} & proportion \\
\hline

Apply delays & - & Bernoulli(0.8) \cite{swaan_timeliness_2018} & boolean \\
Initial max delay & - & Uniform[7, 21] \cite{swaan_timeliness_2018} & days \\
Final max delay & - & Uniform[2, 7] \cite{swaan_timeliness_2018} & days \\
Initial alpha & $\alpha_0$ & 1.0 & - \\
Final alpha & $\alpha_\infty$ & 4.0 & - \\
\hline

\multicolumn{4}{|c|}{\textbf{Hospitalization \& Death}} \\
\hline
Hospitalization probability & $p_{h,i}$ & Uniform[0.02, 0.15] \cite{bajema_severity_2025} & proportion \\
Death probability (of hosp.) & $p_{d,i}$ & Uniform[0.05, 0.3] \cite{bajema_severity_2025} & proportion \\
Hosp. delay shape & $k_h$ & Uniform[2.0, 4.0] & - \\
Hosp. delay scale & $\theta_h$ & Uniform[1.0, 3.0] & days \\
Death delay shape & $k_d$ & Uniform[1.5, 2.5] & - \\
Death delay scale & $\theta_d$ & Uniform[4.5, 8.5] & days \\

\end{longtable}

This large, high dimensional parameter space enables the generation of diverse outbreak scenarios spanning different pathogen characteristics, population contexts, surveillance qualities, and intervention regimes. Each simulation run samples independently from these distributions, creating a rich training corpus that captures the full spectrum of human-to-human transmission dynamics.

\subsection*{Vector-Borne Transmission Simulator}

The vector-borne transmission simulator implements a coupled host-vector compartmental model that captures the complex dynamics of diseases transmitted through arthropod vectors, such as dengue, Zika, chikungunya, and malaria. This simulator models the epidemiological dynamics within both human and vector populations, incorporating realistic vector ecology, seasonal patterns, and intervention strategies.

\subsubsection*{Model Structure}

\paragraph{Coupled Host-Vector Framework}
The model consists of two interacting population systems: humans and vectors (typically mosquitoes). The human population follows a SEAIR structure similar to the human-to-human model, while vectors follow a simpler SEI structure without recovery (as infected vectors typically remain infectious for life).

\paragraph{Human Population Dynamics}
The human compartments evolve according to:

\begin{align}
\frac{dS_h}{dt} &= -\lambda_h S_h + \omega R_h + b_h N_h - d_h S_h \\
\frac{dE_h}{dt} &= \lambda_h S_h - \sigma_h E_h - d_h E_h \\
\frac{dA_h}{dt} &= p_a \sigma_h E_h - \gamma_a A_h - d_h A_h \\
\frac{dI_h}{dt} &= (1-p_a) \sigma_h E_h - \gamma I_h - d_h I_h \\
\frac{dR_h}{dt} &= \gamma_a A_h + \gamma I_h - \omega R_h - d_h R_h
\end{align}

\paragraph{Vector Population Dynamics}
The vector compartments evolve according to:

\begin{align}
\frac{dS_v}{dt} &= \mu_v N_v - \lambda_v S_v - \mu_v S_v \\
\frac{dE_v}{dt} &= \lambda_v S_v - \sigma_v E_v - \mu_v E_v \\
\frac{dI_v}{dt} &= \sigma_v E_v - \mu_v I_v
\end{align}

where $N_v = S_v + E_v + I_v$ is the total vector population.

\paragraph{Force of Infection}
The transmission between humans and vectors occurs through biting interactions. The force of infection for humans (from infected vectors) is:

\begin{align}
\lambda_h(t) = a \cdot b_h \cdot \frac{I_v}{N_v} \cdot \beta(t) \cdot s(t) \cdot f_{intervention}(t)
\end{align}

The force of infection for vectors (from infected humans) is:

\begin{align}
\lambda_v(t) = a \cdot b_v \cdot \frac{I_h + \alpha A_h}{N_h} \cdot \beta(t) \cdot s(t) \cdot f_{intervention}(t)
\end{align}

where:
\begin{itemize}
    \item $a$: Biting rate (bites per vector per day)
    \item $b_h$: Probability of transmission from vector to human per bite
    \item $b_v$: Probability of transmission from human to vector per bite
    \item $\beta(t)$: Time-varying transmission scaling factor
    \item $s(t)$: Seasonal forcing function
    \item $f_{intervention}(t)$: Intervention effectiveness factor
    \item $\alpha$: Relative transmissibility of asymptomatic humans
\end{itemize}

\subsubsection*{Stochastic Implementation}

The continuous-time model is implemented as a discrete-time stochastic process. At each time step $t$, transitions are drawn from binomial distributions:

\paragraph{Human Transitions}
\begin{align}
\text{New human exposures} &\sim \text{Binomial}(S_h(t), 1-e^{-\lambda_h(t)}) \\
\text{E}_h \to \text{I}_h/\text{A}_h &\sim \text{Binomial}(E_h(t), 1-e^{-\sigma_h}) \\
\text{Human recoveries from I} &\sim \text{Binomial}(I_h(t), 1-e^{-\gamma}) \\
\text{Human recoveries from A} &\sim \text{Binomial}(A_h(t), 1-e^{-\gamma_a}) \\
\text{Waning immunity} &\sim \text{Binomial}(R_h(t), 1-e^{-\omega})
\end{align}

\paragraph{Vector Transitions}
\begin{align}
\text{New vector exposures} &\sim \text{Binomial}(S_v(t), 1-e^{-\lambda_v(t)}) \\
\text{E}_v \to \text{I}_v &\sim \text{Binomial}(E_v(t), 1-e^{-\sigma_v}) \\
\text{Vector deaths from S} &\sim \text{Binomial}(S_v(t), \mu_v) \\
\text{Vector deaths from E} &\sim \text{Binomial}(E_v(t), \mu_v) \\
\text{Vector deaths from I} &\sim \text{Binomial}(I_v(t), \mu_v) \\
\text{New vector births} &\sim \text{Poisson}(\mu_v N_v(t))
\end{align}

\subsubsection*{Vector Population Dynamics}

Vector populations exhibit distinct ecological patterns compared to human populations:

\paragraph{Mortality and Reproduction}
Vectors have high mortality rates ($\mu_v$) with typical lifespans of 10-30 days. To maintain population equilibrium, new vectors are born at rate $\mu_v N_v$, entering the susceptible class.

\paragraph{Seasonal Patterns}
Vector abundance and activity are strongly seasonal, captured through enhanced seasonal forcing:
\begin{align}
s(t) = 1.2 + \sum_{j=1}^{n_h} A_j \cos\left(\frac{2\pi(t - \phi_j + \delta_y)}{P_j}\right) + \epsilon_t
\end{align}

The baseline value (1.2) and amplitude ranges ($A_j \in [0.5, 1.25]$) are increased compared to human-to-human transmission to reflect the stronger seasonal dependence of vector-borne diseases.

\subsubsection*{Multi-Wave Dynamics}

Similar to the human-to-human simulator, vector-borne outbreaks can exhibit multiple waves through time-varying transmission scaling:

\begin{align}
\beta(t) = \beta_i \quad \text{for } t \in [t_{i-1}, t_i)
\end{align}

However, for vector-borne diseases, wave changes often reflect environmental conditions affecting vector populations (rainfall, temperature) or intervention campaigns rather than behavioral changes.

\subsubsection*{Intervention Modeling}

Vector control interventions are modeled through the reduction factor $f_{intervention}(t)$ applied to both $\lambda_h$ and $\lambda_v$. Common interventions include:

\begin{itemize}
    \item \textbf{Vector control}: Reduces vector population or biting rates
    \item \textbf{Larvicide applications}: Affects vector reproduction
    \item \textbf{Personal protection}: Reduces effective contact rates
    \item \textbf{Environmental management}: Modifies vector habitat suitability
\end{itemize}

The intervention policy triggers based on human case thresholds but affects the vector reduction parameter:
\begin{align}
f_{intervention}(t) = \begin{cases}
f_{vector} & \text{if intervention active} \\
1.0 & \text{otherwise}
\end{cases}
\end{align}

where $f_{vector} \in [0.2, 0.6]$ represents the effectiveness of vector control measures.

\subsubsection*{Initialization Strategy}

\paragraph{Human Population}
Initial conditions allow for pre-existing immunity in the human population:
\begin{align}
\text{Immune fraction} &\sim \text{Uniform}(0, 0.7) \\
\text{Initial infections} &\sim \text{NegativeBinomial}(n=\max(1, N_h \times 5 \times 10^{-6}), p=0.5)
\end{align}

\paragraph{Vector Population}
Vector populations start with a small fraction exposed/infected:
\begin{align}
N_v &= N_h \times \text{Uniform}(2.0, 10.0) \\
\text{Initial vector exposures} &\sim \text{NegativeBinomial}(n=\max(1, N_v \times 10^{-4}), p=0.3)
\end{align}

The higher initial prevalence in vectors reflects their role as reservoirs and their shorter generation times.

\subsubsection*{Parameter Ranges}

Table~\ref{tab:vector_params} summarizes all parameters specific to the vector-borne transmission simulator.

\begin{longtable}{|p{3.5cm}|p{2.5cm}|p{6cm}|p{2cm}|}

\caption{Vector-Borne Transmission Simulator Parameters}
\label{tab:vector_params} \\
\hline
\textbf{Parameter} & \textbf{Symbol} & \textbf{Range/Distribution} & \textbf{Units} \\
\hline
\endfirsthead 

\multicolumn{4}{c}%
{{\bfseries \tablename\ \thetable{} -- continued from previous page}} \\
\hline
\textbf{Parameter} & \textbf{Symbol} & \textbf{Range/Distribution} & \textbf{Units} \\
\hline
\endhead

\hline \multicolumn{4}{r}{{Continued on next page}} \\ \hline
\endfoot 

\hline
\endlastfoot 

\multicolumn{4}{|c|}{\textbf{Human Population (Shared with H2H)}} \\
\hline
Population size & $N_h$ & Log-uniform[50,000, 40,000,000] \cite{zhou_epidemic_2021,yerlanov_effective_2023,mogelmose_incorporating_2022} & individuals \\
Base transmission scaling & $\beta$ & Uniform[0.5, 0.7] \cite{longini_statistical_1988} & - \\
Recovery rate & $\gamma$ & Uniform[0.1, 0.33] \cite{cdc_dengue_2025,van_boven_estimating_2020} & day$^{-1}$ \\
Latent progression rate & $\sigma_h$ & Uniform[0.2, 0.4] \cite{chan_incubation_2012,paz-bailey_dengue_2024} & day$^{-1}$ \\
Waning immunity rate & $\omega$ & Uniform[0.003, 0.02] \cite{katzelnick_dengue_2021} & day$^{-1}$ \\
Asymptomatic probability & $p_a$ & Beta(3, 7) \cite{asish_global_2023} & proportion \\
Asymp. transmissibility & $\alpha$ & Beta(2, 5) \cite{duong_asymptomatic_2015} & proportion \\
\hline

\multicolumn{4}{|c|}{\textbf{Vector Population Dynamics}} \\
\hline
Vector population size & $N_v$ & $N_h \times$ Uniform[2.0, 10.0] \cite{smith_risk_2004,amaku_maximum_2013} & individuals \\
Vector mortality rate & $\mu_v$ & Uniform[0.03, 0.10] \cite{matthews_mosquito_2025,goindin_parity_2015} & day$^{-1}$ \\
Vector latent progression & $\sigma_v$ & Uniform[0.15, 0.30] \cite{loisel_intra-vector_2025} & day$^{-1}$ \\
Vector lifespan & $1/\mu_v$ & Uniform[10, 33] \cite{matthews_mosquito_2025,doeurk_impact_2025} & days \\
Extrinsic incubation & $1/\sigma_v$ & Uniform[3.3, 6.7] \cite{chan_incubation_2012,loisel_intra-vector_2025} & days \\
\hline

\multicolumn{4}{|c|}{\textbf{Transmission Parameters}} \\
\hline
Biting rate & $a$ & Uniform[0.40, 0.80] \cite{zahid_biting_2023,christofferson_how_2022} & bites/vector/day \\
Vector→human transmission & $b_h$ & Uniform[0.35, 0.75] \cite{nguyet_host_2013,lambrechts_direct_2023} & probability \\
Human→vector transmission & $b_v$ & Uniform[0.35, 0.75] \cite{nguyet_host_2013,lambrechts_direct_2023} & probability \\
\hline

\multicolumn{4}{|c|}{\textbf{Model Structure Options}} \\
\hline
Has latent period & - & Bernoulli(0.7) \cite{zhou_epidemic_2021} & boolean \\
Has asymptomatic compartment & - & Bernoulli(0.5) \cite{duong_asymptomatic_2015} & boolean \\
Has waning immunity & - & Bernoulli(1.0) \cite{katzelnick_dengue_2021} & boolean \\
\hline

\multicolumn{4}{|c|}{\textbf{Enhanced Seasonality}} \\
\hline
Use seasonality & - & Bernoulli(0.8) \cite{lowe_dengue_2014,franklinos_effect_2019} & boolean \\
Base seasonal value & - & 1.2 \cite{lowe_dengue_2014} & - \\
Harmonic amplitude range & $A_j$ & Base $\times$ Uniform[0.3, 1.0] \cite{lowe_dengue_2014} & proportion \\
Base amplitude & - & Uniform[0.5, 1.25] \cite{lowe_dengue_2014} & proportion \\
Number of harmonics & $n_h$ & Uniform[1, 4] \cite{lowe_dengue_2014} & count \\
\hline

\multicolumn{4}{|c|}{\textbf{Multi-Wave Dynamics}} \\
\hline
Multiple waves enabled & - & Configurable (default: True) \cite{franklinos_effect_2019} & boolean \\
Number of wave changes & - & Uniform[0, 4] \cite{franklinos_effect_2019} & count \\
Wave change timing & - & Uniform[50, 1800] \cite{franklinos_effect_2019} & days \\
\hline

\multicolumn{4}{|c|}{\textbf{Vector Control Interventions}} \\
\hline
Enable intervention & - & Bernoulli(0.25) \cite{pryce_insecticide-treated_2018} & boolean \\
On threshold & - & $N_h \times$ Uniform[1e-5, 1e-3] & cases/day \\
Off threshold & - & On threshold $\times$ Uniform[0, 1] & cases/day \\
Vector reduction & $f_{vector}$ & Uniform[0.2, 0.6] \cite{pryce_insecticide-treated_2018} & proportion \\
Trigger delay & - & Uniform[0, 21] & days \\
Minimum duration & - & Uniform[14, 35] \cite{pryce_insecticide-treated_2018} & days \\
Maximum duration & - & Choice[None, Uniform[60, 120]] \cite{pryce_insecticide-treated_2018} & days \\
Consecutive off days & - & Uniform[1, 50] & days \\
\hline

\multicolumn{4}{|c|}{\textbf{Initialization Parameters}} \\
\hline
Human immune fraction & - & Uniform[0.0, 0.7] \cite{katzelnick_dengue_2021} & proportion \\
Initial human infections & - & NegBin$(n=\max(1, N_h \times 5e^{-6}), p=0.5)$ & count \\
Vector:human ratio & - & Uniform[2.0, 10.0] \cite{smith_risk_2004} & ratio \\
Initial vector exposures & - & NegBin$(n=\max(1, N_v \times 1e^{-4}), p=0.3)$ & count \\
\hline

\multicolumn{4}{|c|}{\textbf{Observation \& Outcomes (Shared)}} \\
\hline
Hospitalization probability & $p_h$ & Uniform[0.05, 0.20] \cite{bajema_severity_2025, kharwadkar_clinical_2024} & proportion \\
Death probability (of hosp.) & $p_d$ & Uniform[0.10, 0.40] \cite{kharwadkar_clinical_2024} & proportion \\
Hosp. delay shape & $k_h$ & Uniform[2.0, 4.0] & - \\
Hosp. delay scale & $\theta_h$ & Uniform[1.0, 3.0] & days \\
Death delay shape & $k_d$ & Uniform[1.5, 2.5] & - \\
Death delay scale & $\theta_d$ & Uniform[2.0, 5.0] & days \\
\hline

\multicolumn{4}{|c|}{\textbf{Noise \& Reporting (Identical to H2H)}} \\
\hline
Multiplicative noise std & $\sigma_{mult}$ & 0.1 \cite{dixon_completeness_2017} & log-scale \\
Reporting effects & - & Same as Table~\ref{tab:h2h_params} \cite{swaan_timeliness_2018} & various \\
Day-of-week effects & - & Same as Table~\ref{tab:h2h_params} \cite{dixon_completeness_2017} & various \\
Lab noise parameters & - & Same as Table~\ref{tab:h2h_params} \cite{effler_statewide_1999} & various \\
Delay distributions & - & Same as Table~\ref{tab:h2h_params} \cite{swaan_timeliness_2018} & various \\

\end{longtable}

\subsubsection*{Key Differences from Human-to-Human Transmission}

The vector-borne simulator incorporates several critical differences that distinguish it from direct human-to-human transmission:

\paragraph{Dual Population Dynamics}
Unlike human-to-human models, vector-borne diseases require explicit modeling of vector population dynamics, including birth, death, and infection processes within the arthropod population.

\paragraph{Transmission Pathway}
Transmission requires a two-step process (human → vector → human) rather than direct contact, fundamentally altering the epidemic dynamics and introducing additional delay and complexity.

\paragraph{Environmental Sensitivity}
Vector-borne diseases exhibit stronger seasonal patterns due to vector ecology dependence on temperature, rainfall, and other environmental factors, reflected in enhanced seasonal forcing parameters.

\paragraph{Vector-Specific Interventions}
Control strategies focus on vector population reduction rather than behavioral modifications, affecting both $\lambda_h$ and $\lambda_v$ through environmental rather than social mechanisms.

\paragraph{Basic Reproduction Number}
$R_0$ calculations must account for the round-trip transmission cycle and vector population dynamics, resulting in a more complex formulation that depends on vector:human ratios and vector survival probabilities.

This comprehensive parameter space enables realistic simulation of diverse vector-borne disease scenarios, capturing the full spectrum of arthropod-transmitted pathogen dynamics across different ecological contexts, intervention strategies, and surveillance systems.

\subsection*{Environmental Transmission Simulator}

The environmental transmission simulator implements a hybrid compartmental model that captures diseases transmitted through both direct human-to-human contact and indirect environmental pathways via contaminated water sources. This simulator models pathogens such as cholera, typhoid, hepatitis A, and norovirus, which exhibit complex transmission dynamics involving multiple routes of exposure and environmental persistence.

\subsubsection*{Model Structure}

\paragraph{Dual-Route Transmission Framework}
The waterborne model extends the basic SEAIR structure to include an additional environmental compartment representing pathogen concentration in water. The model simultaneously tracks both direct contact transmission (similar to respiratory diseases) and indirect transmission through contaminated water or environmental sources.

\paragraph{Human Population Dynamics}
The human compartments follow the same structure as the human-to-human model:

\begin{align}
\frac{dS}{dt} &= -\lambda_h S + \omega R + b N - d S \\
\frac{dE}{dt} &= \lambda_h S - \sigma E - d E \\
\frac{dA}{dt} &= p_a \sigma E - \gamma_a A - d A \\
\frac{dI}{dt} &= (1-p_a) \sigma E - \gamma I - d I \\
\frac{dR}{dt} &= \gamma_a A + \gamma I - \omega R - d R
\end{align}

\paragraph{Environmental Pathogen Dynamics}
The environmental water compartment evolves according to:

\begin{align}
\frac{dW}{dt} &= \eta_I I + \eta_A A - \mu_w W
\end{align}

where:
\begin{itemize}
    \item $W$: Pathogen concentration in water sources
    \item $\eta_I$: Contamination rate from symptomatic individuals
    \item $\eta_A$: Contamination rate from asymptomatic individuals  
    \item $\mu_w$: Pathogen decay/removal rate in water
\end{itemize}

\paragraph{Composite Force of Infection}
The total force of infection combines both transmission routes:

\begin{align}
\lambda_h(t) = \lambda_{contact}(t) + \lambda_{water}(t)
\end{align}

where:
\begin{align}
\lambda_{contact}(t) &= \beta(t) \cdot s(t) \cdot \frac{I + \alpha A}{N} \cdot f_{contact}(t) \\
\lambda_{water}(t) &= \delta(t) \cdot s(t) \cdot W \cdot f_{water}(t)
\end{align}

The parameters are:
\begin{itemize}
    \item $\beta(t)$: Time-varying contact transmission rate
    \item $\delta(t)$: Time-varying waterborne transmission rate
    \item $s(t)$: Seasonal forcing factor (affects both routes)
    \item $f_{contact}(t)$, $f_{water}(t)$: Intervention effectiveness factors
    \item $\alpha$: Relative transmissibility of asymptomatic individuals
\end{itemize}

\subsubsection*{Multi-Route Wave Dynamics}

Waterborne diseases can exhibit complex temporal patterns due to independent variation in both transmission routes. The model implements time-varying transmission coefficients for both pathways:

\begin{align}
\beta(t) &= \beta_i \quad \text{for } t \in [t_{i-1}, t_i) \\
\delta(t) &= \delta_i \quad \text{for } t \in [t_{i-1}, t_i)
\end{align}

Wave changes can reflect:
\begin{itemize}
    \item \textbf{Contact route changes}: Behavioral modifications, crowding patterns, hygiene practices
    \item \textbf{Water route changes}: Infrastructure failures, seasonal water quality, sanitation disruptions
    \item \textbf{Coordinated changes}: Public health interventions affecting both routes simultaneously
\end{itemize}

\subsubsection*{Environmental Contamination Process}

\paragraph{Contamination Sources}
Both symptomatic and asymptomatic individuals contribute to environmental contamination:

\begin{align}
\text{Daily contamination} = \eta_I \cdot I + \eta_A \cdot A \cdot \alpha_{env}
\end{align}

where $\eta_I$ and $\eta_A$ represent base contamination rates, and $\alpha_{env}$ is the relative environmental shedding from asymptomatic individuals.

\paragraph{Pathogen Persistence and Decay}
Environmental pathogens undergo exponential decay:

\begin{align}
\frac{dW}{dt}\Big|_{\text{decay}} = -\mu_w W
\end{align}

The decay rate $\mu_w$ encompasses multiple removal mechanisms:
\begin{itemize}
    \item Natural pathogen die-off
    \item Water treatment processes
    \item Dilution effects
    \item UV radiation and temperature effects
\end{itemize}

\subsubsection*{Stochastic Implementation}

The continuous-time model is implemented stochastically with daily time steps:

\paragraph{Human Transmission Events}
\begin{align}
\text{Contact exposures} &\sim \text{Binomial}(S(t), 1-e^{-\lambda_{contact}(t)}) \\
\text{Water exposures} &\sim \text{Binomial}(S(t), 1-e^{-\lambda_{water}(t)}) \\
\text{Total new exposures} &= \text{Contact exposures} + \text{Water exposures}
\end{align}

Note: In the stochastic implementation, total exposures are calculated as:
\begin{align}
\text{Total exposures} &\sim \text{Binomial}(S(t), 1-e^{-(\lambda_{contact}(t) + \lambda_{water}(t))})
\end{align}

\paragraph{Environmental Process Updates}
\begin{align}
\text{New contamination} &= \eta_I \cdot I(t) + \eta_A \cdot A(t) \cdot \alpha_{env} \\
\text{Pathogen decay} &= \mu_w \cdot W(t) \\
W(t+1) &= \max(0, W(t) + \text{New contamination} - \text{Pathogen decay})
\end{align}

\paragraph{Compartment Transitions}
Human compartment transitions follow the same binomial processes as the human-to-human model:
\begin{align}
\text{E} \to \text{I/A} &\sim \text{Binomial}(E(t), 1-e^{-\sigma}) \\
\text{Recovery from I} &\sim \text{Binomial}(I(t), 1-e^{-\gamma}) \\
\text{Recovery from A} &\sim \text{Binomial}(A(t), 1-e^{-\gamma_a}) \\
\text{Waning immunity} &\sim \text{Binomial}(R(t), 1-e^{-\omega})
\end{align}

\subsubsection*{Enhanced Seasonality}

Waterborne diseases exhibit strong seasonal patterns due to environmental factors affecting both transmission routes and pathogen survival:

\begin{align}
s(t) = 1.0 + \sum_{j=1}^{n_h} A_j \cos\left(\frac{2\pi(t - \phi_j + \delta_y)}{P_j}\right) + \epsilon_t
\end{align}

\paragraph{Seasonal Factors}
Seasonality in waterborne diseases reflects:
\begin{itemize}
    \item \textbf{Rainfall patterns}: Affecting water contamination and treatment efficacy
    \item \textbf{Temperature effects}: Influencing pathogen survival in water
    \item \textbf{Flooding events}: Disrupting sanitation infrastructure
    \item \textbf{Social patterns}: Seasonal migration, gatherings, and water usage
\end{itemize}

\paragraph{Enhanced Amplitude Range}
Waterborne diseases show stronger seasonal variation with amplitude parameters:
\begin{align}
A_j \in [0.6, 1.5] \text{ (compared to } [0.1, 0.5] \text{ for human-to-human)}
\end{align}

\subsubsection*{Dual-Route Intervention Modeling}

Interventions can target either or both transmission routes with different effectiveness:

\begin{align}
\lambda_{contact,eff}(t) &= \lambda_{contact}(t) \cdot f_{contact}(t) \\
\lambda_{water,eff}(t) &= \lambda_{water}(t) \cdot f_{water}(t)
\end{align}

\paragraph{Contact Route Interventions}
Similar to human-to-human diseases:
\begin{itemize}
    \item Social distancing measures
    \item Hygiene promotion
    \item Isolation of cases
\end{itemize}

\paragraph{Water Route Interventions}
Specific to environmental transmission:
\begin{itemize}
    \item Water treatment and chlorination
    \item Sanitation infrastructure improvements
    \item Safe water distribution
    \item Waste management systems
\end{itemize}

\paragraph{Intervention Policy}
The universal intervention policy can independently affect both routes:
\begin{align}
f_{contact}(t) &\in [0.2, 0.6] \text{ when intervention active} \\
f_{water}(t) &\in [0.3, 0.7] \text{ when intervention active}
\end{align}

\subsubsection*{Super-Spreading Events}

Super-spreading affects only the contact transmission route, as waterborne transmission is mediated by environmental persistence rather than individual variation:

\begin{align}
\lambda_{contact,ss}(t) = \lambda_{contact}(t) \cdot m_{ss}(t)
\end{align}

where $m_{ss}(t)$ follows the same gamma-distributed multiplier as in human-to-human transmission.

\subsubsection*{Initialization Strategy}

\paragraph{Human Population}
Initial conditions allow for higher pre-existing immunity due to endemic circulation:
\begin{align}
\text{Immune fraction} &\sim \text{Beta}(10, 3.5) \text{ (higher mean than H2H)} \\
\text{Initial infections} &\sim \text{NegativeBinomial}(n=N_h \times 10^{-6}, p=0.5)
\end{align}

\paragraph{Environmental Pathogen Load}
The water compartment initializes at zero, building up through contamination from initial infections:
\begin{align}
W(0) = 0
\end{align}

\subsubsection*{Parameter Ranges}

Table~\ref{tab:waterborne_params} summarizes all parameters specific to the waterborne transmission simulator.

\begin{longtable}{|p{3.5cm}|p{2.5cm}|p{6cm}|p{2cm}|}

\caption{Waterborne Transmission Simulator Parameters}
\label{tab:waterborne_params} \\
\hline
\textbf{Parameter} & \textbf{Symbol} & \textbf{Range/Distribution} & \textbf{Units} \\
\hline
\endfirsthead 

\multicolumn{4}{c}
{{\bfseries \tablename\ \thetable{} -- continued from previous page}} \\
\hline
\textbf{Parameter} & \textbf{Symbol} & \textbf{Range/Distribution} & \textbf{Units} \\
\hline
\endhead 

\hline \multicolumn{4}{r}{{Continued on next page}} \\ \hline
\endfoot 

\hline
\endlastfoot 

\multicolumn{4}{|c|}{\textbf{Human Population (Shared)}} \\
\hline
Population size & $N$ & Log-uniform[50,000, 40,000,000] \cite{zhou_epidemic_2021,mogelmose_incorporating_2022,yerlanov_effective_2023} & individuals \\
Recovery rate & $\gamma$ & Uniform[0.1, 0.33] \cite{kanungo_cholera_2022,kuehn_enteric_2025} & day$^{-1}$ \\
Latent progression rate & $\sigma$ & Uniform[0.2, 0.4] \cite{azman_incubation_2013,awofisayo-okuyelu_incubation_2018} & day$^{-1}$ \\
Waning immunity rate & $\omega$ & Uniform[0.001, 0.01] \cite{king_inapparent_2008} & day$^{-1}$ \\
Asymptomatic probability & $p_a$ & Beta(3, 7) \cite{king_inapparent_2008} & proportion \\
Asymp. transmissibility & $\alpha$ & Beta(2, 5) \cite{kanungo_cholera_2022} & proportion \\
\hline

\multicolumn{4}{|c|}{\textbf{Dual-Route Transmission}} \\
\hline
Contact transmission rate & $\beta$ & Uniform[0.0, 0.05] \cite{kanungo_cholera_2022} & day$^{-1}$ \\
Water transmission rate & $\delta$ & Uniform[0.0005, 0.01] \cite{kanungo_cholera_2022} & day$^{-1}$ \\
Water contamination rate & $\eta$ & Uniform[0.001, 0.01] \cite{li_survivability_2015,kingsley_functional_2018} & day$^{-1}$ \\
Pathogen decay rate & $\mu_w$ & Uniform[0.05, 0.3] \cite{li_survivability_2015,alam_viable_2007} & day$^{-1}$ \\
Pathogen half-life & $\ln(2)/\mu_w$ & Uniform[2.3, 13.9] \cite{li_survivability_2015} & days \\
\hline

\multicolumn{4}{|c|}{\textbf{Model Structure Options}} \\
\hline
Has latent period & - & Bernoulli(0.7) \cite{kanungo_cholera_2022} & boolean \\
Has asymptomatic compartment & - & Bernoulli(0.6) \cite{king_inapparent_2008} & boolean \\
Has waning immunity & - & Bernoulli(0.9) \cite{king_inapparent_2008} & boolean \\
\hline

\multicolumn{4}{|c|}{\textbf{Multi-Wave Dynamics}} \\
\hline
Multiple waves enabled & - & Configurable (default: True) \cite{semenza_waterborne_2023} & boolean \\
Number of wave changes & - & Uniform[0, 7] \cite{semenza_waterborne_2023} & count \\
Wave change timing & - & Uniform[50, 1800] \cite{semenza_waterborne_2023} & days \\
Contact wave segments & - & Wave count + 1 & count \\
Water wave segments & - & Wave count + 1 & count \\
\hline

\multicolumn{4}{|c|}{\textbf{Enhanced Seasonality}} \\
\hline
Use seasonality & - & Bernoulli(0.9) \cite{shackleton_seasonality_2023,lawrence_grant_influence_2023} & boolean \\
Number of harmonics & $n_h$ & Uniform[1, 4] \cite{shackleton_seasonality_2023} & count \\
Base amplitude & - & Uniform[0.6, 1.5] \cite{shackleton_seasonality_2023} & proportion \\
Harmonic amplitude & $A_j$ & Base $\times$ Uniform[0.3, 1.0] \cite{shackleton_seasonality_2023} & proportion \\
Phase offset & $\phi_j$ & Uniform[0, 365] + Uniform[-60, 60] \cite{lawrence_grant_influence_2023} & days \\
Possible periods & $P_j$ & Choice[365.0, 182.5, 91.25] \cite{shackleton_seasonality_2023} & days \\
Daily noise std & $\sigma_{\epsilon}$ & 0.05 & proportion \\
Annual peak jitter & $\delta_y$ & Uniform[-30, 30] & days \\
\hline

\multicolumn{4}{|c|}{\textbf{Dual-Route Interventions}} \\
\hline
Enable intervention & - & Bernoulli(0.25) \cite{clasen_interventions_2015,wolf_effectiveness_2022} & boolean \\
On threshold & - & $N \times$ Uniform[1e-5, 1e-3] & cases/day \\
Off threshold & - & On threshold $\times$ Uniform[0, 1] & cases/day \\
Contact reduction & $f_{contact}$ & Uniform[0.2, 0.6] \cite{wolf_effectiveness_2022} & proportion \\
Water reduction & $f_{water}$ & Uniform[0.3, 0.7] \cite{clasen_interventions_2015,rao_systematic_2025} & proportion \\
Trigger delay & - & Uniform[0, 21] & days \\
Minimum duration & - & Uniform[14, 35] & days \\
Maximum duration & - & Choice[None, Uniform[60, 120]] & days \\
Consecutive off days & - & Uniform[1, 50] & days \\
\hline

\multicolumn{4}{|c|}{\textbf{Super-Spreading (Contact Only)}} \\
\hline
Super-spreader probability & $p_{ss}$ & Uniform[0.0002, 0.02] \cite{kanungo_cholera_2022} & proportion \\
Multiplier shape & $k_{ss}$ & 4.0 & - \\
Multiplier scale & $\theta_{ss}$ & 1.5 & - \\
\hline

\multicolumn{4}{|c|}{\textbf{Endemic Demographics}} \\
\hline
Enable endemic & - & Bernoulli(0.8) \cite{king_inapparent_2008} & boolean \\
Birth rate & $b$ & Uniform[0.00002, 0.00012] \cite{king_inapparent_2008} & day$^{-1}$ \\
Death rate factor & - & Uniform[0.8, 1.3] & proportion \\
Importation rate & $\iota$ & Exp(Uniform[ln(0.01), ln(0.5)]) \cite{kanungo_cholera_2022} & day$^{-1}$ \\
\hline

\multicolumn{4}{|c|}{\textbf{Initialization Parameters}} \\
\hline
Novel outbreak probability & - & 0.0 (always endemic) \cite{king_inapparent_2008} & proportion \\
Immune fraction (if endemic) & - & Beta(10, 3.5) \cite{king_inapparent_2008} & proportion \\
Initial infections & - & NegBin$(n=N \times 10^{-6}, p=0.5)$ & count \\
Initial water pathogen load & $W(0)$ & 0.0 \cite{kanungo_cholera_2022} & concentration \\
\hline

\multicolumn{4}{|c|}{\textbf{Observation \& Outcomes}} \\
\hline
Hospitalization probability & $p_h$ & Uniform[0.02, 0.15] \cite{vega_ocasio_cholera_2023,tembo_rapid_2025} & proportion \\
Death probability (of hosp.) & $p_d$ & Uniform[0.05, 0.3] \cite{chowdhury_diagnosis_2022} & proportion \\
Hosp. delay shape & $k_h$ & Uniform[2.0, 4.0] & - \\
Hosp. delay scale & $\theta_h$ & Uniform[1.0, 3.0] & days \\
Death delay shape & $k_d$ & Uniform[1.5, 2.5] & - \\
Death delay scale & $\theta_d$ & Uniform[2.0, 5.0] & days \\
\hline

\multicolumn{4}{|c|}{\textbf{Noise \& Reporting (Identical to H2H)}} \\
\hline
Multiplicative noise std & $\sigma_{mult}$ & 0.1 & log-scale \\
Overdispersion parameter & $r$ & 100 & - \\
Reporting effects & - & Same as Table~\ref{tab:h2h_params} & various \\
Day-of-week effects & - & Same as Table~\ref{tab:h2h_params} & various \\
Lab noise parameters & - & Same as Table~\ref{tab:h2h_params} & various \\
Delay distributions & - & Same as Table~\ref{tab:h2h_params} & various \\

\end{longtable}

\subsubsection*{Key Differences from Other Transmission Modes}

The waterborne simulator incorporates several distinguishing features:

\paragraph{Dual Transmission Pathways}
Unlike single-route models, waterborne diseases require simultaneous modeling of direct contact and environmental transmission, each with independent temporal dynamics and intervention susceptibilities.

\paragraph{Environmental Persistence}
The explicit water compartment captures pathogen persistence in the environment, creating temporal delays and amplification effects not present in direct transmission models.

\paragraph{Route-Specific Interventions}
Control strategies must address both transmission pathways, with water-focused interventions (treatment, sanitation) having different effectiveness profiles than contact-focused measures.

\paragraph{Enhanced Seasonal Dependence}
Environmental factors create stronger seasonal patterns than purely social transmission modes, reflected in increased amplitude ranges and environmental sensitivity.

\paragraph{Endemic Circulation Patterns}
Higher baseline immunity levels reflect the endemic nature of many waterborne diseases in affected populations, influencing outbreak dynamics and intervention effectiveness.

This comprehensive dual-route framework enables realistic simulation of waterborne disease dynamics across diverse environmental contexts, intervention strategies, and population immunity profiles, capturing the complex interplay between direct and indirect transmission pathways characteristic of environmentally-mediated pathogens.

\section{Model Architecture}\label{appendix:architecture}

Mantis is implemented as a hybrid convolutional-transformer architecture specifically designed for epidemiological time series forecasting. The model consists of five main components: (1) multi-scale input embeddings, (2) hybrid CNN-transformer encoder blocks, (3) an epidemic pattern memory bank, (4) an autoregressive quantile decoder, and (5) comprehensive temporal feature engineering. This section provides detailed specifications for each component.

\subsection{Input Processing and Feature Engineering}

\subsubsection{Multi-Scale Convolutional Embedding}

The input embedding layer processes time series data using parallel convolutions at multiple temporal scales to capture epidemiological patterns across different time horizons. For input tensor $\mathbf{X} \in \mathbb{R}^{B \times L \times 2}$ where $B$ is batch size, $L$ is sequence length, and the feature dimension includes both target values and covariates (e.g., deaths and hospitalizations), the multi-scale embedding applies four parallel 1D convolutions:

\begin{align}
\mathbf{H}_{\text{short}} &= \text{Conv1D}(\mathbf{X}; k=3, d_{\text{out}}=d/4) \\
\mathbf{H}_{\text{med}} &= \text{Conv1D}(\mathbf{X}; k=7, d_{\text{out}}=d/4) \\
\mathbf{H}_{\text{long}} &= \text{Conv1D}(\mathbf{X}; k=15, d_{\text{out}}=d/4) \\
\mathbf{H}_{\text{vlong}} &= \text{Conv1D}(\mathbf{X}; k=31, d_{\text{out}}=d/4)
\end{align}

where $k$ denotes kernel size, $d$ is the target hidden dimension (default 512), and each convolution uses appropriate padding to maintain sequence length. Each scale captures different temporal patterns:
\begin{itemize}
\item Short-term (k=3): Weekly variations and immediate trends
\item Medium-term (k=7): Monthly patterns and intervention responses
\item Long-term (k=15): Seasonal cycles and extended epidemic phases
\item Very long-term (k=31): Annual seasonality and yearly trends
\end{itemize}

Each convolutional output is independently normalized and activated:
\begin{align}
\tilde{\mathbf{H}}_i &= \text{Dropout}(\text{SiLU}(\text{LayerNorm}(\mathbf{H}_i))) \quad \text{for } i \in \{\text{short, med, long, vlong}\}
\end{align}

The multi-scale features are concatenated and projected to the target hidden dimension:
\begin{align}
\mathbf{H}_{\text{conv}} &= \text{LayerNorm}(\mathbf{W}_{\text{proj}} \cdot \text{Concat}([\tilde{\mathbf{H}}_{\text{short}}, \tilde{\mathbf{H}}_{\text{med}}, \tilde{\mathbf{H}}_{\text{long}}, \tilde{\mathbf{H}}_{\text{vlong}}]))
\end{align}

where $\mathbf{W}_{\text{proj}} \in \mathbb{R}^{d \times d}$ is a learned projection matrix.

\subsubsection{Static and Temporal Feature Embeddings}

Mantis incorporates rich contextual information through multiple embedding layers:

\paragraph{Disease Type Embedding} Disease categories are embedded using a learned lookup table:
\begin{align}
\mathbf{e}_{\text{disease}} = \text{Embedding}(\text{disease\_type}; |V|=3, d_{\text{disease}}=64)
\end{align}
where the three disease types correspond to human-to-human, waterborne, and vector-borne transmission modes from the training simulations.

\paragraph{Population Embedding} Population size is processed through a multi-layer perceptron after log-normalization:
\begin{align}
\mathbf{e}_{\text{pop}} = \text{GELU}(\text{LayerNorm}(\mathbf{W}_{\text{pop}} \cdot \log(1 + \text{population})))
\end{align}
where $\mathbf{W}_{\text{pop}} \in \mathbb{R}^{64 \times 1}$ projects the scalar population to a 64-dimensional embedding.

\paragraph{Enhanced Temporal Embeddings} Temporal patterns are captured through multiple cyclic embeddings derived from absolute day indices. For day index $d_t$ at timestep $t$:

\begin{align}
\mathbf{e}_{\text{dow}} &= \text{LayerNorm}(\text{Embedding}(d_t \bmod 7; |V|=7, d=128)) \\
\mathbf{e}_{\text{month}} &= \text{LayerNorm}(\text{Embedding}(\lfloor d_t/30 \rfloor \bmod 12; |V|=12, d=128)) \\
\mathbf{e}_{\text{doy}} &= \text{LayerNorm}(\text{Embedding}(d_t \bmod 366; |V|=366, d=128))
\end{align}

These embeddings capture day-of-week effects (reporting cycles), monthly seasonality, and annual patterns respectively.

\subsubsection{Feature Integration}

All embedding components are concatenated and projected to the model's hidden dimension. Static embeddings are expanded across the temporal dimension:

\begin{align}
\mathbf{H}_{\text{input}} &= \text{LayerNorm}(\mathbf{W}_{\text{feat}} \cdot \text{Concat}([\mathbf{H}_{\text{conv}}, \mathbf{E}_{\text{disease}}, \mathbf{E}_{\text{pop}}, \mathbf{e}_{\text{dow}}, \mathbf{e}_{\text{month}}, \mathbf{e}_{\text{doy}}]))
\end{align}

where $\mathbf{E}_{\text{disease}}, \mathbf{E}_{\text{pop}} \in \mathbb{R}^{B \times L \times d_{\text{embed}}}$ are the temporally expanded static embeddings, and $\mathbf{W}_{\text{feat}} \in \mathbb{R}^{d \times (d + 64 + 64 + 128 + 128 + 128)}$ projects the concatenated features to hidden dimension $d$.

\subsection{Hybrid CNN-Transformer Encoder}

The encoder consists of $N$ stacked hybrid blocks (default $N=16$), each combining local convolutional processing with global attention mechanisms.

\subsubsection{Temporal Pattern Attention}

Each encoder block uses a custom attention mechanism (`TemporalPatternAttention`) that incorporates relative positional encoding specifically designed for temporal sequences:

\begin{align}
\mathbf{Q}, \mathbf{K}, \mathbf{V} &= \mathbf{H}\mathbf{W}_Q, \mathbf{H}\mathbf{W}_K, \mathbf{H}\mathbf{W}_V \\
\mathbf{A}_{i,j} &= \frac{\mathbf{Q}_i \mathbf{K}_j^T}{\sqrt{d_k}} + \mathbf{R}_{i-j+L-1} \\
\text{Attention}(\mathbf{H}) &= \text{softmax}(\mathbf{A})\mathbf{V}
\end{align}

where $\mathbf{R} \in \mathbb{R}^{2L-1 \times h}$ is a learned relative positional bias matrix with $h$ attention heads, and $\mathbf{R}_{i-j+L-1}$ provides position-dependent bias for the attention score between positions $i$ and $j$.

The relative positional encoding allows the model to learn temporal relationships that are translation-invariant, which is crucial for epidemiological patterns that may occur at different absolute times but with similar relative timing.

\subsubsection{CNN-Transformer Hybrid Block}

Each encoder block (`CNNTransformerBlock`) follows a three-stage design with post-normalization:

\paragraph{Stage 1: Local Pattern Extraction}
\begin{align}
\mathbf{H}^{(1)} &= \text{LayerNorm}(\mathbf{H}^{(0)} + \text{Dropout}(\text{LocalConv}(\mathbf{H}^{(0)})))
\end{align}

where LocalConv applies depthwise separable convolution:
\begin{align}
\text{LocalConv}(\mathbf{H}) &= \text{Conv1D}(\text{DepthwiseConv1D}(\mathbf{H}; k=5); k=1)
\end{align}

The depthwise convolution (kernel size 5) captures local temporal dependencies within each feature channel, followed by a pointwise convolution that mixes information across channels.

\paragraph{Stage 2: Global Context Attention}
\begin{align}
\mathbf{H}^{(2)} &= \text{LayerNorm}(\mathbf{H}^{(1)} + \text{Dropout}(\text{TemporalPatternAttention}(\mathbf{H}^{(1)}, \mathbf{M})))
\end{align}

where $\mathbf{M} \in \{0,1\}^{B \times L}$ is the padding mask indicating valid sequence positions.

\paragraph{Stage 3: Position-wise Feed-Forward}
\begin{align}
\mathbf{H}^{(3)} &= \text{LayerNorm}(\mathbf{H}^{(2)} + \text{Dropout}(\text{FFN}(\mathbf{H}^{(2)})))
\end{align}

where the feed-forward network is:
\begin{align}
\text{FFN}(\mathbf{H}) &= \mathbf{W}_2 \text{GELU}(\mathbf{W}_1 \mathbf{H} + \mathbf{b}_1) + \mathbf{b}_2
\end{align}

with $\mathbf{W}_1 \in \mathbb{R}^{d \times d_{\text{ffn}}}$, $\mathbf{W}_2 \in \mathbb{R}^{d_{\text{ffn}} \times d}$, and $d_{\text{ffn}} = 2048$ by default.

\subsection{Epidemic Pattern Memory Bank}

The pattern memory module (`EpidemicPatternMemory') implements a learnable memory bank that stores prototypical epidemic patterns discovered during training.

\subsubsection{Pattern Bank Initialization}

The pattern bank consists of $K=256$ learnable pattern vectors:
\begin{align}
\mathbf{P} = [\mathbf{p}_1, \mathbf{p}_2, \ldots, \mathbf{p}_K] \in \mathbb{R}^{K \times d}
\end{align}

Pattern vectors are initialized with small random values: $\mathbf{p}_k \sim \mathcal{N}(0, 0.02^2 \mathbf{I})$.

\subsubsection{Pattern Matching and Retrieval}

For encoder output $\mathbf{H}_{\text{enc}} \in \mathbb{R}^{B \times L \times d}$, pattern matching computes attention weights over the pattern bank:

\begin{align}
\mathbf{S} &= \text{FFN}_{\text{match}}(\text{LayerNorm}(\mathbf{H}_{\text{enc}})) \in \mathbb{R}^{B \times L \times K} \\
\mathbf{A}_{\text{pattern}} &= \text{softmax}(\mathbf{S} \odot \mathbf{M}_{\text{exp}}) \\
\mathbf{H}_{\text{retrieved}} &= \mathbf{A}_{\text{pattern}} \mathbf{P} \in \mathbb{R}^{B \times L \times d}
\end{align}

where $\mathbf{M}_{\text{exp}} \in \mathbb{R}^{B \times L \times 1}$ is the expanded padding mask, $\odot$ denotes element-wise multiplication, and $\text{FFN}_{\text{match}}$ is a two-layer network:

\begin{align}
\text{FFN}_{\text{match}}(\mathbf{x}) = \mathbf{W}_3 \text{GELU}(\mathbf{W}_4 \mathbf{x})
\end{align}

with $\mathbf{W}_4 \in \mathbb{R}^{d \times d}$ and $\mathbf{W}_3 \in \mathbb{R}^{d \times K}$.

\subsubsection{Pattern Integration}

Retrieved patterns are projected and integrated with the encoder output using residual connections:

\begin{align}
\mathbf{H}_{\text{memory}} &= \text{LayerNorm}(\mathbf{H}_{\text{enc}} + \text{Dropout}(\mathbf{W}_{\text{proj}} \mathbf{H}_{\text{retrieved}}))
\end{align}

where $\mathbf{W}_{\text{proj}} \in \mathbb{R}^{d \times d}$ is a learned projection matrix.

\subsection{Autoregressive Quantile Decoder}

The decoder generates probabilistic forecasts using an autoregressive approach with a two-layer GRU and cross-attention to encoder memory.

\subsubsection{Decoder Initialization}

The decoder state is initialized using masked pooling of the encoder output:

\begin{align}
\mathbf{h}_0 &= \mathbf{W}_{\text{init}} \cdot \text{MaskedPool}(\mathbf{H}_{\text{memory}}, \mathbf{M}) \\
\text{MaskedPool}(\mathbf{H}, \mathbf{M}) &= \frac{\sum_{t=1}^L \mathbf{M}_{:,t} \odot \mathbf{H}_{:,t,:}}{\sum_{t=1}^L \mathbf{M}_{:,t} + \epsilon}
\end{align}

where $\mathbf{W}_{\text{init}} \in \mathbb{R}^{d \times d}$ and $\epsilon = 10^{-10}$ prevents division by zero. The initial state is replicated for the two-layer GRU: $\mathbf{h}_0^{(1)} = \mathbf{h}_0^{(2)} = \mathbf{h}_0$.

\subsubsection{Autoregressive Generation}

For each forecast step $\tau \in \{1, 2, \ldots, H\}$ where $H$ is the forecast horizon:

\paragraph{Step 1: Input Processing}
The previous prediction $y_{\tau-1}$ is embedded:
\begin{align}
\mathbf{e}_{\tau} = \text{GELU}(\text{LayerNorm}(\mathbf{W}_{\text{input}} y_{\tau-1})) \in \mathbb{R}^{B \times d/4}
\end{align}

\paragraph{Step 2: Cross-Attention}
The top GRU layer output attends to encoder memory:
\begin{align}
\mathbf{q}_{\tau} &= (\mathbf{h}_{\tau-1}^{(2)})^T \in \mathbb{R}^{B \times 1 \times d} \\
\mathbf{c}_{\tau} &= \text{MultiHeadAttention}(\mathbf{q}_{\tau}, \mathbf{H}_{\text{memory}}, \mathbf{H}_{\text{memory}}, \neg\mathbf{M})
\end{align}

where $\neg\mathbf{M}$ serves as the key padding mask (True for padded positions).

\paragraph{Step 3: GRU Update}
The concatenated input is processed through the two-layer GRU:
\begin{align}
\mathbf{x}_{\tau} &= \text{Concat}([\mathbf{c}_{\tau}, \mathbf{e}_{\tau}]) \in \mathbb{R}^{B \times 1 \times (d + d/4)} \\
\mathbf{h}_{\tau}^{(1)}, \mathbf{h}_{\tau}^{(2)} &= \text{GRU}(\mathbf{x}_{\tau}, [\mathbf{h}_{\tau-1}^{(1)}, \mathbf{h}_{\tau-1}^{(2)}])
\end{align}

\paragraph{Step 4: Quantile Prediction}
23 quantiles are predicted using separate projection heads:
\begin{align}
\hat{q}_{\tau}^{(i)} &= \mathbf{W}_{\text{out}}^{(i)} \text{GELU}(\text{LayerNorm}(\mathbf{W}_{\text{hidden}}^{(i)} \mathbf{h}_{\tau}^{(2)})) \\
&\quad \text{for } i \in \{1, 2, \ldots, 23\}
\end{align}

where each quantile head has parameters $\mathbf{W}_{\text{hidden}}^{(i)} \in \mathbb{R}^{(d/2) \times d}$ and $\mathbf{W}_{\text{out}}^{(i)} \in \mathbb{R}^{1 \times (d/2)}$.

The 23 quantiles are: [0.01, 0.025, 0.05, 0.1, 0.15, 0.2, 0.25, 0.3, 0.35, 0.4, 0.45, 0.5, 0.55, 0.6, 0.65, 0.7, 0.75, 0.8, 0.85, 0.9, 0.95, 0.975, 0.99].

\subsubsection{Teacher Forcing}

During training, teacher forcing is applied with linearly decaying probability:
\begin{align}
p_{\text{tf}}(s) &= p_{\text{initial}} \cdot \left(1 - \frac{s}{S_{\text{total}}}\right) + p_{\text{final}} \cdot \frac{s}{S_{\text{total}}}
\end{align}

where $s$ is the current training step, $S_{\text{total}}$ is the total number of training steps, $p_{\text{initial}} = 0.2$, and $p_{\text{final}} = 0.0$.

At each step, with probability $p_{\text{tf}}(s)$, the true target value is used as input instead of the model's prediction:
\begin{align}
y_{\tau} = \begin{cases}
\text{target}_{\tau} & \text{if } U(0,1) < p_{\text{tf}}(s) \text{ and training} \\
\hat{q}_{\tau}^{(5)} & \text{otherwise (median prediction)}
\end{cases}
\end{align}

\subsection{Training Configuration}

\subsubsection{Weight Initialization}

Model parameters are initialized using a careful scheme to ensure stable training:

\begin{align}
\mathbf{W}_{\text{linear}} &\sim \text{Xavier uniform with gain } 0.01 \\
\mathbf{b} &= \mathbf{0} \\
\mathbf{E}_{\text{embedding}} &\sim \mathcal{N}(0, 0.01^2) \\
\mathbf{P}_{\text{pattern}} &\sim \mathcal{N}(0, 0.02^2)
\end{align}

\subsubsection{Loss Function}

The model is trained using quantile loss (pinball loss) averaged across all quantiles:

\begin{align}
\mathcal{L}_{\text{quantile}} &= \frac{1}{BH|Q|} \sum_{b=1}^B \sum_{\tau=1}^H \sum_{q \in Q} \rho_q(y_{b,\tau} - \hat{q}_{b,\tau}^{(q)}) \\
\rho_q(e) &= \max(qe, (q-1)e)
\end{align}

where $Q = \{0.05, 0.1, 0.25, 0.4, 0.5, 0.6, 0.75, 0.9, 0.95\}$ is the set of quantile levels, and $e$ is the prediction error.

\subsubsection{Optimization}

The model is trained using AdamW optimizer with OneCycle learning rate scheduling:

\begin{align}
\text{AdamW}&: \beta_1 = 0.9, \beta_2 = 0.999, \text{weight decay} = 0.01 \\
\text{OneCycle}&: \text{max\_lr} = 10^{-4}, \text{pct\_start} = 0.05, \\
&\quad \text{div\_factor} = 10.0, \text{final\_div\_factor} = 100.0
\end{align}

Gradient clipping is applied with maximum norm 1.0, and training uses gradient accumulation every 64 steps to simulate larger effective batch sizes.

\subsection{Model Scale and Efficiency}

The full model configuration uses:
\begin{itemize}
    \item Hidden dimension: $d = 1024$
    \item Feed-forward dimension: $d_{\text{ffn}} = 2048$  
    \item Number of encoder layers: $N = 16$
    \item Number of attention heads: $h = 16$
    \item Pattern memory size: $K = 256$
    \item Total parameters: approximately 150M
\end{itemize}

Training requires approximately 2 days on a single NVIDIA L4 GPU for one complete pass through the synthetic dataset (we only train with one epoch, since we can generate as much data as we need), while inference can be performed in real-time on standard hardware.

\section{Evaluation Data}\label{appendix:evaluation}

After training, Mantis is evaluated on real-world forecasting tasks spanning multiple diseases, regions, and targets. For each task, the model is run out-of-the-box: no retraining, fine-tuning, or exposure to real data is performed. Forecast accuracy is assessed using both pointwise metrics such as mean absolute error (MAE), weighted interval score (WIS), and coverage metrics. Mantis's performance is compared to established forecasting benchmarks, including the CDC Forecast Hub ensemble and individual submissions \cite{evaluationpnas}, as well as published models such as Chronos \cite{chronos2024}.

\subsection{Primary Evaluation Tasks}

\subsubsection{COVID-19 Mortality Across U.S. States}
To assess generalization to novel outbreak scenarios, we evaluated Mantis on weekly COVID-19 death forecasts across all 50 U.S. states, the District of Columbia, and Puerto Rico during the first two years of the pandemic (April 2020 through November 2021). This evaluation period matches the timeframe used in \cite{evaluationpnas}, enabling direct comparison with published CDC COVID-19 Forecast Hub results.

\paragraph{Data Characteristics}
\begin{itemize}
    \item \textbf{Temporal Coverage}: Weekly data from April 2020 through November 2021 (87 weeks)
    \item \textbf{Geographic Resolution}: 52 jurisdictions (50 states + DC + Puerto Rico)
    \item \textbf{Outcome Variable}: Weekly COVID-19 deaths
    \item \textbf{Population Range}: 580,000 (Wyoming) to 39.5 million (California)
    \item \textbf{Total Forecasts}: Approximately 22,776 individual forecasts across all jurisdictions and horizons
    \item \textbf{Data Source}: CDC COVID-19 surveillance data \cite{cdccovid}
\end{itemize}

\paragraph{Baseline Comparisons}
We benchmarked against models from the U.S. CDC COVID-19 Forecast Hub, which represented expert-curated approaches from modeling teams throughout the pandemic:
\begin{itemize}
    \item \textbf{CDC Forecast Hub ensemble}: Weighted combination of top-performing individual models
    \item \textbf{Individual Hub models}: 28 distinct modeling approaches including statistical, mechanistic, and machine learning methods
    \item \textbf{Covariate utilization}: Most models incorporated extensive covariate information including case counts, mobility data, and syndromic surveillance signals
    \item \textbf{Continuous retraining}: Baselines were retrained as new data became available, reflecting operational practice
\end{itemize}

\subsection{Distribution Shift Evaluation Tasks}

To further assess generalization capabilities under varying degrees of distribution shift, we conducted four additional evaluations on diseases and contexts that challenge different aspects of model robustness.

\subsubsection{Dengue Fever in Brazil (Vector-Borne Transmission)}
We evaluated Mantis on dengue fever case forecasting across Brazilian states from 1993 to 2005, representing real-world vector-borne transmission dynamics within the model's training distribution.

\paragraph{Data Characteristics}
\begin{itemize}
    \item \textbf{Temporal Coverage}: Monthly data from 1993 through 2005 (156 months)
    \item \textbf{Geographic Resolution}: Alagoas, Brazil
    \item \textbf{Outcome Variable}: Monthly dengue fever cases
    \item \textbf{Population Range}: Roughly 3 million
    \item \textbf{Total Forecasts}: Approximately 4,096 individual forecasts
    \item \textbf{Data Source}: Project Tycho \cite{tychodengue}
    \item \textbf{Transmission Mode}: Vector-borne (included in training simulations)
\end{itemize}

\subsubsection{Scarlet Fever in the United States (Historical Context)}
Scarlet fever forecasting across U.S. states from 1910 to 1950 represents historical surveillance contexts with limited infrastructure and irregular reporting patterns.

\paragraph{Data Characteristics}
\begin{itemize}
    \item \textbf{Temporal Coverage}: Weekly data from 1910 through 1950 (40 years)
    \item \textbf{Geographic Resolution}: 48 contiguous U.S. states (Alaska and Hawaii not yet states)
    \item \textbf{Outcome Variable}: Weekly scarlet fever cases
    \item \textbf{Population Range}: 200,000 to 10 million per state (1910-1950 demographics)
    \item \textbf{Total Forecasts}: Approximately 70,560 individual forecasts
    \item \textbf{Data Source}: Project Tycho \cite{tychoscarletfever}
    \item \textbf{Distribution Shift}: Early 20th-century surveillance systems with limited medical infrastructure
\end{itemize}

\subsubsection{Hepatitis B in the United States (Blood-Borne Transmission)}
Hepatitis B forecasting across U.S. states from 1978 to 1988 represents chronic blood-borne infections with long latency periods—a transmission mode explicitly excluded from Mantis's training data.

\paragraph{Data Characteristics}
\begin{itemize}
    \item \textbf{Temporal Coverage}: Weekly data from 1978 through 1988 (11 years)
    \item \textbf{Geographic Resolution}: 50 U.S. states
    \item \textbf{Outcome Variable}: Weekly hepatitis B cases
    \item \textbf{Population Range}: 400,000 to 25 million per state (1980s demographics)
    \item \textbf{Total Forecasts}: Approximately 57,344 individual forecasts
    \item \textbf{Data Source}: Project Tycho \cite{tychohepatitisb}
    \item \textbf{Distribution Shift}: Chronic blood-borne transmission not included in training simulations
\end{itemize}

\subsubsection{Smallpox in the United States (Eradicated Disease)}
Smallpox forecasting across U.S. states from 1925 to 1933 represents an eradicated disease with no modern analog in the training dataset.

\paragraph{Data Characteristics}
\begin{itemize}
    \item \textbf{Temporal Coverage}: Weekly data from 1925 through 1933 (9 years)
    \item \textbf{Geographic Resolution}: 48 contiguous U.S. states
    \item \textbf{Outcome Variable}: Weekly smallpox cases
    \item \textbf{Population Range}: 500,000 to 12 million per state (1920s-1930s demographics)
    \item \textbf{Total Forecasts}: Approximately 21,888 individual forecasts
    \item \textbf{Data Source}: Project Tycho \cite{tychosmallpox}
    \item \textbf{Distribution Shift}: Eradicated disease with no training analog; pre-vaccine era dynamics
\end{itemize}

\subsubsection{Influenza-Like Illness (ILI) in the United States (Syndromic Data)}
ILI forecasting across U.S. states from 2013 to 2025 represents a challenge to Mantis due to the lack of syndromic indicators in its training set.

\paragraph{Data Characteristics}
\begin{itemize}
    \item \textbf{Temporal Coverage}: Weekly data from 2013 through 2025 (13 years) \cite{mistartmap}
    \item \textbf{Geographic Resolution}: 50 U.S. states, Washington D.C., and Puerto Rico.
    \item \textbf{Outcome Variable}: Weekly percentage of outpatient visits due to ILI.
    \item \textbf{Population Range}: 500,000 to 40 million per state
    \item \textbf{Total Forecasts}: Approximately 35,000 individual forecasts
    \item \textbf{Data Source}: CMU DELPHI \cite{delphi_epidata}
    \item \textbf{Distribution Shift}: Syndromic indicator is a data type not seen during training
\end{itemize}

\subsubsection{Communicable Diseases in Thailand}

To assess generalization across endemic disease settings, we evaluated Mantis on weekly communicable disease forecasts across all 77 Thai provinces and the Bangkok Metropolitan Region. This evaluation spans multiple pathogens tracked by Thailand's national surveillance system.

\paragraph{Data Characteristics}
\begin{itemize}
    \item \textbf{Temporal Coverage}: Monthly data from January 2003 through December 2019 (87 months)
    \item \textbf{Geographic Resolution}: 77 jurisdictions (76 Provinces + Bangkok Metropolitan Region)
    \item \textbf{Outcome Variable}: Weekly Communicable Disease Cases
    \item \textbf{Population Range}: 187,000 (Samut Songkhram) to 39.5 million (Bangkok Metropolitan Region)
    \item \textbf{Total Forecasts}: Approximately 121,680 individual forecasts across all jurisdictions and horizons
    \item \textbf{Data Source}: Thai National Notifiable Disease Surveillance System (R506) \cite{Thai_ST}
\end{itemize}

\subsubsection{Meteorological Trends in Thailand}
To assess the impact of weather covariates on forecasting performance, we incorporated provincial-level meteorological data alongside communicable disease case series across all 76 Thai provinces and the Bangkok Metropolitan Region.

\paragraph{Data Characteristics}
\begin{itemize}
    \item \textbf{Temporal Coverage}: Monthly data from January 2003 through December 2019 (87 months)
    \item \textbf{Geographic Resolution}: 77 jurisdictions (76 Provinces + Bangkok Metropolitan Region)
    \item \textbf{Outcome Variable}: Provincial meteorological indicators (e.g., relative humidity, temperature)
    \item \textbf{Population Range}: 187,000 (Samut Songkhram) to 39.5 million (Bangkok Metropolitan Region)
    \item \textbf{Total Forecasts}: Approximately 121,680 individual forecasts across all jurisdictions and horizons
    \item \textbf{Data Source}: ERA5-Land
\end{itemize}

\subsubsection{Ebola Outbreak in Democratic Republic of the Congo}
To assess generalization to novel outbreak scenarios, we evaluated Mantis on weekly Ebola case forecasts across four provinces in the Democratic Republic of the Congo during the 2018--2020 outbreak, the second-largest Ebola outbreak in history.

\paragraph{Data Characteristics}
\begin{itemize}
    \item \textbf{Temporal Coverage}: Weekly data from April 2018 through July 2020 (117 weeks)
    \item \textbf{Geographic Resolution}: 4 jurisdictions (Ituri, North Kivu, South Kivu, Équateur)
    \item \textbf{Outcome Variable}: Weekly Ebola cases and deaths
    \item \textbf{Population Range}: 1,712,000 (Équateur) to 7,574,000 (North Kivu)
    \item \textbf{Total Forecasts}: Approximately 8,848 individual forecasts across all jurisdictions and horizons
    \item \textbf{Data Source}: WHO Ebola outbreak surveillance data \cite{moh_drc_ebola_2020}
\end{itemize}

\subsubsection{Mpox in WHO Africa Countries}
To assess generalization to emerging zoonotic outbreaks, we evaluated Mantis on weekly Mpox case forecasts in Sierra Leone and Uganda, two countries experiencing significant Mpox transmission during the recent WHO-declared public health emergency.

\paragraph{Data Characteristics}
\begin{itemize}
    \item \textbf{Temporal Coverage}: Weekly data from July 2024 to January 2026
    \item \textbf{Geographic Resolution}: 2 countries (Sierra Leone, Uganda)
    \item \textbf{Outcome Variable}: Weekly Mpox cases
    \item \textbf{Population Range}: 8.6 million (Sierra Leone) to 50 million people (Uganda)
    \item \textbf{Total Forecasts}: Approximately 1,328 individual forecasts across all jurisdictions and horizons
    \item \textbf{Data Source}: WHO Mpox surveillance data \cite{who_mpox_global_2024}
\end{itemize}

\subsubsection{Cholera in Ethiopia}
To assess generalization to enteric disease outbreaks in low-resource settings, we evaluated Mantis on weekly cholera case forecasts across Ethiopian regions during a period of sustained transmission.

\paragraph{Data Characteristics}
\begin{itemize}
    \item \textbf{Temporal Coverage}: Weekly data from January 2023 through June 2025 (251 weeks)
    \item \textbf{Geographic Resolution}: Country level
    \item \textbf{Outcome Variable}: Weekly cholera cases
    \item \textbf{Population Range}: 132.1 million people
    \item \textbf{Total Forecasts}: Approximately 3,152 individual forecasts across all jurisdictions and horizons
    \item \textbf{Data Source}: Ethiopian Public Health Institute cholera surveillance data
\end{itemize}

\subsection{Baseline Model Implementations}

For the distribution shift evaluation tasks, we implemented strong statistical baselines to establish performance benchmarks:

\subsubsection{Naïve Persistence Model}
The persistence model assumes that future values will equal the most recent observed value:
\begin{align}
\hat{y}_{t+h} = y_t \quad \text{for all forecast horizons } h
\end{align}

\subsubsection{Exponential Smoothing (ETS)}
We implemented exponential smoothing models with optimal configuration selection via Akaike Information Criterion (AIC). For each disease and forecast window, we evaluated a comprehensive range of ETS variants:
\begin{itemize}
    \item \textbf{Error types}: Additive and multiplicative
    \item \textbf{Trend types}: None, additive, additive damped, multiplicative, multiplicative damped
    \item \textbf{Seasonal types}: None, additive, multiplicative
    \item \textbf{Training procedure}: Rolling forecasts with 104 weeks (or maximum available) of historical context
    \item \textbf{Model selection}: Best configuration selected per disease based on AIC across training period
\end{itemize}

\subsubsection{Seasonal ARIMA (SARIMA)}
We implemented seasonal autoregressive integrated moving average models with automatic order selection. For each disease and forecast window, the model specification was determined using the following procedure:
\begin{itemize}
    \item \textbf{Order selection}: Auto ARIMA algorithm to determine optimal $(p,d,q) \times (P,D,Q)_s$ parameters based on AIC
    \item \textbf{Seasonality}: Period determined by data frequency (e.g., 52 for weekly)
    \item \textbf{Training procedure}: Model refit at each time step using all available historical data up to the forecast origin
    \item \textbf{Forecast generation}: Rolling forecasts with the refitted model producing multi-step-ahead predictions
    \item \textbf{Model selection}: Best SARIMA specification selected per disease based on AIC performance during the initial training period
\end{itemize}

\subsubsection{Long Short-Term Memory Network (LSTM)}
We implemented a deep recurrent neural network baseline using a multi-layer LSTM architecture. The model was retrained at each forecast point to incorporate all available historical data:
\begin{itemize}
    \item \textbf{Architecture}: 6-layer LSTM with hidden dimension of 128 units per layer
    \item \textbf{Regularization}: Dropout rate of 0.5 applied between LSTM layers
    \item \textbf{Data preprocessing}: Standard scaler applied to normalize input time series
    \item \textbf{Training procedure}: Model retrained from scratch at each time step using all available historical data
    \item \textbf{Training duration}: 50 epochs per retraining with early stopping based on validation loss
    \item \textbf{Forecast generation}: Rolling forecasts produced after retraining at each forecast origin
\end{itemize}

\subsection{Data Preprocessing and Quality Control}

\subsubsection{Standardization}
All time series were preprocessed using consistent procedures:
\begin{itemize}
    \item \textbf{Missing value handling}: Linear interpolation for isolated missing points; exclusion of series with $\geq 10\%$ missing data (this leads to most states being excluded for the historical diseases)
    \item \textbf{Minimum length requirements}: Series with $ \leq 52$ observations excluded from evaluation
\end{itemize}

\subsubsection{Forecast Windows}
Evaluation employed rolling forecast windows with consistent protocols:
\begin{itemize}
    \item \textbf{Context length}: Up to 112 weeks of historical data for model input
    \item \textbf{Forecast horizons}: 4 and 8 weeks ahead for primary tasks; 8 weeks for distribution shift tasks
    \item \textbf{Rolling evaluation}: New forecasts generated weekly/monthly as new data became available
    \item \textbf{Evaluation metrics}: Mean Absolute Error (MAE), Mean Absolute Percentage Error (MAPE), 90\% prediction interval coverage
\end{itemize}

\end{appendices}


\begin{thebibliography}{100}

\bibitem{mcbryde2020role}
Emma~S McBryde et~al.
\newblock Role of modelling in covid-19 policy development.
\newblock {\em Paediatric Respiratory Reviews}, 35:57--60, Sep 2020.

\bibitem{evaluationpnas}
Estee~Y. Cramer et~al.
\newblock Evaluation of individual and ensemble probabilistic forecasts of covid-19 mortality in the united states.
\newblock {\em Proceedings of the National Academy of Sciences}, 119(15):e2113561119, 2022.

\bibitem{deepcovid}
Alexander Rodriguez et~al.
\newblock Deepcovid: An operational deep learning-driven framework for explainable real-time covid-19 forecasting.
\newblock In {\em Proceedings of the AAAI Conference on Artificial Intelligence}, volume~35, 2021.

\bibitem{deepgleam}
Dongxia Wu, Liyao Gao, Xinyue Xiong, Matteo Chinazzi, Alessandro Vespignani, Yi-An Ma, and Rose Yu.
\newblock Deepgleam: A hybrid mechanistic and deep learning model for covid-19 forecasting, 2021.

\bibitem{einav_surge_2014}
Sharon Einav, John~L. Hick, Dan Hanfling, Brian~L. Erstad, Eric~S. Toner, Richard~D. Branson, Robert~K. Kanter, Niranjan Kissoon, Jeffrey~R. Dichter, Asha~V. Devereaux, Michael~D. Christian, {Task Force for Mass Critical Care}, and {Task Force for Mass Critical Care}.
\newblock Surge capacity logistics: care of the critically ill and injured during pandemics and disasters: {CHEST} consensus statement.
\newblock {\em Chest}, 146(4 Suppl):e17S--43S, October 2014.

\bibitem{dichter_mass_2022}
Jeffrey~R. Dichter, Asha~V. Devereaux, Charles~L. Sprung, Vikramjit Mukherjee, Jason Persoff, Karyn~D. Baum, Douglas Ornoff, Amit Uppal, Tanzib Hossain, Kiersten~N. Henry, Marya Ghazipura, Kasey~R. Bowden, Henry~J. Feldman, Mitchell~T. Hamele, Lisa~D. Burry, Anne Marie~O. Martland, Meredith Huffines, Pritish~K. Tosh, James Downar, John~L. Hick, Michael~D. Christian, and Ryan~C. Maves.
\newblock Mass {Critical} {Care} {Surge} {Response} {During} {COVID}-19.
\newblock {\em Chest}, 161(2):429--447, February 2022.

\bibitem{funk2019ebola}
Sebastian Funk, Anton Camacho, Adam~J Kucharski, Rachel Lowe, Rosalind~M Eggo, and W~John Edmunds.
\newblock Assessing the performance of real-time epidemic forecasts: A case study of ebola in the western area region of sierra leone, 2014--15.
\newblock {\em PLOS Computational Biology}, 15(2):e1006785, 2019.

\bibitem{reich2022}
Nicholas~G. Reich et~al.
\newblock Collaborative hubs: Making the most of predictive epidemic modeling.
\newblock {\em American Journal of Public Health}, 112(6):839--842, 2022.
\newblock Epub 2022 Apr 14.

\bibitem{runge2024}
Michael~C. Runge et~al.
\newblock Scenario design for infectious disease projections: Integrating concepts from decision analysis and experimental design.
\newblock {\em Epidemics}, 47:100775, 2024.
\newblock Epub 2024 May 24.

\bibitem{gpt3}
Tom~B. Brown, Benjamin Mann, Nick Ryder, Melanie Subbiah, Jared Kaplan, et~al.
\newblock Language models are few-shot learners.
\newblock {\em arXiv}, 2020.

\bibitem{alphafold2}
John Jumper et~al.
\newblock Highly accurate protein structure prediction with alphafold.
\newblock {\em Nature}, 596:583--589, 2021.

\bibitem{clip}
Alec Radford, Jong~Wook Kim, Chris Hallacy, Aditya Ramesh, Gabriel Goh, Sandhini Agarwal, Girish Sastry, Amanda Askell, Pamela Mishkin, Jack Clark, Gretchen Krueger, and Ilya Sutskever.
\newblock Learning transferable visual models from natural language supervision, 2021.

\bibitem{sgnns2025}
Carson Dudley, Reiden Magdaleno, Christopher Harding, and Marisa Eisenberg.
\newblock Simulation as supervision: Mechanistic pretraining for scientific discovery.
\newblock {\em arXiv preprint arXiv:2507.08977}, 2025.

\bibitem{mistartmap}
University of~Michigan.
\newblock Mi safe start map.
\newblock Accessed August 15, 2025.

\bibitem{delphi_epidata}
David~C. Farrow, Logan~C. Brooks, Ryan~J. Tibshirani, and Roni Rosenfeld.
\newblock Delphi epidata api.
\newblock \url{https://github.com/cmu-delphi/delphi-epidata}, 2015.
\newblock Carnegie Mellon University, Delphi Research Group.

\bibitem{cdccovid}
{CDC COVID-19 Response}.
\newblock Weekly united states covid-19 cases and deaths by state - archived.
\newblock \url{https://data.cdc.gov/Case-Surveillance/Weekly-United-States-COVID-19-Cases-and-Deaths-by-/pwn4-m3yp}, 2025.

\bibitem{tychodengue}
Willem Van~Panhuis, Anne Cross, and Donald Burke.
\newblock Counts of dengue without warning signs reported in brazil: 1980--2005 (2.0) [data set], 2018.
\newblock Project Tycho.

\bibitem{tychohepatitisb}
Willem Van~Panhuis, Anne Cross, and Donald Burke.
\newblock Counts of viral hepatitis type b reported in united states of america: 1951--2007 (2.0) [data set], 2018.
\newblock Project Tycho.

\bibitem{tychosmallpox}
Willem Van~Panhuis, Anne Cross, and Donald Burke.
\newblock Counts of smallpox reported in united states of america: 1888--1952 (2.0) [data set], 2018.
\newblock Project Tycho.

\bibitem{tychoscarletfever}
Willem Van~Panhuis, Anne Cross, and Donald Burke.
\newblock Counts of scarlet fever reported in united states of america: 1888--1969 (2.0) [data set], 2018.
\newblock Project Tycho.

\bibitem{wis}
Johannes Bracher, Evan~L. Ray, Tilmann Gneiting, and Nicholas~G. Reich.
\newblock Evaluating epidemic forecasts in an interval format.
\newblock {\em PLOS Computational Biology}, 17(2):e1008618, 2021.

\bibitem{chronos2024}
Abdul~Fatir Ansari et~al.
\newblock Chronos: Transformer-based language models for time-series forecasting, 2024.

\bibitem{time_series_fms}
Suprabhath Kalahasti et~al.
\newblock Foundation time series models for forecasting and policy evaluation in infectious disease epidemics.
\newblock {\em medRxiv}, February 2025.
\newblock preprint, not peer-reviewed.

\bibitem{timepfn}
Ege~Onur Taga, Muhammed~Emrullah Ildiz, and Samet Oymak.
\newblock Time{PFN}: Effective multivariate time series forecasting with synthetic data.
\newblock In {\em NeurIPS Workshop on Time Series in the Age of Large Models}, 2024.

\bibitem{defsi}
Lijing Wang, Jiangzhuo Chen, and Madhav Marathe.
\newblock Defsi: Deep learning based epidemic forecasting with synthetic information.
\newblock In {\em Proceedings of the AAAI Conference on Artificial Intelligence}, volume~33, 2019.

\bibitem{brouwer2026infectious}
Andrew~F Brouwer, Marisa~C Eisenberg, Natalie~E Dean, Harry Hochheiser, Philip Huang, Joseph~R Coyle, and Lior Rennert.
\newblock Infectious disease modeling for public health practice: projections, scenarios, and uncertainty in three phases of outbreak response.
\newblock {\em American Journal of Epidemiology}, page kwag058, 2026.

\bibitem{cobey2020modeling}
Sarah Cobey.
\newblock Modeling infectious disease dynamics.
\newblock {\em Science}, 368(6492):713--714, 2020.

\bibitem{keeling2008modeling}
Matt~J Keeling and Pejman Rohani.
\newblock {\em Modeling infectious diseases in humans and animals}.
\newblock Princeton university press, 2008.

\bibitem{blumberg2013inference}
Seth Blumberg and James~O Lloyd-Smith.
\newblock Inference of r 0 and transmission heterogeneity from the size distribution of stuttering chains.
\newblock {\em PLoS computational biology}, 9(5):e1002993, 2013.

\bibitem{marani2021}
Marco Marani, Gabriel~G. Katul, William~K. Pan, and Anthony~J. Parolari.
\newblock Intensity and frequency of extreme novel epidemics.
\newblock {\em Proceedings of the National Academy of Sciences}, 118(35):e2105482118, August 2021.

\bibitem{paul2021emergence}
Ayan Paul, Jayanta~Kumar Bhattacharjee, Akshay Pal, and Sagar Chakraborty.
\newblock Emergence of universality in the transmission dynamics of covid-19.
\newblock {\em Scientific reports}, 11(1):18891, 2021.

\bibitem{das2021scaling}
Subir~K Das.
\newblock A scaling investigation of pattern in the spread of covid-19: universality in real data and a predictive analytical description.
\newblock {\em Proceedings of the Royal Society A: Mathematical, Physical and Engineering Sciences}, 477(2246), 2021.

\bibitem{riley2007large}
Steven Riley.
\newblock Large-scale spatial-transmission models of infectious disease.
\newblock {\em Science}, 316(5829):1298--1301, 2007.

\bibitem{sattenspiel2009geographic}
Lisa Sattenspiel.
\newblock The geographic spread of infectious diseases: models and applications.
\newblock {\em The Lancet Infectious Diseases}, 2009.

\bibitem{chowell2018spatial}
G~Chowell and R~Rothenberg.
\newblock Spatial infectious disease epidemiology: on the cusp.
\newblock {\em BMC medicine}, 16(1):192, 2018.

\bibitem{bertuzzo2010spatially}
Enrico Bertuzzo, Renato Casagrandi, Marino Gatto, I~Rodriguez-Iturbe, and Andrea Rinaldo.
\newblock On spatially explicit models of cholera epidemics.
\newblock {\em Journal of the Royal Society Interface}, 7(43):321--333, 2010.

\bibitem{gattospatial}
Marino Gatto, Enrico Bertuzzo, Lorenzo Mari, Stefano Miccoli, Luca Carraro, Renato Casagrandi, and Andrea Rinaldo.
\newblock Spread and dynamics of the {COVID}-19 epidemic in {I}taly: {E}ffects of emergency containment measures.
\newblock {\em Proceedings of the National Academy of Sciences}, 117(19), April 2020.

\bibitem{adamw}
Ilya Loshchilov and Frank Hutter.
\newblock Decoupled weight decay regularization, 2019.

\bibitem{LSTM}
Sepp Hochreiter and J{\"u}rgen Schmidhuber.
\newblock Long short-term memory.
\newblock {\em Neural computation}, 9(8):1735--1780, 1997.

\bibitem{dmtest}
Francis~X. Diebold and Roberto~S. Mariano.
\newblock Comparing predictive accuracy.
\newblock {\em Journal of Business \& Economic Statistics}, 13(3):253--263, 1995.

\bibitem{masucci2013gravity}
A~Paolo Masucci, Joan Serras, Anders Johansson, and Michael Batty.
\newblock Gravity versus radiation models: On the importance of scale and heterogeneity in commuting flows.
\newblock {\em Physical Review E—Statistical, Nonlinear, and Soft Matter Physics}, 88(2):022812, 2013.

\bibitem{zhou_epidemic_2021}
Jibiao Zhou, Sheng Dong, Changxi Ma, Yao Wu, and Xiao Qiu.
\newblock Epidemic spread simulation in an area with a high-density crowd using a {SEIR}-based model.
\newblock {\em PloS One}, 16(6):e0253220, 2021.

\bibitem{yerlanov_effective_2023}
Madi Yerlanov, Piyush Agarwal, Caroline Colijn, and Jessica~E. Stockdale.
\newblock Effective population size in simple infectious disease models.
\newblock {\em Journal of Mathematical Biology}, 87(6):80, November 2023.

\bibitem{mogelmose_incorporating_2022}
Signe Møgelmose, Karel Neels, and Niel Hens.
\newblock Incorporating human dynamic populations in models of infectious disease transmission: a systematic review.
\newblock {\em BMC infectious diseases}, 22(1):862, November 2022.

\bibitem{getz_adequacy_2019}
Wayne~M. Getz, Richard Salter, and Whitney Mgbara.
\newblock Adequacy of {SEIR} models when epidemics have spatial structure: {Ebola} in {Sierra} {Leone}.
\newblock {\em Philosophical Transactions of the Royal Society of London. Series B, Biological Sciences}, 374(1775):20180282, June 2019.

\bibitem{hussein_meta-analysis_2021}
Mohammad Hussein, Eman Toraih, Rami Elshazli, Manal Fawzy, August Houghton, Danielle Tatum, Mary Killackey, Emad Kandil, and Juan Duchesne.
\newblock Meta-analysis on {Serial} {Intervals} and {Reproductive} {Rates} for {SARS}-{CoV}-2.
\newblock {\em Annals of Surgery}, 273(3):416--423, March 2021.

\bibitem{guerra_basic_2017}
Fiona~M. Guerra, Shelly Bolotin, Gillian Lim, Jane Heffernan, Shelley~L. Deeks, Ye~Li, and Natasha~S. Crowcroft.
\newblock The basic reproduction number ({R0}) of measles: a systematic review.
\newblock {\em The Lancet. Infectious Diseases}, 17(12):e420--e428, December 2017.

\bibitem{carrat_time_2008}
Fabrice Carrat, Elisabeta Vergu, Neil~M. Ferguson, Magali Lemaitre, Simon Cauchemez, Steve Leach, and Alain-Jacques Valleron.
\newblock Time lines of infection and disease in human influenza: a review of volunteer challenge studies.
\newblock {\em American Journal of Epidemiology}, 167(7):775--785, April 2008.

\bibitem{stone_risk_2025}
Erin~C. Stone, Devon~L. Okasako-Schmucker, Joanna Taliano, Melissa Schaefer, and David~T. Kuhar.
\newblock Risk period for transmission of {SARS}-{CoV}-2 and seasonal influenza: a rapid review.
\newblock {\em Infection Control and Hospital Epidemiology}, 46(3):1--9, February 2025.

\bibitem{lessler_incubation_2009}
Justin Lessler, Nicholas~G. Reich, Ron Brookmeyer, Trish~M. Perl, Kenrad~E. Nelson, and Derek A.~T. Cummings.
\newblock Incubation periods of acute respiratory viral infections: a systematic review.
\newblock {\em The Lancet. Infectious Diseases}, 9(5):291--300, May 2009.

\bibitem{lauer_incubation_2020}
Stephen~A. Lauer, Kyra~H. Grantz, Qifang Bi, Forrest~K. Jones, Qulu Zheng, Hannah~R. Meredith, Andrew~S. Azman, Nicholas~G. Reich, and Justin Lessler.
\newblock The {Incubation} {Period} of {Coronavirus} {Disease} 2019 ({COVID}-19) {From} {Publicly} {Reported} {Confirmed} {Cases}: {Estimation} and {Application}.
\newblock {\em Annals of Internal Medicine}, 172(9):577--582, May 2020.

\bibitem{wu_incubation_2022}
Yu~Wu, Liangyu Kang, Zirui Guo, Jue Liu, Min Liu, and Wannian Liang.
\newblock Incubation {Period} of {COVID}-19 {Caused} by {Unique} {SARS}-{CoV}-2 {Strains}: {A} {Systematic} {Review} and {Meta}-analysis.
\newblock {\em JAMA Network Open}, 5(8):e2228008, August 2022.

\bibitem{berec_protection_2022}
Luděk Berec, Martin Šmíd, Lenka Přibylová, Ondřej Májek, Tomáš Pavlík, Jiří Jarkovský, Milan Zajíček, Jakub Weiner, Tamara Barusová, and Jan Trnka.
\newblock Protection provided by vaccination, booster doses and previous infection against covid-19 infection, hospitalisation or death over time in {Czechia}.
\newblock {\em PloS One}, 17(7):e0270801, 2022.

\bibitem{goldberg_protection_2022}
Yair Goldberg, Micha Mandel, Yinon~M. Bar-On, Omri Bodenheimer, Laurence~S. Freedman, Nachman Ash, Sharon Alroy-Preis, Amit Huppert, and Ron Milo.
\newblock Protection and {Waning} of {Natural} and {Hybrid} {Immunity} to {SARS}-{CoV}-2.
\newblock {\em New England Journal of Medicine}, 386(23):2201--2212, June 2022.

\bibitem{ferdinands_intraseason_2017}
Jill~M. Ferdinands, Alicia~M. Fry, Sue Reynolds, Joshua Petrie, Brendan Flannery, Michael~L. Jackson, and Edward~A. Belongia.
\newblock Intraseason waning of influenza vaccine protection: {Evidence} from the {US} {Influenza} {Vaccine} {Effectiveness} {Network}, 2011-12 through 2014-15.
\newblock {\em Clinical Infectious Diseases}, 64(5):544--550, March 2017.

\bibitem{oran_proportion_2021}
Daniel~P. Oran and Eric~J. Topol.
\newblock The {Proportion} of {SARS}-{CoV}-2 {Infections} {That} {Are} {Asymptomatic}: {A} {Systematic} {Review}.
\newblock {\em Annals of Internal Medicine}, 174(5):655--662, May 2021.

\bibitem{alene_magnitude_2021}
Muluneh Alene, Leltework Yismaw, Moges~Agazhe Assemie, Daniel~Bekele Ketema, Belayneh Mengist, Bekalu Kassie, and Tilahun~Yemanu Birhan.
\newblock Magnitude of asymptomatic {COVID}-19 cases throughout the course of infection: {A} systematic review and meta-analysis.
\newblock {\em PloS One}, 16(3):e0249090, 2021.

\bibitem{wu_assessing_2021}
Peng Wu, Fengfeng Liu, Zhaorui Chang, Yun Lin, Minrui Ren, Canjun Zheng, Yu~Li, Zhibin Peng, Yin Qin, Jianxing Yu, Mengjie Geng, Xiaokun Yang, Hongting Zhao, Zhili Li, Sheng Zhou, Lu~Ran, Benjamin~J. Cowling, Shengjie Lai, Qiulan Chen, Liping Wang, Tim~K. Tsang, and Zhongjie Li.
\newblock Assessing {Asymptomatic}, {Presymptomatic}, and {Symptomatic} {Transmission} {Risk} of {Severe} {Acute} {Respiratory} {Syndrome} {Coronavirus} 2.
\newblock {\em Clinical Infectious Diseases}, 73(6):e1314--e1320, September 2021.

\bibitem{buitrago-garcia_occurrence_2022}
Diana Buitrago-Garcia, Aziz~Mert Ipekci, Leonie Heron, Hira Imeri, Lucia Araujo-Chaveron, Ingrid Arevalo-Rodriguez, Agustín Ciapponi, Muge Cevik, Anthony Hauser, Muhammad~Irfanul Alam, Kaspar Meili, Eric~A. Meyerowitz, Nirmala Prajapati, Xueting Qiu, Aaron Richterman, William~Gildardo Robles-Rodriguez, Shabnam Thapa, Ivan Zhelyazkov, Georgia Salanti, and Nicola Low.
\newblock Occurrence and transmission potential of asymptomatic and presymptomatic {SARS}-{CoV}-2 infections: {Update} of a living systematic review and meta-analysis.
\newblock {\em PLoS medicine}, 19(5):e1003987, May 2022.

\bibitem{an_seasonality_2024}
Tai~Joon An, Jangwon Lee, Myoungin Shin, and Chin~Kook Rhee.
\newblock Seasonality of common respiratory viruses: {Analysis} of nationwide time-series data.
\newblock {\em Respirology}, 29(11):985--993, November 2024.

\bibitem{adam_clustering_2020}
Dillon~C. Adam, Peng Wu, Jessica~Y. Wong, Eric H.~Y. Lau, Tim~K. Tsang, Simon Cauchemez, Gabriel~M. Leung, and Benjamin~J. Cowling.
\newblock Clustering and superspreading potential of {SARS}-{CoV}-2 infections in {Hong} {Kong}.
\newblock {\em Nature Medicine}, 26(11):1714--1719, November 2020.

\bibitem{wang_superspreading_2021}
Jingxuan Wang, Xiao Chen, Zihao Guo, Shi Zhao, Ziyue Huang, Zian Zhuang, Eliza Lai-Yi Wong, Benny Chung-Ying Zee, Marc Ka~Chun Chong, Maggie~Haitian Wang, and Eng~Kiong Yeoh.
\newblock Superspreading and heterogeneity in transmission of {SARS}, {MERS}, and {COVID}-19: {A} systematic review.
\newblock {\em Computational and Structural Biotechnology Journal}, 19:5039--5046, 2021.

\bibitem{goyal_viral_2021}
Ashish Goyal, Daniel~B. Reeves, E.~Fabian Cardozo-Ojeda, Joshua~T. Schiffer, and Bryan~T. Mayer.
\newblock Viral load and contact heterogeneity predict {SARS}-{CoV}-2 transmission and super-spreading events.
\newblock {\em eLife}, 10:e63537, February 2021.

\bibitem{price_association_2019}
Rory Henry~Macgregor Price, Catriona Graham, and Sandeep Ramalingam.
\newblock Association between viral seasonality and meteorological factors.
\newblock {\em Scientific Reports}, 9:929, January 2019.

\bibitem{kucharski_effectiveness_2020}
Adam~J. Kucharski, Petra Klepac, Andrew J.~K. Conlan, Stephen~M. Kissler, Maria~L. Tang, Hannah Fry, Julia~R. Gog, W.~John Edmunds, and {CMMID COVID-19 working group}.
\newblock Effectiveness of isolation, testing, contact tracing, and physical distancing on reducing transmission of {SARS}-{CoV}-2 in different settings: a mathematical modelling study.
\newblock {\em The Lancet. Infectious Diseases}, 20(10):1151--1160, October 2020.

\bibitem{dixon_completeness_2017}
Brian~E. Dixon, Zuoyi Zhang, Patrick T.~S. Lai, Uzay Kirbiyik, Jennifer Williams, Rebecca Hills, Debra Revere, P.~Joseph Gibson, and Shaun~J. Grannis.
\newblock Completeness and timeliness of notifiable disease reporting: a comparison of laboratory and provider reports submitted to a large county health department.
\newblock {\em BMC medical informatics and decision making}, 17(1):87, June 2017.

\bibitem{swaan_timeliness_2018}
Corien Swaan, Anouk van~den Broek, Mirjam Kretzschmar, and Jan~Hendrik Richardus.
\newblock Timeliness of notification systems for infectious diseases: {A} systematic literature review.
\newblock {\em PloS One}, 13(6):e0198845, 2018.

\bibitem{effler_statewide_1999}
Paul Effler, Myra Ching-Lee, April Bogard, Man-Cheng Ieong, Trudi Nekomoto, and Daniel Jernigan.
\newblock Statewide {System} of {Electronic} {Notifiable} {Disease} {Reporting} {From} {Clinical} {Laboratories}: {Comparing} {Automated} {Reporting} {With} {Conventional} {Methods}.
\newblock {\em JAMA}, 282(19):1845--1850, November 1999.

\bibitem{bajema_severity_2025}
Kristina~L. Bajema, David~P. Bui, Lei Yan, Yuli Li, Nallakkandi Rajeevan, Robert Vergun, Kristin Berry, Yuan Huang, Hung-Mo Lin, Mihaela Aslan, and George~N. Ioannou.
\newblock Severity and {Long}-{Term} {Mortality} of {COVID}-19, {Influenza}, and {Respiratory} {Syncytial} {Virus}.
\newblock {\em JAMA Internal Medicine}, 185(3):324--334, March 2025.

\bibitem{longini_statistical_1988}
Ira~M. Longini, James~S. Koopman, Michael Haber, and George~A. Cotsonis.
\newblock Statistical {Inference} for {Infectious} {Diseases}.
\newblock {\em American Journal of Epidemiology}, 128(4):845--859, October 1988.

\bibitem{cdc_dengue_2025}
CDC.
\newblock Dengue, September 2025.

\bibitem{van_boven_estimating_2020}
Michiel van Boven, Anne~C Teirlinck, Adam Meijer, Mariëtte Hooiveld, Christiaan~H van Dorp, Rachel~M Reeves, Harry Campbell, Wim van~der Hoek, and {RESCEU Investigators}.
\newblock Estimating {Transmission} {Parameters} for {Respiratory} {Syncytial} {Virus} and {Predicting} the {Impact} of {Maternal} and {Pediatric} {Vaccination}.
\newblock {\em The Journal of Infectious Diseases}, 222(Supplement\_7):S688--S694, August 2020.

\bibitem{chan_incubation_2012}
Miranda Chan and Michael~A. Johansson.
\newblock The incubation periods of {Dengue} viruses.
\newblock {\em PloS One}, 7(11):e50972, 2012.

\bibitem{paz-bailey_dengue_2024}
Gabriela Paz-Bailey, Laura~E. Adams, Jacqueline Deen, Kathryn~B. Anderson, and Leah~C. Katzelnick.
\newblock Dengue.
\newblock {\em Lancet}, 403(10427):667--682, February 2024.

\bibitem{katzelnick_dengue_2021}
Leah~C. Katzelnick, Jose~Victor Zambrana, Douglas Elizondo, Damaris Collado, Nadezna Garcia, Sonia Arguello, Juan~Carlos Mercado, Tatiana Miranda, Oscarlett Ampie, Brenda~Lopez Mercado, César Narvaez, Lionel Gresh, Raquel~A. Binder, Sergio Ojeda, Nery Sanchez, Miguel Plazaola, Krista Latta, Amy Schiller, Josefina Coloma, Fausto~Bustos Carrillo, Federico Narvaez, M.~Elizabeth Halloran, Aubree Gordon, Guillermina Kuan, Angel Balmaseda, and Eva Harris.
\newblock Dengue and {Zika} virus infections in children elicit cross-reactive protective and enhancing antibodies that persist long term.
\newblock {\em Science Translational Medicine}, 13(614):eabg9478, October 2021.

\bibitem{asish_global_2023}
Purushothaman~Rajamani Asish, Sauvik Dasgupta, Gladys Rachel, Bhavani~Shankara Bagepally, and Chethrapilly~Purushothaman Girish~Kumar.
\newblock Global prevalence of asymptomatic dengue infections - a systematic review and meta-analysis.
\newblock {\em International Journal of Infectious Diseases}, 134:292--298, September 2023.

\bibitem{duong_asymptomatic_2015}
Veasna Duong, Louis Lambrechts, Richard~E. Paul, Sowath Ly, Rath~Srey Lay, Kanya~C. Long, Rekol Huy, Arnaud Tarantola, Thomas~W. Scott, Anavaj Sakuntabhai, and Philippe Buchy.
\newblock Asymptomatic humans transmit dengue virus to mosquitoes.
\newblock {\em Proceedings of the National Academy of Sciences of the United States of America}, 112(47):14688--14693, November 2015.

\bibitem{smith_risk_2004}
David~L. Smith, Jonathan Dushoff, and F.~Ellis McKenzie.
\newblock The risk of a mosquito-borne infection in a heterogeneous environment.
\newblock {\em PLoS biology}, 2(11):e368, November 2004.

\bibitem{amaku_maximum_2013}
Marcos Amaku, Marcelo~Nascimento Burattini, Francisco Antonio~Bezerra Coutinho, Luis~Fernandez Lopez, and Eduardo Massad.
\newblock Maximum equilibrium prevalence of mosquito-borne microparasite infections in humans.
\newblock {\em Computational and Mathematical Methods in Medicine}, 2013:659038, 2013.

\bibitem{matthews_mosquito_2025}
Justin Matthews.
\newblock Mosquito survival from mark-recapture studies releasing at known age.
\newblock {\em Parasites \& Vectors}, 18(1):455, November 2025.

\bibitem{goindin_parity_2015}
Daniella Goindin, Christelle Delannay, Cédric Ramdini, Joël Gustave, and Florence Fouque.
\newblock Parity and longevity of {Aedes} aegypti according to temperatures in controlled conditions and consequences on dengue transmission risks.
\newblock {\em PloS One}, 10(8):e0135489, 2015.

\bibitem{loisel_intra-vector_2025}
Léa Loisel, Vincent Raquin, Maxime Ratinier, Pauline Ezanno, and Gaël Beaunée.
\newblock Intra-vector infection dynamics challenge how to model the extrinsic incubation period for major arboviruses: dengue, {Zika}, and chikungunya.
\newblock {\em PLoS computational biology}, 21(8):e1013393, August 2025.

\bibitem{doeurk_impact_2025}
Bros Doeurk, Sokkeang Leng, Zanory Long, Pierre-Olivier Maquart, and Sébastien Boyer.
\newblock Impact of temperature on survival, development and longevity of {Aedes} aegypti and {Aedes} albopictus ({Diptera}: {Culicidae}) in {Phnom} {Penh}, {Cambodia}.
\newblock {\em Parasites \& Vectors}, 18(1):362, August 2025.

\bibitem{zahid_biting_2023}
Mondal~Hasan Zahid, Hannah Van~Wyk, Amy~C. Morrison, Josefina Coloma, Gwenyth~O. Lee, Varsovia Cevallos, Patricio Ponce, and Joseph N.~S. Eisenberg.
\newblock The biting rate of {Aedes} aegypti and its variability: {A} systematic review (1970-2022).
\newblock {\em PLoS neglected tropical diseases}, 17(8):e0010831, August 2023.

\bibitem{christofferson_how_2022}
Rebecca~C. Christofferson, Helen~J. Wearing, Erik~A. Turner, Christine~S. Walsh, Henrik Salje, Cécile Tran-Kiem, and Simon Cauchemez.
\newblock How do i bite thee? let me count the ways: {Exploring} the implications of individual biting habits of {Aedes} aegypti for dengue transmission.
\newblock {\em PLoS neglected tropical diseases}, 16(10):e0010818, October 2022.

\bibitem{nguyet_host_2013}
Minh~Nguyen Nguyet, Thi Hue~Kien Duong, Vu~Tuan Trung, Than Ha~Quyen Nguyen, Chau N.~B. Tran, Vo~Thi Long, Le~Thi Dui, Hoa~Lan Nguyen, Jeremy~J. Farrar, Edward~C. Holmes, Maia~A. Rabaa, Juliet~E. Bryant, Truong~Thanh Nguyen, Huong Thi~Cam Nguyen, Lan Thi~Hong Nguyen, Mai~Phuong Pham, Hung~The Nguyen, Tai Thi~Hue Luong, Bridget Wills, Chau Van~Vinh Nguyen, Marcel Wolbers, and Cameron~P. Simmons.
\newblock Host and viral features of human dengue cases shape the population of infected and infectious {Aedes} aegypti mosquitoes.
\newblock {\em Proceedings of the National Academy of Sciences of the United States of America}, 110(22):9072--9077, May 2013.

\bibitem{lambrechts_direct_2023}
Louis Lambrechts, Robert~C. Reiner, M.~Veronica Briesemeister, Patricia Barrera, Kanya~C. Long, William~H. Elson, Alfonso Vizcarra, Helvio Astete, Isabel Bazan, Crystyan Siles, Stalin Vilcarromero, Mariana Leguia, Anna~B. Kawiecki, T.~Alex Perkins, Alun~L. Lloyd, Lance~A. Waller, Uriel Kitron, Sarah~A. Jenkins, Robert~D. Hontz, Wesley~R. Campbell, Lauren~B. Carrington, Cameron~P. Simmons, J.~Sonia Ampuero, Gisella Vasquez, John~P. Elder, Valerie~A. Paz-Soldan, Gonzalo~M. Vazquez-Prokopec, Alan~L. Rothman, Christopher~M. Barker, Thomas~W. Scott, and Amy~C. Morrison.
\newblock Direct mosquito feedings on dengue-2 virus-infected people reveal dynamics of human infectiousness.
\newblock {\em PLoS neglected tropical diseases}, 17(9):e0011593, September 2023.

\bibitem{lowe_dengue_2014}
Rachel Lowe, Christovam Barcellos, Caio A.~S. Coelho, Trevor~C. Bailey, Giovanini~Evelim Coelho, Richard Graham, Tim Jupp, Walter~Massa Ramalho, Marilia~Sá Carvalho, David~B. Stephenson, and Xavier Rodó.
\newblock Dengue outlook for the {World} {Cup} in {Brazil}: an early warning model framework driven by real-time seasonal climate forecasts.
\newblock {\em The Lancet. Infectious Diseases}, 14(7):619--626, July 2014.

\bibitem{franklinos_effect_2019}
Lydia H.~V. Franklinos, Kate~E. Jones, David~W. Redding, and Ibrahim Abubakar.
\newblock The effect of global change on mosquito-borne disease.
\newblock {\em The Lancet. Infectious Diseases}, 19(9):e302--e312, September 2019.

\bibitem{pryce_insecticide-treated_2018}
Joseph Pryce, Marty Richardson, and Christian Lengeler.
\newblock Insecticide-treated nets for preventing malaria.
\newblock {\em The Cochrane Database of Systematic Reviews}, 11(11):CD000363, November 2018.

\bibitem{kharwadkar_clinical_2024}
Sahil Kharwadkar and Nipun Herath.
\newblock Clinical manifestations of dengue, {Zika} and chikungunya in the {Pacific} {Islands}: {A} systematic review and meta-analysis.
\newblock {\em Reviews in Medical Virology}, 34(2):e2521, March 2024.

\bibitem{kanungo_cholera_2022}
Suman Kanungo, Andrew~S. Azman, Thandavarayan Ramamurthy, Jaqueline Deen, and Shanta Dutta.
\newblock Cholera.
\newblock {\em Lancet}, 399(10333):1429--1440, April 2022.

\bibitem{kuehn_enteric_2025}
Rebecca Kuehn, Paul Rahden, Huma~Syed Hussain, Abhilasha Karkey, Farah~Naz Qamar, Priscilla Rupali, and Christopher~M. Parry.
\newblock Enteric (typhoid and paratyphoid) fever.
\newblock {\em Lancet}, 406(10509):1283--1294, September 2025.

\bibitem{azman_incubation_2013}
Andrew~S. Azman, Kara~E. Rudolph, Derek A.~T. Cummings, and Justin Lessler.
\newblock The incubation period of cholera: a systematic review.
\newblock {\em The Journal of Infection}, 66(5):432--438, May 2013.

\bibitem{awofisayo-okuyelu_incubation_2018}
Adedoyin Awofisayo-Okuyelu, Noel McCarthy, Ifunanya Mgbakor, and Ian Hall.
\newblock Incubation period of typhoidal salmonellosis: a systematic review and meta-analysis of outbreaks and experimental studies occurring over the last century.
\newblock {\em BMC infectious diseases}, 18(1):483, September 2018.

\bibitem{king_inapparent_2008}
Aaron~A. King, Edward~L. Ionides, Mercedes Pascual, and Menno~J. Bouma.
\newblock Inapparent infections and cholera dynamics.
\newblock {\em Nature}, 454(7206):877--880, August 2008.

\bibitem{li_survivability_2015}
X.~Q. Li, M.~Wang, Z.~A. Deng, J.~C. Shen, X.~Q. Zhang, Y.~F. Liu, Y.~S. Cai, X.~W. Wu, and B.~Di.
\newblock Survivability and molecular variation in {Vibrio} cholerae from epidemic sites in {China}.
\newblock {\em Epidemiology and Infection}, 143(2):288--297, January 2015.

\bibitem{kingsley_functional_2018}
Robert~A. Kingsley, Gemma Langridge, Sarah~E. Smith, Carine Makendi, Maria Fookes, Tom~M. Wileman, Moataz~Abd El~Ghany, A.~Keith~Turner, Zoe~A. Dyson, Sushmita Sridhar, Derek Pickard, Sally Kay, Nicholas Feasey, Vanessa Wong, Lars Barquist, and Gordon Dougan.
\newblock Functional analysis of {Salmonella} {Typhi} adaptation to survival in water.
\newblock {\em Environmental Microbiology}, 20(11):4079--4090, November 2018.

\bibitem{alam_viable_2007}
Munirul Alam, Marzia Sultana, G.~Balakrish Nair, A.~K. Siddique, Nur~A. Hasan, R.~Bradley Sack, David~A. Sack, K.~U. Ahmed, A.~Sadique, H.~Watanabe, Christopher~J. Grim, A.~Huq, and Rita~R. Colwell.
\newblock Viable but nonculturable {Vibrio} cholerae {O1} in biofilms in the aquatic environment and their role in cholera transmission.
\newblock {\em Proceedings of the National Academy of Sciences of the United States of America}, 104(45):17801--17806, November 2007.

\bibitem{semenza_waterborne_2023}
Jan~C. Semenza and Albert~I. Ko.
\newblock Waterborne {Diseases} {That} {Are} {Sensitive} to {Climate} {Variability} and {Climate} {Change}.
\newblock {\em New England Journal of Medicine}, 389(23):2175--2187, December 2023.
\newblock \_eprint: https://www.nejm.org/doi/pdf/10.1056/NEJMra2300794.

\bibitem{shackleton_seasonality_2023}
Debbie Shackleton, Theo Economou, Fayyaz~Ali Memon, Albert Chen, Shanta Dutta, Suman Kanungo, and Alok Deb.
\newblock Seasonality of cholera in {Kolkata} and the influence of climate.
\newblock {\em BMC infectious diseases}, 23(1):572, September 2023.

\bibitem{lawrence_grant_influence_2023}
Stephen Lawrence~Grant, Sina Lange, Sara Almeida, Bilqis Hoque, and Peter Kjær Mackie~Jensen.
\newblock Influence of {Seasonal} {Hazards} on {Water}, {Sanitation}, and {Hygiene}-{Related} {Behavior} and {Implications} for {Cholera} {Transmission} in {Bangladesh}.
\newblock {\em The American Journal of Tropical Medicine and Hygiene}, 108(3):518--523, March 2023.

\bibitem{clasen_interventions_2015}
Thomas~F. Clasen, Kelly~T. Alexander, David Sinclair, Sophie Boisson, Rachel Peletz, Howard~H. Chang, Fiona Majorin, and Sandy Cairncross.
\newblock Interventions to improve water quality for preventing diarrhoea.
\newblock {\em The Cochrane Database of Systematic Reviews}, 2015(10):CD004794, October 2015.

\bibitem{wolf_effectiveness_2022}
Jennyfer Wolf, Sydney Hubbard, Michael Brauer, Argaw Ambelu, Benjamin~F. Arnold, Robert Bain, Valerie Bauza, Joe Brown, Bethany~A. Caruso, Thomas Clasen, John~M. Colford, Matthew~C. Freeman, Bruce Gordon, Richard~B. Johnston, Andrew Mertens, Annette Prüss-Ustün, Ian Ross, Jeffrey Stanaway, Jeff~T. Zhao, Oliver Cumming, and Sophie Boisson.
\newblock Effectiveness of interventions to improve drinking water, sanitation, and handwashing with soap on risk of diarrhoeal disease in children in low-income and middle-income settings: a systematic review and meta-analysis.
\newblock {\em Lancet}, 400(10345):48--59, July 2022.

\bibitem{rao_systematic_2025}
Gouthami Rao, Emma Wells, Catherine Reynolds, Rebecca Yoo, Erin Kowalsky, Jennifer DeFrance, Karl Linden, and Joe Brown.
\newblock Systematic {Review} of the {Microbiological} {Performance} of {Household} {Water} {Treatment} {Technologies}.
\newblock {\em Environmental Science \& Technology}, 59(41):21776--21789, October 2025.

\bibitem{vega_ocasio_cholera_2023}
Denisse Vega~Ocasio, Stanley Juin, David Berendes, Kristen Heitzinger, Graeme Prentice-Mott, Anne~Marie Desormeaux, Phaimyr~D. Jn~Charles, Jonas Rigodon, Valerie Pelletier, Reginald~Jean Louis, John Vertefeuille, Jacques Boncy, Gerard Joseph, Valusnor Compère, Donald Lafontant, Lesly~L. Andrecy, Edwige Michel, Katilla Pierre, Evenel Thermidor, David Fitter, Yoran Grant-Greene, Matthew Lozier, Samson Marseille, and {CDC Haiti Cholera Response Group}.
\newblock Cholera {Outbreak} - {Haiti}, {September} 2022-{January} 2023.
\newblock {\em MMWR. Morbidity and Mortality Weekly Report}, 72(2):21--25, January 2023.

\bibitem{tembo_rapid_2025}
Deborah Tembo, Nedah~Chikonde Musonda, Miyanda Simwaka, Chipo Nkwemu, and Samson Shumba.
\newblock Rapid progression and short-term mortality during the {January} 2023 to {July} 2024 {Cholera} outbreaks in {Zambia}.
\newblock {\em PloS One}, 20(12):e0330969, 2025.

\bibitem{chowdhury_diagnosis_2022}
Fahima Chowdhury, Allen~G. Ross, Md~Taufiqul Islam, Nigel A.~J. McMillan, and Firdausi Qadri.
\newblock Diagnosis, {Management}, and {Future} {Control} of {Cholera}.
\newblock {\em Clinical Microbiology Reviews}, 35(3):e0021121, September 2022.

\bibitem{Thai_ST}
Thailand bureau of epidemiology department of disease control moph. scrub typhus. accessed january 2, 2021.

\bibitem{moh_drc_ebola_2020}
{DRC Ministry of Health}.
\newblock Ebola cases and deaths in the {North Kivu} {Ebola} outbreak in the {Democratic Republic of the Congo} ({DRC}).
\newblock \url{https://data.humdata.org/dataset/ebola-cases-and-deaths-drc-north-kivu}, 2020.
\newblock Accessed: 2026-04-09.

\bibitem{who_mpox_global_2024}
{World Health Organization}.
\newblock 2022--24 mpox ({Monkeypox}) outbreak: Global trends.
\newblock \url{https://worldhealthorg.shinyapps.io/mpx_global/}, 2024.
\newblock Accessed: 2026-04-09.

\end{thebibliography}
\end{document}